\documentclass[10pt,twocolumn,letterpaper]{article}

\usepackage[pagenumbers]{wacv} 

\usepackage{graphicx}
\usepackage{amsmath}
\usepackage{amssymb}
\usepackage{booktabs}
\usepackage{multirow}
\usepackage{subfigure} 
\usepackage{algorithm}
\usepackage{algorithmic}

\usepackage[pagebackref,breaklinks,colorlinks]{hyperref}

\usepackage[capitalize]{cleveref}
\crefname{section}{Sec.}{Secs.}
\Crefname{section}{Section}{Sections}
\Crefname{table}{Table}{Tables}
\crefname{table}{Tab.}{Tabs.}

\def\wacvPaperID{252} 
\def\confName{WACV}
\def\confYear{2024}

\begin{document}

\title{On the Fly Neural Style Smoothing for Risk-Averse Domain Generalization}

\author{Akshay Mehra\textsuperscript{1}, Yunbei Zhang\textsuperscript{1}, Bhavya Kailkhura\textsuperscript{2}, and Jihun Hamm\textsuperscript{1}\\
{\small \textsuperscript{1}Tulane University \quad \textsuperscript{2}Lawrence Livermore National Laboratory}\\ 
{\tt\small\{amehra, yzhang111, jhamm3\}@tulane.edu, kailkhura1@llnl.gov}\\
}


\maketitle

\begin{abstract}
Achieving high accuracy on data from domains unseen during training is a fundamental challenge in domain generalization (DG).
While state-of-the-art (SOTA) DG classifiers have demonstrated impressive performance across various tasks, they have shown a bias towards domain-dependent information, such as image styles, rather than domain-invariant information, such as image content. This bias renders them unreliable for deployment in risk-sensitive scenarios such as autonomous driving where a misclassification could lead to catastrophic consequences.
To enable risk-averse predictions from a DG classifier, we propose a novel inference procedure, Test-Time Neural Style Smoothing (TT-NSS), that uses a ``style-smoothed'' version of the DG classifier for prediction at test time.
Specifically, the style-smoothed classifier classifies a test image as the most probable class predicted by the DG classifier on random re-stylizations of the test image.
TT-NSS uses a neural style transfer module to stylize a test image on the fly, requires only black-box access to the DG classifier, and crucially, abstains when predictions of the DG classifier on the stylized test images lack consensus.
Additionally, we propose a neural style smoothing (NSS) based training procedure that can be seamlessly integrated with existing DG methods. 
This procedure enhances prediction consistency, improving the performance of TT-NSS on non-abstained samples.
Our empirical results demonstrate the effectiveness of TT-NSS and NSS at producing and improving risk-averse predictions on unseen domains from DG classifiers trained with SOTA training methods on various benchmark datasets and their variations.
\end{abstract}

\section{Introduction}
\label{sec:introduction}
The objective of Domain Generalization (DG) \cite{wang2021generalizing} is to develop models 
that demonstrate remarkable resilience to domain shifts during testing, even without prior knowledge of the test domain during training
This represents a challenging problem, as it is impractical to train a model to be robust to all potential variations that may arise at test time.
For example, previous works \cite{bulusu2020anomalous, hendrycks2019benchmarking, alcorn2019strike,geirhos2018imagenet,beery2018recognition} have demonstrated that variations in styles/textures, weather changes, etc., unseen during training can drastically reduce the classifier's performance. 
Recent works  \cite{geirhos2018imagenet,nam2021reducing,hermann2020origins,baker2018deep} brought to light the fact that predictions from state-of-the-art (SOTA) neural networks are biased towards the information unrelated to the content of the images but are dependent on the image styles, a characteristic that can vary across domains.
Due to the vast practical implications of this problem
many works have studied this problem both analytically \cite{ben2007analysis,ben2010theory,mansour2009domain,shen2018wasserstein,zhao2019learning,johansson2019support,blanchet2019quantifying,mehra2021understanding} and empirically \cite{albuquerque2019generalizing,zhang2021quantifying,ganin2016domain,zhao2018adversarial,qiao2020learning,gulrajani2020search,mehra2022do}.
However, in scenarios such as in autonomous driving, medical diagnoses, or rescue operations involving drones, where misclassifications can have severe consequences, it becomes essential to augment classifiers with abstaining mechanisms or involve humans in the decision-making process \cite{settles2009active,cortes2016learning}.
In this work, we focus on the problem of image classification under distribution shifts which comprise of differences in image styles.

To safeguard the classifier against risky misclassification (and enable risk-averse predictions) we augment the classifier with a capability to defer making a prediction on samples, when it lacks confidence.
However, since the softmax score of the classifier is known to be uncalibrated \cite{hein2019relu,hendrycks2019augmix,hendrycks2016baseline} on data from unseen domains, we propose a novel test-time method that uses neural style information to estimate classifier's confidence in its prediction under style changes. 
\begin{figure*}[tb]
  \centering{\includegraphics[width=0.99\textwidth]{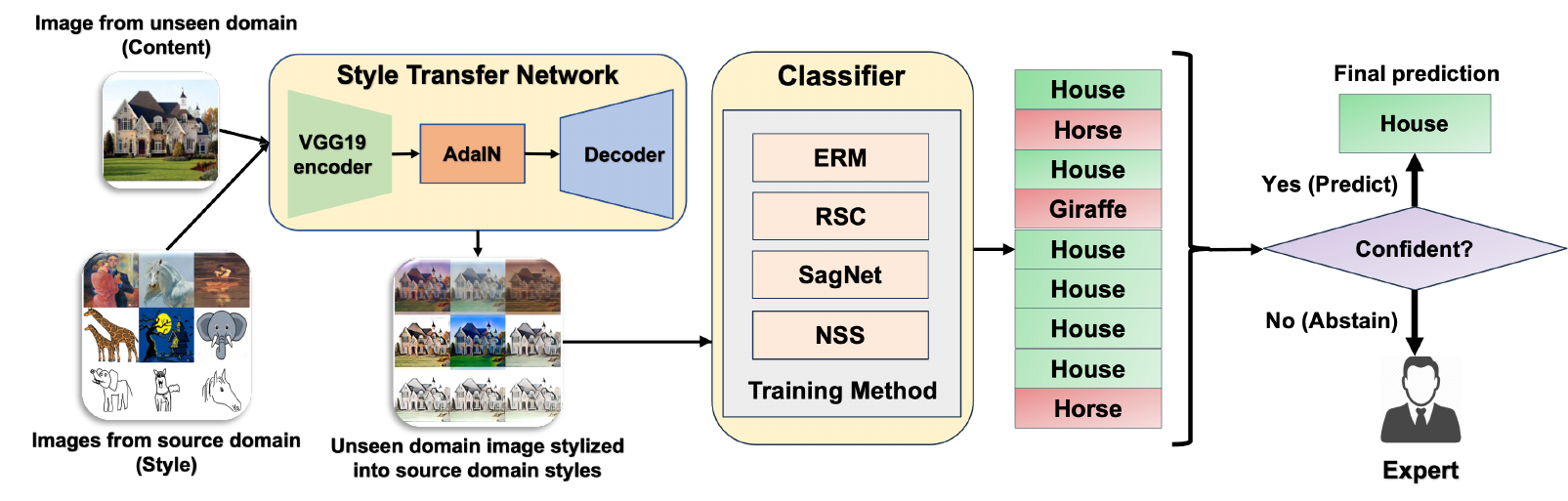}}
  \caption{Overview of our Test-Time Neural Style Smoothing (TT-NSS) inference procedure for obtaining risk-averse predictions. TT-NSS works by stylizing a test sample into source domain styles and classifies the sample as the most probable class assigned by the base DG classifier to the stylized samples if that class is much more likely than the other classes. Otherwise, it abstains from making a prediction and refers the sample to an expert thereby avoiding a risky misclassification.}
  \label{fig:overview}
\end{figure*}
Our inference procedure, Test-Time Neural Style Smoothing (TT-NSS), depicted in Fig.~\ref{fig:overview}, first transforms a classifier (base classifier) into a style-smoothed classifier and then uses it to either predict the label of an incoming test sample or abstain 
on it. 
Specifically, the prediction of the style smoothed classifier, $\psi$, constructed from a base classifier $f$, on a test input $x$ is defined as the class that the base classifier $f$ predicts most 
 frequently on stylized versions of the input.
TT-NSS uses a style transfer network based on AdaIN \cite{huang2017arbitrary} to produce stylized versions of the test input in real-time.
While AdaIN can transform the style of $x$ to any arbitrary style,
we specifically transform it into the style of the data from the domains used for training. 
This choice is based on the assumption that $f$ can be made agnostic to the styles of the data from domains used for training. 
Moreover, changing the styles of $x$ to arbitrary styles, unknown to $f$, can worsen the classifier's performance due to a widened distribution shift. 

TT-NSS can be used to evaluate any DG classifier with only black-box access to it, i.e., it does not require the knowledge of weights, architecture, or training procedure used to train the classifier and only needs its predictions on stylized test samples.
However, computing the prediction of a style-smoothed classifier requires computing the probability with which the base classifier classifies the stylized images of $x$. 
Following works in Randomized Smoothing \cite{cohen2019certified}, we propose a Monte Carlo algorithm to estimate this probability.
When this estimated probability exceeds a set threshold it implies that the predictions of the classifier $f$ on stylized images of $x$ achieve a desired level of consensus and the prediction is reliable. 
In other cases, TT-NSS abstains due to a lack of consensus among the predictions of the base DG classifier. 
Recently, test-time adaptation \cite{iwasawa2021test,zhang2022memo} (TTA) approaches have been shown to be effective in the DG setup which adapts some or all parameters of the classifier using multiple incoming data samples from the unseen domains. 
However, our work differs significantly from these since we consider a black-box setting where parameters of the classifier are not accessible at test time  making our approach much more practically useful compared to TTA approaches. 

Furthermore, we propose a novel training procedure based on neural style smoothing (NSS) to improve the consistency of the predictions of the DG classifier on stylized images. 
The improved consistency leads to improved performance of the DG classifier on non-abstained samples at lower abstaining rates making them more reliable.
Our training method creates a style-smoothed version of the soft base DG classifier and uses stylized versions of the source domain data (generated by stylizing the source domain images into random styles of other source domain images) to train the base DG classifier. 
Similar to previous works \cite{jeong2020consistency,sohn2020fixmatch,sun2021certified}, 
we incorporate consistency regularization during training to further boost 
the performance of the classifier on non-abstained samples at various abstaining rates. 
Similar to TT-NSS which can be used with any classifier, our NSS-based training losses can be combined with any training method and can help improve the reliability of the classifier's predictions without significantly degrading their accuracy or requiring access to auxiliary data from unseen domains \cite{hendrycks2018deep,choi2023conservative}. 
We present results of using our inference and training procedures on PACS \cite{li2017deeper}, VLCS \cite{fang2013unbiased}, OfficeHome \cite{venkateswara2017deep} 
and their variations generated by applying style changes and common corruptions, in both single and multiple source domain settings.
Our results show the effectiveness of our proposed methods at enabling and improving risk-averse predictions from classifiers trained with SOTA DG methods on data from unseen domains.
Our main contributions are summarized below:
\begin{itemize}
    \item We focus on the problem of obtaining risk-averse predictions in a DG setup with black-box access to the classifier. 
    We propose an efficient inference procedure 
    relying on AdaIN-based style transfer and a style-smoothed classifier for classification and abstaining. 

    
    \item To improve the quality of risk-averse predictions, we propose losses that enforce prediction consistency on the random stylization of the source data and can be seamlessly combined with losses of any DG method.
    
    
    \item We demonstrate the effectiveness of our inference and training methods on benchmark datasets and their variations generated by stylizing and using corruptions.
    
\end{itemize}

\section{Related work}
\label{sec:related_work}
{\bf Domain generalization:} The goal of domain generalization (DG) is to produce classifiers whose accuracy remains high when faced with data from domains unseen during training.
Many works have proposed to address this problem by capturing invariances in the data by learning a representation space that reduces the divergence between multiple source domains thereby promoting the use of only domain invariant features for prediction \cite{albuquerque2019generalizing,zhang2021quantifying,ganin2016domain,zhao2018adversarial,qiao2020learning,gulrajani2020search}. 
Another line of work learns to disentangle the style and content information from the source domains and trains the classifier to be agnostic to the styles of the source domains \cite{arjovsky2019invariant,zhang2021towards,dittadi2020transfer,montero2020role}. 
Yet another line of research focuses on diversifying the source domain data to encompass possible variations that may be encountered at test time \cite{hendrycks2019augmix,wang2021augmax,kireev2021effectiveness,calian2021defending,sun2021certified}.
Unlike previous works which focus on improving classifier accuracy on unseen domains, we focus on making DG risk-averse on data from 
unseen domains.

{\bf Certified robustness via randomized smoothing:}
Many works have demonstrated the failure of SOTA machine learning classifiers on adversarial examples 
\cite{szegedy2013intriguing,chen2017ead,xiao2018generating,CPY17zoo,ilyas2018black}. 
In response, many works proposed to provide empirical \cite{athalye2018obfuscated} and provable \cite{li2018second,lecuyer2019certified,cohen2019certified,raghunathan2018certified,zhang2018efficient,mehra2021robust} robustness to these examples. 
Among them, Randomized Smoothing (RS) \cite{li2018second,lecuyer2019certified,cohen2019certified} is a popular method which 
considers a smoothed version of the original classifier and certifies that no adversarial perturbation exists within a certified radius (in $\ell_2$ norm) around a test sample that can change the prediction of the classifier. 
RS uses Gaussian noise to produce a smoothed version of the base classifier and classifies a test sample to be the class most likely to be predicted by the base classifier on Gaussian perturbations of the test sample.
While RS was proposed to certify the robustness to additive noise, the idea has been extended to certify robustness to parameterized transformations of the data such as geometric transformation \cite{fischer2020certified,li2021tss} where the noise is added to the parameters of the transformations. 
Our neural style smoothing procedure is similar to RS with crucial differences. 
Firstly, we use neural styles for smoothing (which cannot be parameterized) instead of adding Gaussian noise to the input or parameters of specific transformations.
Secondly, our goal is not to provide certified robustness guarantees against style changes but to provide a practical method to produce reliable predictions on test samples and an abstaining mechanism to curb incorrect predictions.

{\bf Neural style transfer:}
Following \cite{gatys2016image}, which demonstrated the effectiveness of using the convolutional layers of a convolutional neural network for style transfer, several ways have been proposed to improve style transfer \cite{gatys2017controlling,johnson2016perceptual,ulyanov2016texture,wang2017multimodal,ulyanov2017improved,dumoulin2016learned}. 
AdaIN \cite{huang2017arbitrary} is a popular approach that allows style transfer by changing only the mean and variance of the convolutional feature maps.
Other ways of generating stylized images include mixing \cite{zhou2021domain} or exchanging \cite{tang2020selfnorm, zhao2022source} styles, or using adversarial learning \cite{zhong2022adversarial,shu2021encoding}.

{\bf Test-time adaptation (TTA):}
Recent works have demonstrated the effectiveness of using TTA for improving generalization to unseen domains, where the classifier is updated partially or fully using incoming batches of test samples \cite{wang2020tent,sun2020test,zhang2022memo}. 
This approach has also been shown to be effective in the DG setup \cite{iwasawa2021test}. 
Our approach is different from these methods since we do not assume access to the parameters of the DG classifier or assume that data from unseen domains arrive in batches. 

{\bf Classification with abstaining:} 
A learning framework allowing a classifier to abstain on samples has been studied extensively \cite{chow1970optimum,bartlett2008classification,ni2019calibration,charoenphakdee2021classification,cortes2016learning}. Two main approaches in these works include a confidence-based rejection where the classifier's confidence is used to abstain based on a predefined threshold and a classifier-rejector approach where the classifier and rejector are trained together. 
Our work is closer to the former since we do not train a rejector and abstain when the top class is not much more likely than other classes. 

\section{Neural style smoothing}
\label{sec:style_adaptation}
\subsection{Background}
{\bf Domain Generalization (DG) setup:} Given data samples $\mathcal{D}^i_{\mathrm{source}} = \{(x^i_j, y^i_j)\}_{j=1}^{N^i}$, with $N^i$ samples, from $N_S$ source domains each following a distribution $P^i_S(X,Y)$, the goal of DG is to learn a classifier $f(X)$ whose performance does not degrade on a sample from an unseen test domain with distribution $P_T(X,Y) \neq P^i_S(X,Y)$, for all $i \in \{1, \cdots, N_S\}$.
Depending on the number of source domains available during training the setup can be termed as single or multi-domain. 
The lack of information about the target domain makes the problem setup challenging and many previous works have proposed training methods focusing on capturing domain invariant information from source domain data to improve performance on unseen domains at test time. 
In the multi-domain setup, learning a classifier by minimizing its empirical risk on all available source domains achieves competitive performance on various benchmark datasets \cite{gulrajani2020search}. 

{\bf Neural style transfer with AdaIN \cite{huang2017arbitrary}:} Given a content image, $x_{c}$ and a style image $x_{s}$, AdaIN generates an image having the content of $x_{c}$ and style of $x_{s}$. 
AdaIN works by first extracting the intermediate features (output of \texttt{block4\_conv1}) of the style and content image by passing them through a VGG-19 \cite{simonyan2014very} encoder, $g$, pretrained on Imagenet. 
Using these features AdaIN aligns the mean ($\mu$) and variance ($\sigma$) of the two feature maps using
\begin{equation}
\begin{aligned}
    t &= \mathrm{AdaIN}(g(x_{c}), g(x_{s})) \\
      &= \sigma(g(x_{s})) \left(\frac{g(x_{c}) - \mu(g(x_{c}))}{\sigma(g(x_{c}))} \right) + \mu(g(x_{s})).
\end{aligned}
\end{equation}
A decoder, $h$, is then used to map the AdaIN-generated feature back to the input space to produce a stylized image $x_{\mathrm{stylized}} = h(t)$.
We follow the design of the decoder as proposed in \cite{huang2017arbitrary} and train the 
decoder to minimize the content loss between the features of the stylized image, $g(x_{\mathrm{stylized}})$ and the AdaIN transformed features of the content image, i.e.  
\begin{equation}
\mathcal{L}_{\mathrm{content}} = \|g(x_{\mathrm{stylized}}) - t\|^2_2,
\end{equation}
along with a style loss that measures the distance between the feature statistics of the style and the stylized image using $L$ layers of the pretrained VGG-19 network, $\phi$. 
In particular, the style loss is computed as 
\begin{equation}
\begin{aligned}
\mathcal{L}_{\mathrm{style}} &= \sum_{i=1}^{L} \|\mu(\phi_i(x_s)) - \mu(\phi_i(x_{stylized})\|_2^2 \\ &+ \sum_{i=1}^{L} \|\sigma(\phi_i(x_s)) - \sigma(\phi_i(x_{\mathrm{stylized}})\|_2^2.
\end{aligned}
\end{equation}
We measure the style loss, using \texttt{block1\_conv1, block2\_conv1, block3\_conv1}, and \texttt{block5\_conv1} layers of the VGG-19 network.
We pre-train the decoder with MS-COCO \cite{lin2014microsoft} images as content and Wikiart \cite{wikiart} images as style.

\subsection{Neural style smoothing-based inference}
Consider a classification problem from $\mathbb{R}^d$ to the label space $\mathcal{Y}$. 
Neural style smoothing produces an output, for a test image $x$, that a base DG classifier, $f:\mathbb{R}^d \rightarrow \mathcal{Y}$ is most likely to return when $x$ is stylized into the style of the source domain data, i.e., the data used for training $f$.
Formally, given a base DG classifier $f$, we construct a style-smoothed  classifier $\psi: \mathbb{R}^d \rightarrow \mathcal{Y}$, whose prediction on a test image $x$ is the most probable output of $f$ on $x$ converted into the style of the source domain data, i.e.,
\begin{equation}
    \psi(x) := \arg\max_{y \in \mathcal{Y}} \; \mathbb{P}(f(h(t))=y), 
\end{equation}
where $t=\mathrm{AdaIN}(g(x), g(x_s))$, $x_s \sim P_S$, and $P_S$ is the distribution of the source domain.
When data from multiple source domains are available 
we combine the data from all the domains and use the combined data as source domain data.
If the base  DG classifier, $f$, correctly classifies the test image $x$ when stylized into the styles of the source domain, then the style-smoothed classifier also correctly classifies that sample. 
However, computing the actual prediction of the style-smoothed classifier requires computing the exact probabilities with which the base DG classifier classifies the stylized test samples into each class.
Thus, following \cite{cohen2019certified}, we propose a Monte Carlo algorithm to estimate these probabilities and the prediction of the style-smoothed classifier.
The first step in estimating the prediction of the style-smoothed classifier on a test image $x$ is to generate stylized versions of the image using the styles from the source domain.
To achieve the style conversion in real-time, we use the AdaIN framework described previously with the content image as the test image $x$ and $n$ randomly chosen images from the dataset used for training the DG classifier as style images. 
The style transfer network then transforms $x$ into $n$ stylized images, each having the style of the source domain data, as illustrated in Fig.~\ref{fig:overview}.
The stylized images are then passed through the $f$ and the class that is predicted the most often (majority class) is returned as the prediction of the test image. 
This procedure of Test-Time Neural Style Smoothing (TT-NSS) is detailed in Alg.~\ref{alg:main}. 
\begin{algorithm}[t] 
\caption{Test-Time Neural Style Smoothing (TT-NSS)
} 
\label{alg:main}
\textbf{Input}: Test image $x$, base  DG classifier $f$, VGG-19 encoder $g$, $\mathrm{AdaIN}$ decoder $h$, number of source style images $n$, $\mathcal{D}_{\mathrm{styles}} = \{x_\mathrm{s}^i\}_{i=1}^n$, threshold $\alpha$. \\
\textbf{Output}: Prediction for $x$ or $\mathrm{ABSTAIN}$.

\begin{algorithmic}
\STATE{Initialize class-wise counts $\mathrm{class\_counts}$ to zeros}
\STATE{}
\STATE{\# Generate $n$ stylized images from $x$ using $\mathcal{D}_{\mathrm{styles}}$}
\FOR{$i=1,\;\cdots\;,n$}
    \STATE{$t = \mathrm{AdaIN}(g(x), g(x_\mathrm{s}^i))$}
    \STATE{$x_{\mathrm{stylized}} = h(t)$}
    \STATE{$\mathrm{prediction} = f(x_{\mathrm{stylized}})$}
    \STATE{$\mathrm{class\_counts}[\mathrm{prediction}] += 1$}
\ENDFOR
\STATE{}
\STATE{\# Get the top predicted class on stylized images}
\STATE{$c_{\mathrm{max}}$ = index of $\mathrm{class\_counts}$ with highest count}
\STATE{$n_{\mathrm{max}} = \mathrm{class\_counts}[c_{\mathrm{max}}]$}
\STATE{}
\STATE{\# Predict or ABSTAIN}
\IF{$\frac{n_{\mathrm{max}}}{n} < \alpha$}
\STATE{return $\mathrm{ABSTAIN}$}
\ELSE
\STATE{return $c_{\mathrm{max}}$}
\ENDIF
\end{algorithmic}
\end{algorithm}

To ascertain that the prediction returned by TT-NSS is reliable,
we estimate the confidence of the style-smoothed classifier in its prediction.
In particular, we compute the proportion of the re-stylized test images that are classified as a particular class by the base DG classifier and obtain the counts of how often each class is predicted.
Based on these counts, we compute the class which has the highest occurrence and if the proportion of the highest class exceeds a threshold $\alpha$, TT-NSS classifies the test image as this class. 
However, if the proportion remains less than the threshold, then TT-NSS abstains due to a lack of consensus among the predictions. 
The abstained samples can then be sent for further processing to experts and save the system from returning a potentially incorrect prediction. 
A high value of $\alpha$ in TT-NSS improves the accuracy on non-abstained samples but it also increases the number of abstained samples. 
On the other hand, a low value of $\alpha$ leads to decreased abstaining with an increased chance that the DG classifier may not be confident in its prediction, leading to a risky misclassification. 
In our empirical analysis in Sec.~\ref{sec:experiments}, we use various values of $\alpha$ ranging from $0$ to $1$ and show how the accuracy on non-abstained samples and the proportion of abstained samples change as the value of $\alpha$ is varied.

\begin{figure*}[tb]
  \centering{
  \subfigure[Original  style]{\includegraphics[width=0.24\textwidth]{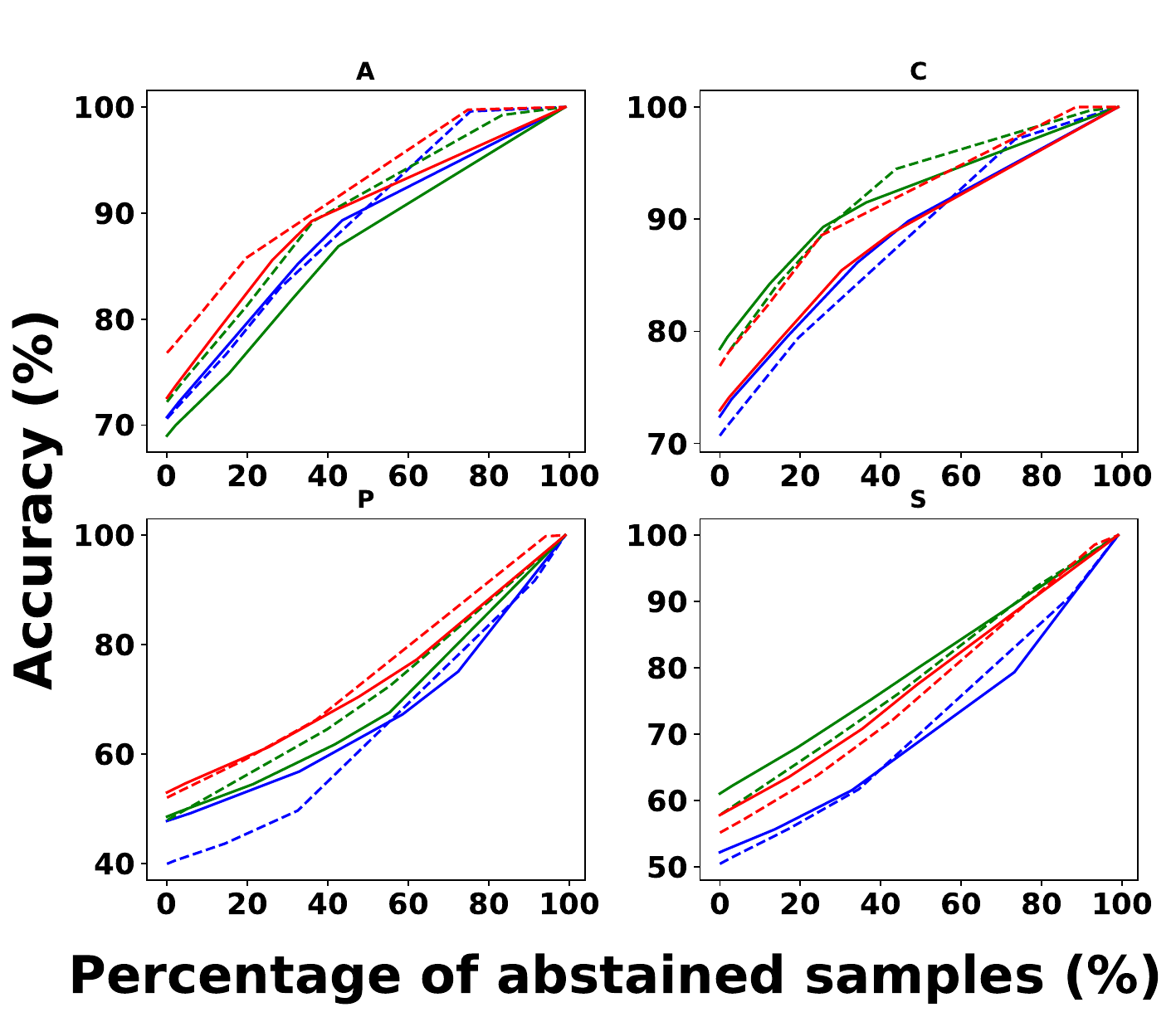}}
  \subfigure[Wikiart  style]{\includegraphics[width=0.24\textwidth]{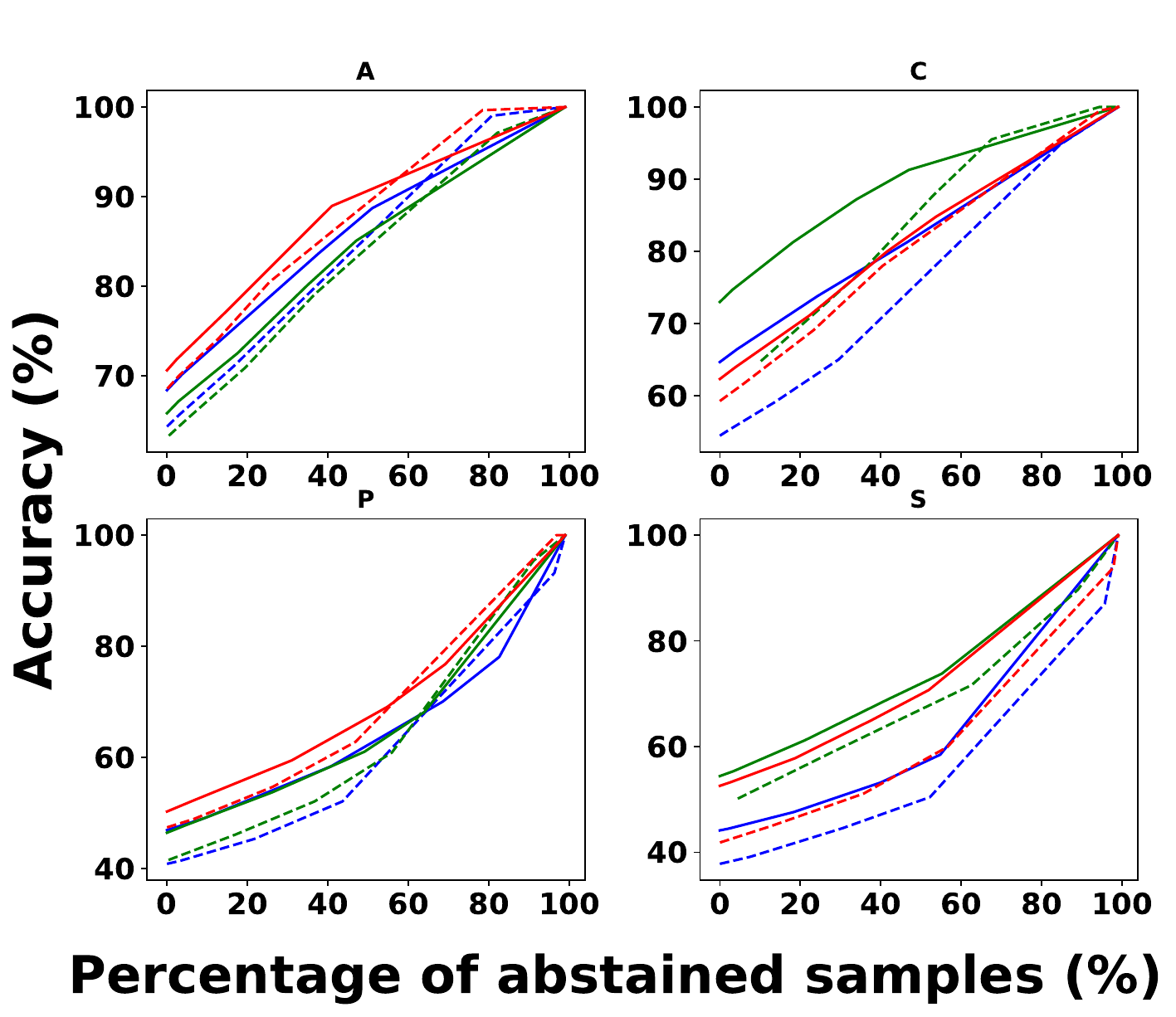}}
  \subfigure[Severity 3 corruptions]{\includegraphics[width=0.24\textwidth]{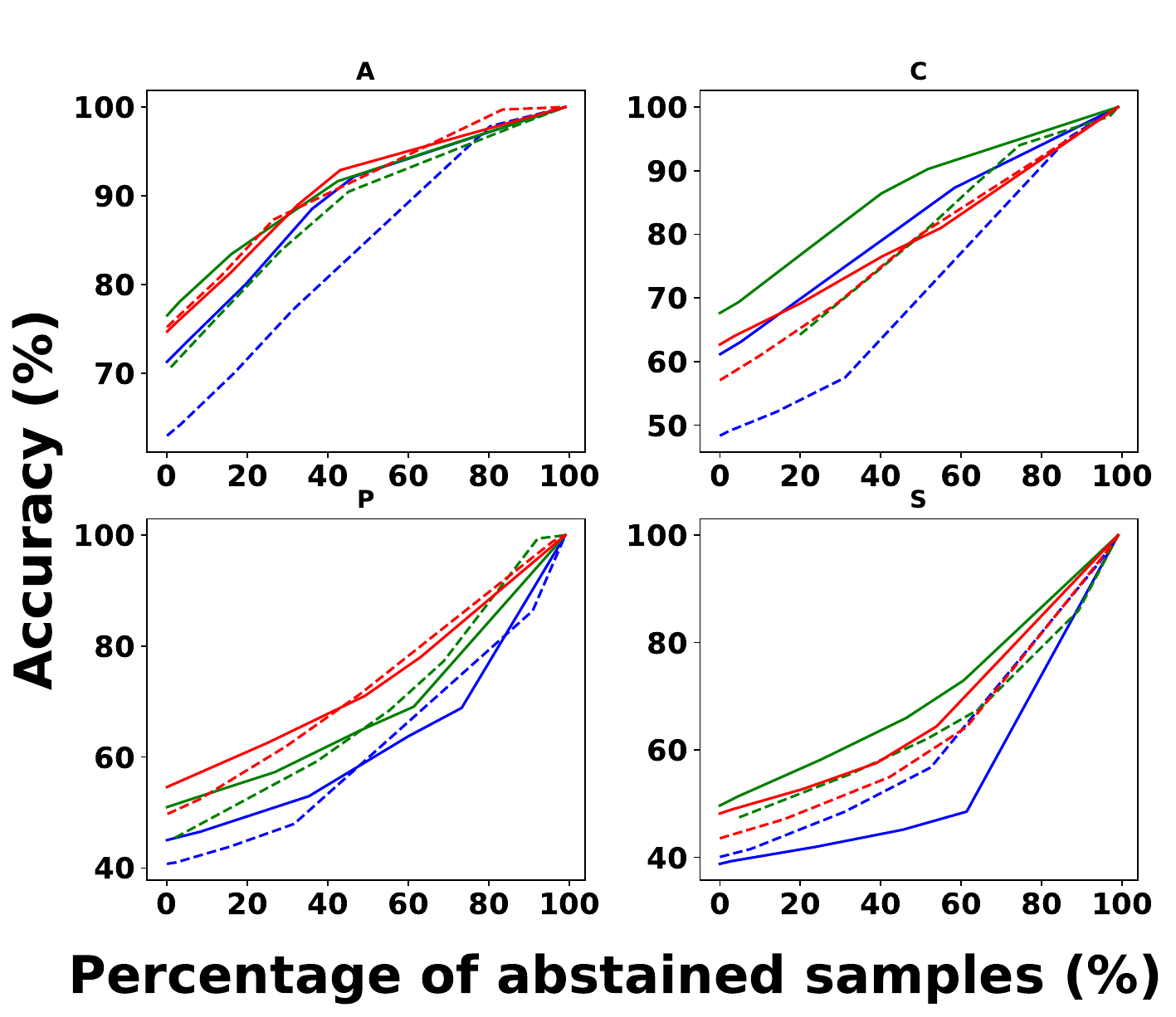}}
  \subfigure[Severity 5 corruptions]{\includegraphics[width=0.24\textwidth]{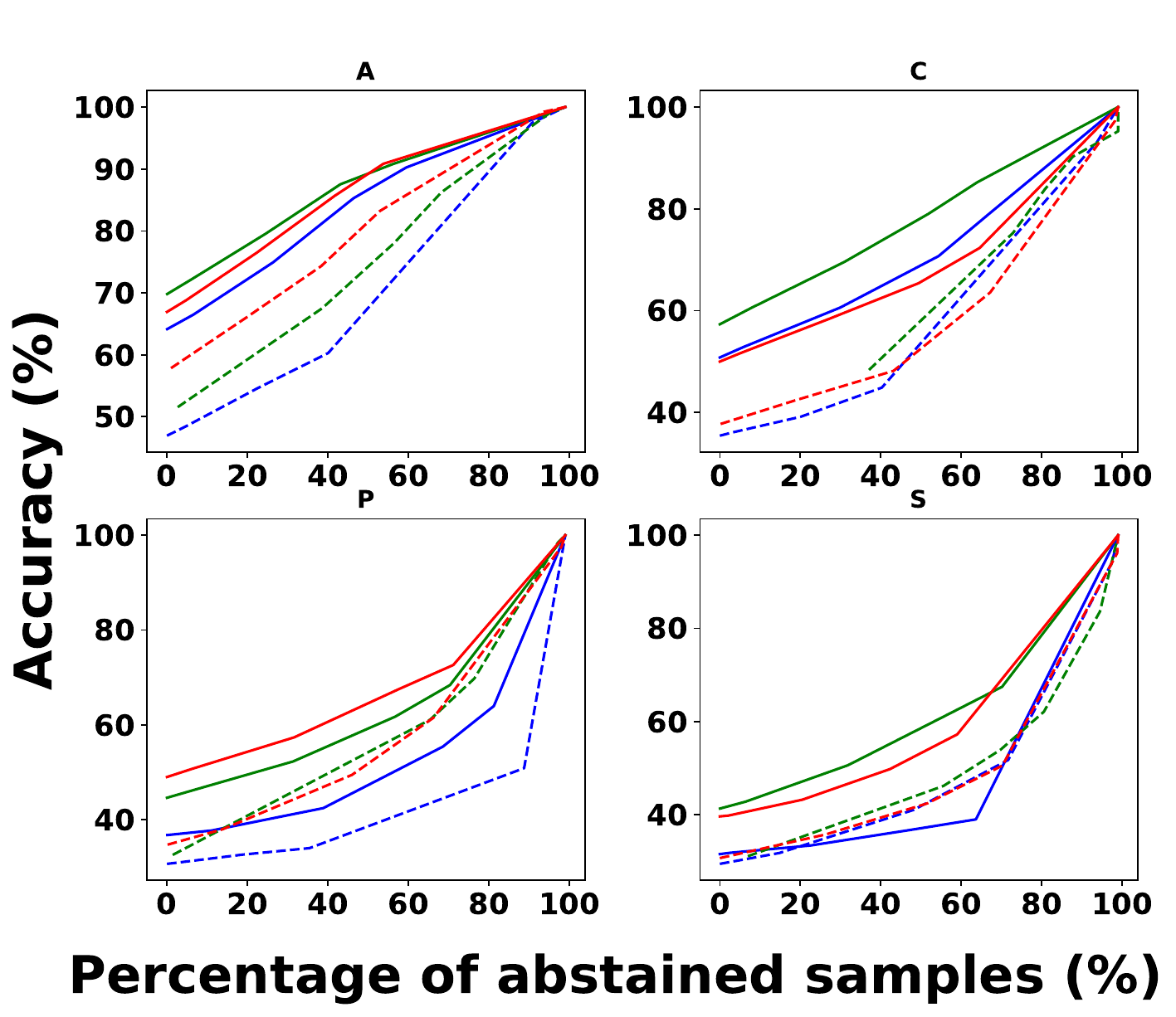}}
  \includegraphics[width=0.25\textwidth]{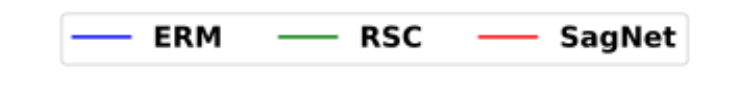}
  }
  \caption{
    Comparison of TT-NSS (solid lines) and confidence-based abstaining method (dashed lines) at producing risk-averse predictions in a {\bf single} source domain setup on classifiers trained with SOTA DG methods. 
    The graphs show accuracy vs abstained points on different variants of the {\bf PACS} dataset ((a) original, (b) wikiart, (c,d) corrupted).
    In most domains, the accuracy of TT-NSS is higher than the corresponding accuracy of the confidence-based method for most of the range of the percentage of abstained samples demonstrating the superiority of TT-NSS 
    at producing risk-averse predictions.
    (Note: The source domain from PACS used for training is denoted in the title.)
    } 
  \label{fig:inference_advantage_sconf_vs_conf_pacs}
\end{figure*}

\subsection{Neural style smoothing-based training}
The performance of our inference procedure, TT-NSS, relies on the assumption that the base  classifier, $f$, can classify the test image stylized into the source domain styles correctly and consistently. 
This requires that the base classifier be accurate on the images generated by the decoder used in the AdaIN-based neural style transfer network. 
However, our empirical evaluation of using TT-NSS on classifiers trained with existing DG methods on benchmark datasets shows a relatively low accuracy on non-abstained samples at smaller abstaining rates. 
This suggests that the base classifier cannot accurately classify the stylized images generated through the AdaIN decoder.
Thus, we propose a new training procedure based on neural style smoothing (NSS) that enables consistent and accurate predictions from the  classifiers when evaluated using TT-NSS. 
The proposed loss functions can be combined with any DG training algorithm and can be used to improve the reliability of the predictions from classifiers when evaluated with TT-NSS. 
To achieve this, we propose to augment the losses of an existing DG method with two additional loss functions.
The first loss penalizes misclassification of the stylized images w.r.t. the label of the content image i.e., 
given a sample $(x,y) \sim \mathcal{D}_\mathrm{source}$, the stylized misclassification loss is 
\begin{equation}
   \mathcal{L}_{stylized\_aug} = \mathbb{E}_{x_s \sim P_S}[\ell(f(h(t)), y)],
\end{equation}
where $t=\mathrm{AdaIN}(g(x), g(x_s))$ and $\ell$ is the cross entropy loss. 
Specifically, we first stylize a sample $x$ from the source domain using multiple randomly sampled style images from the source domain and then penalize the misclassification loss of the classifier $f$ on these stylized images. 
For a single source domain problem, even though all images from a domain may be considered as being in the same broad set of styles such as Art or Photos, individually the images have different non-semantic information such as textures, colors, patterns, etc., and thus stylizing an image into the styles of other source domain images is still effective and meaningful.
The second loss which helps improve the trustworthiness of the predictions enforces consistency among the predictions of the stylized versions of the content image, generated using AdaIN.
Previous works \cite{sohn2020fixmatch,jeong2020consistency,sun2021certified,zhao2022style}, have also demonstrated the effectiveness of enforcing consistency among the predictions of the classifier to be helpful in various setups such as semi-supervised learning and randomized smoothing.
To define the style consistency loss, let $(x,y) \sim \mathcal{D}_\mathrm{source}$, $F:\mathbb{R}^d \rightarrow \Delta^{K-1}$ be the softmax output of the classifier such that the prediction of the base  classifier $f(x) = \arg \max_{k\in\mathcal{Y}}F(x)$, $\Delta^{K-1}$ be the probability simplex in $\mathbb{R}^K$, $\overline{F}(x) = \mathbb{E}_{x_s \sim P_S}[F(h(t))]$ with $t=\mathrm{AdaIN}(g(x), g(x_s))$ be the average softmax output of the  classifier on stylized images, $\mathrm{KL}(\cdot\|\cdot)$ be the Kullback–Leibler divergence (KLD)~\cite{joyce2011kullback} and $\mathrm{H}(\cdot)$ be the entropy.
Then the style consistency loss is given by 
\begin{equation}
    \begin{aligned}
    \mathcal{L}_{consistency} = \mathbb{E}&_{x_s \sim P_S}[\mathrm{KL}(\overline{F}(x) \| F(h(t)))] \\
    &+ \mathrm{H}(\overline{F}(x),y).
    \end{aligned}
\end{equation}
In practice, we minimize the empirical version of the two losses using multiple-style images sampled randomly from the available source domain data. 
The trained classifier can then be evaluated using TT-NSS as  in Alg.~\ref{alg:main} to gauge the reliability of their predictions on unseen domains. 

\section{Experiments}
\label{sec:experiments}
In this section, we present the evaluation results of using our inference and training procedures for obtaining and improving the risk-averse predictions from DG classifiers. 
We present evaluations and comparisons with three popular DG methods, namely Empirical Risk Minimization (ERM), Style Agnostic Networks (SagNet), \cite{nam2021reducing} and networks trained with Representation Self-Challenging (RSC) \cite{huang2020self}.
Our evaluation includes three popular benchmark datasets, namely PACS \cite{li2017deeper}, VLCS \cite{fang2013unbiased} and OfficeHome \cite{venkateswara2017deep}, all of which contain four domains (see Appendix~\ref{app:experimental_details}).
We also create and present evaluations on variations of these datasets generated by stylizing the images into the styles of Wikiart \cite{wikiart} and changing styles based on changes in weather, lighting, blurring, and addition of noise by using common corruptions \cite{hendrycks2018benchmarking} including \texttt{\{frost, fog, brightness, contrast, gaussian blur, defocus blur, zoom blur, gaussian noise, shot noise, impulse noise\}}. 
These variations allow us to evaluate the performance of DG classifiers on realistic changes that do not affect the semantic content of the images. 
To generate images from benchmark datasets stylized into the style of Wikiart, we use  an AdaIN decoder pre-trained using images from MS-COCO\cite{lin2014microsoft} as content images and images from Wikiart \cite{wikiart} as style images. 
To create corrupted versions, we follow \cite{hendrycks2018benchmarking} and use corruption with severity levels 3 and 5. For reporting results over corrupted versions we use a subsample of the test set described in App.~\ref{app:subsample} where as for original/wikiart styles we report results on the entire test set.

Following previous works \cite{gulrajani2020search}, we used ResNet50 pre-trained on the ImageNet dataset as our backbone network augmented with a fully connected layer with softmax activation. 
We use this network for training ERM and for neural style smoothing (combined with ERM as the DG method). 
For other baselines, we train the classifiers using the source codes from the official repositories of RSC \cite{huang2020self} and SagNet \cite{nam2021reducing}.
For all experiments in the single source domain setup, we train the  classifiers with a single source domain and evaluate the performance of the remaining three domains.
For multi-domain setup, we train the classifiers with three domains and test on the fourth unseen domain.

We compare the performance of TT-NSS (Alg.~\ref{alg:main}) with an abstaining mechanism that uses the classifier's max confidence on the original test sample for abstaining.
In this method, we abstain if the highest softmax score for a sample is below a set threshold.
We note that, compared to TT-NSS, which only requires prediction of the classifier on a sample the confidence-based mechanism additionally requires the classifier's confidence in the prediction and hence has access to more information than that available to TT-NSS, making TT-NSS more practically viable. 
For TT-NSS we use 10 randomly sampled style images ($n = 10$) for the single source domain setup and 15 for the multiple source domain setup (see Sec.~\ref{sec:exp_num_styles}). 
We present the accuracy of the DG classifier on non-abstained samples as a function of the proportion of abstained samples and the area under this curve (AUC) to demonstrate the effectiveness of TT-NSS (Alg.~\ref{alg:main}) and the confidence-based abstaining mechanism for producing risk-averse predictions.
A higher AUC is desired since it indicates that the accuracy of the DG classifier at different abstaining rates remains high suggesting that whenever the inference procedure does not abstain, it is likely that the prediction is correct. 
This improves the reliability of the predictions from a DG classifier.
We present additional experimental results in App.~\ref{app:additional_experiments} followed by dataset and implementation details in App.~\ref{app:experimental_details}. Our codes are present at \url{https://github.com/akshaymehra24/RiskAverseDG}

\begin{table*}
  \begin{center}
    \captionof{table}{%
       Effectiveness of NSS at producing a better AUC score compared to classifiers trained with ERM in a {\bf single} source domain setting on PACS, VLCS, and OfficeHome datasets and their variations when evaluated with TT-NSS. The source domain used for training is denoted in the columns. (In all tables, the best result is marked in bold if the difference in the AUC is at least 0.01.)
      \label{table:erm_vs_nss_auc}
    }
    \resizebox{0.75\textwidth}{!}{
    \begin{tabular}{|c|cccc|cccc|cccc|}

      \hline
      & \multicolumn{4}{|c|}{PACS} & \multicolumn{4}{|c|}{VLCS} & \multicolumn{4}{|c|}{OfficeHome}  \\

      \hline
      \multicolumn{1}{|c|}{Alg.} & A & C & P & S & C & L & S & V & A & C & P & R \\
      
      \hline

       & \multicolumn{12}{|c|}{Original Style}  \\
      
      \hline
      
    ERM & 0.875 & 0.878 & 0.662 & 0.702 & 0.567 & {\bf 0.724} & 0.851 & 0.751 & 0.689 & 0.553 & 0.549 & 0.685
    \\
    NSS & {\bf 0.884} & {\bf 0.911} & {\bf 0.694} & {\bf 0.745} & {\bf 0.619} & 0.685 & 0.853 & {\bf 0.796} & {\bf 0.727} & {\bf 0.683} & {\bf 0.675} & {\bf 0.767}
    \\
      
      \hline

      & \multicolumn{12}{|c|}{Wikiart Style}  \\
      
      \hline
      
    ERM & 0.854 & 0.816 & 0.643 & 0.626 & 0.477 & 0.682 & 0.785 & 0.704 & 0.552 & 0.344 & 0.321 & 0.5
    \\
    NSS & 0.855 & {\bf 0.888} & {\bf 0.71} & {\bf 0.706} & {\bf 0.528} & {\bf 0.673} & {\bf 0.845} & {\bf 0.788} & {\bf 0.696} & {\bf 0.643} & {\bf 0.625} & {\bf 0.725}
    \\
      
      \hline

      & \multicolumn{12}{|c|}{Corrupted with severity 3}  \\
      
      \hline
      
    ERM & 0.886 & 0.812 & 0.622 & 0.545 & 0.468 & 0.551 & 0.689 & 0.471 & 0.573 & 0.358 & 0.312 & 0.54
    \\
    NSS & {\bf 0.901} & {\bf 0.853} & {\bf 0.717} & {\bf 0.683} & {\bf 0.573} & {\bf 0.686} & {\bf 0.775} & {\bf 0.608} & {\bf 0.625} & {\bf 0.576} & {\bf 0.56} & {\bf 0.67}
    \\
      
      \hline

      & \multicolumn{12}{|c|}{Corrupted with severity 5}  \\
      
      \hline
      
    ERM & 0.834 & 0.708 & 0.519 & 0.468 & 0.411 & 0.439 & 0.567 & 0.415 & 0.445 & 0.235 & 0.196 & 0.383
    \\
    NSS & {\bf 0.871} & {\bf 0.792} & {\bf 0.682} & {\bf 0.606} & {\bf 0.512} & {\bf 0.61} & {\bf 0.722} & {\bf 0.537} & {\bf 0.545} & {\bf 0.478} & {\bf 0.466} & {\bf 0.565}
    \\
      
      \hline

    \end{tabular}
    }
  \end{center}

\end{table*}

\subsection{TT-NSS improves the reliability of the predictions from existing DG classifiers}
In this section, we demonstrate the effectiveness of TT-NSS at producing reliable predictions from classifiers trained with ERM, RSC, and SagNet when evaluated on domains unseen during training.
The results in Fig.~\ref{fig:inference_advantage_sconf_vs_conf_pacs} and Figs.~\ref{fig:inference_advantage_sconf_vs_conf_vlcs},~\ref{fig:inference_advantage_sconf_vs_conf_pacs_M} (in the Appendix) show the advantage of using the style-smoothed classifier over the confidence of the original classifier for producing risk-averse predictions on a test sample on PACS and VLCS datasets in both single and multiple source domain setting. 
This superiority of TT-NSS is also evident from the results in Tables~\ref{table:conf_vs_sconf_pacs}, \ref{table:conf_vs_sconf_pacs_M}, \ref{table:conf_vs_sconf_vlcs}, \ref{table:conf_vs_sconf_vlcs_M} (in the Appendix) which show the area under the curve for accuracy versus percentage of abstained samples for different settings.  
The high accuracy of the classifiers with TT-NSS at the same abstaining rates compared to the confidence-based strategy shows the advantage of TT-NSS at producing better risk-averse predictions.
This advantage of TT-NSS becomes more apparent on stylized and corrupted variants of the PACS dataset where the standard accuracy of the  classifier  drops significantly and necessitates abstaining for safeguarding against risky misclassifications.
The classifier's high confidence incorrect predictions on unseen domains is the primary reason that prevents the confidence-based strategy from producing risk-averse predictions.
This is in line with the findings from previous works which have shown that a classifier can produce high-confidence misclassification on samples from unseen domains \cite{hein2019relu,mallick2020probabilistic,hendrycks2016baseline, zhang2017mixup, zhang2020leveraging}.
On the other hand, using the confidence of the style-smoothed classifier, by stylizing the test sample into source domain styles, can mitigate the classifier's bias to non-semantic information in the test samples and produce better quality predictions even without abstaining. 
This is evident from Fig.~\ref{fig:inference_advantage_sconf_vs_conf_pacs} and Figs.~\ref{fig:inference_advantage_sconf_vs_conf_vlcs},~\ref{fig:inference_advantage_sconf_vs_conf_pacs_M} (in the Appendix) where TT-NSS (solid lines) achieve higher accuracy even at an abstaining rate of 0\%.

Another crucial insight obtained from our evaluation on variations of benchmark datasets created by style changes is the significant decrease in the performance of the DG classifiers compared to the evaluation on original styles of the benchmark datasets both with confidence-based abstaining and TT-NSS.
This suggests that classifiers trained with existing DG methods are susceptible to non-semantic variations in the data and improving the performance on these benchmark datasets while important may not be enough to achieve the goal of DG.
However, while data augmentation and style diversification methods have been shown to be effective at improving the performance of DG methods on potential variations, it is not practical to train classifiers to be robust to all possible variations.
Due to this limitation, improving the test time methods which either adapt the classifier to unseen domains or abstain from making predictions such as TT-NSS by explicitly transforming the test sample into known styles are essential for DG.

\begin{figure*}[tb]
  \centering{
  \subfigure[Original  style]{\includegraphics[width=0.22\textwidth]{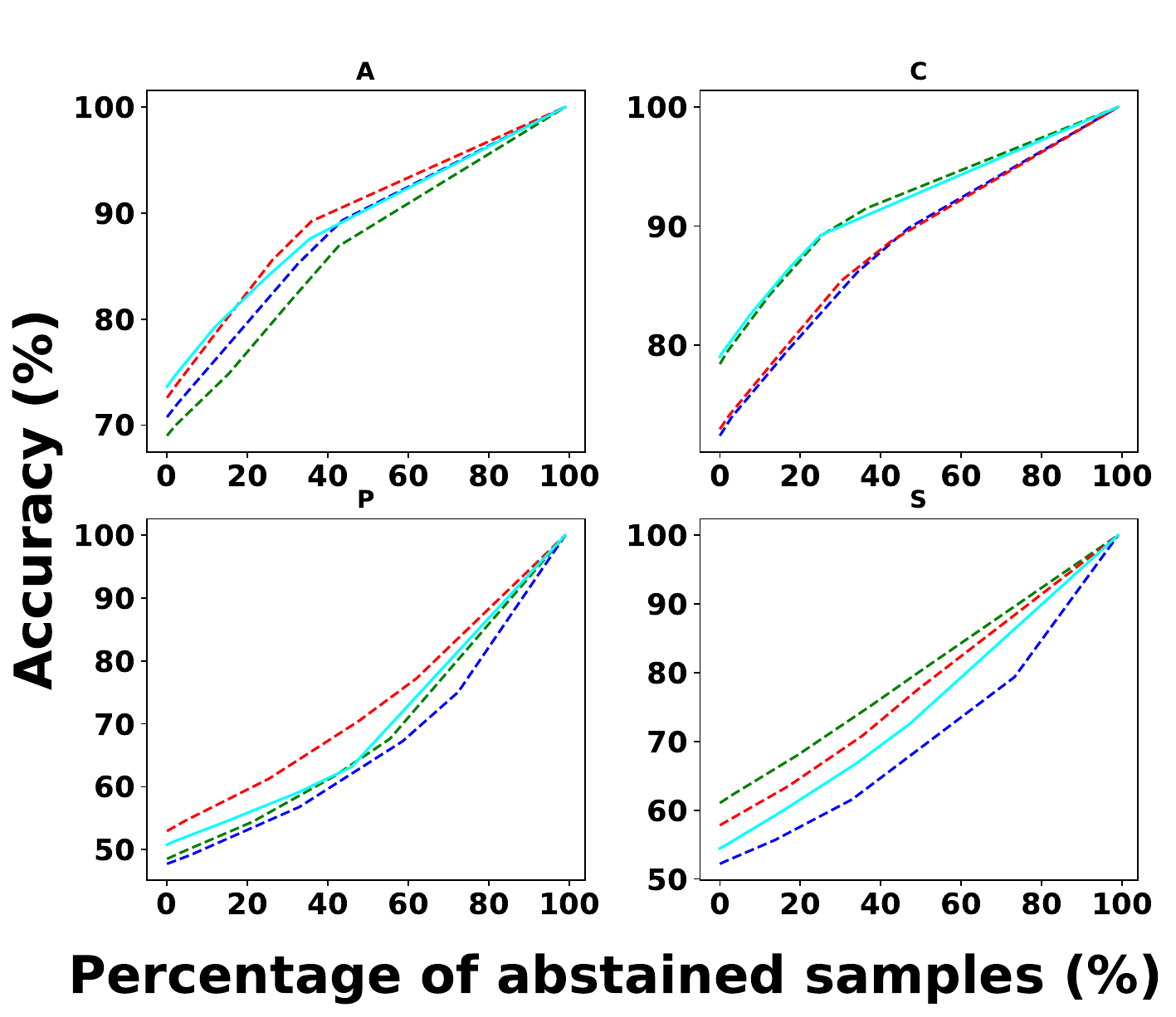}}
  \subfigure[Wikiart  style]{\includegraphics[width=0.22\textwidth]{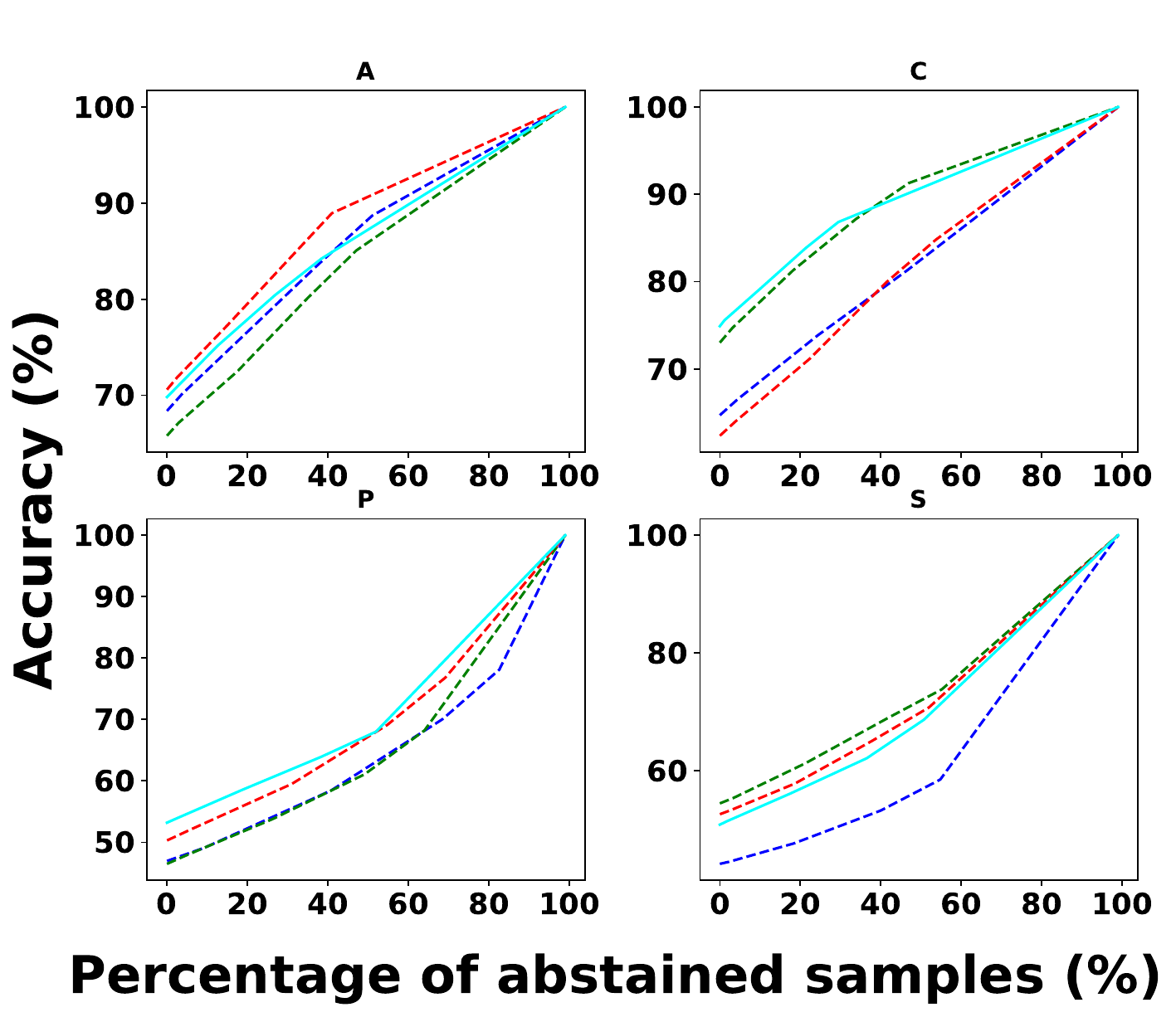}}
  \subfigure[Severity 3 corruptions]{\includegraphics[width=0.22\textwidth]{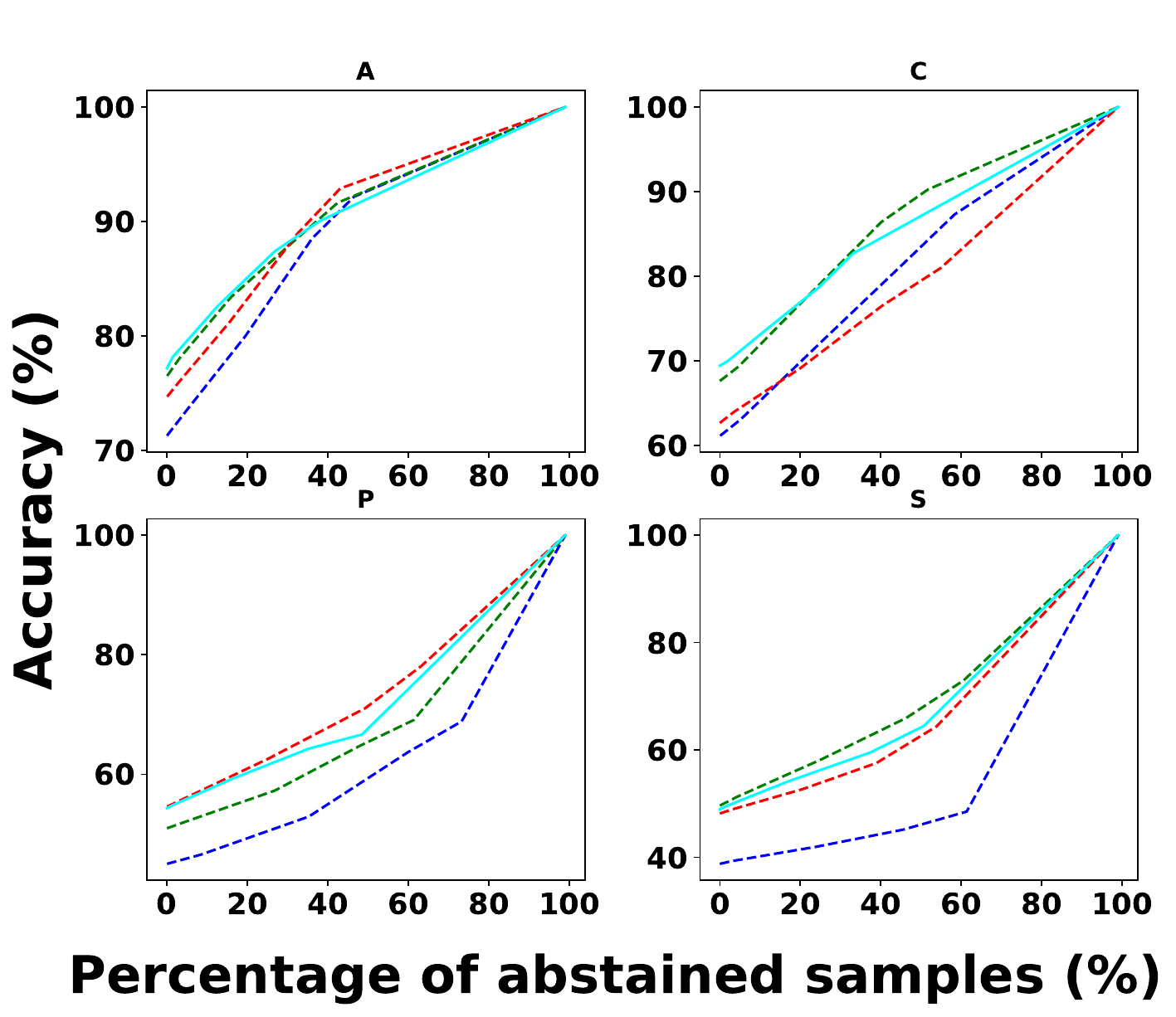}}
  \subfigure[Severity 5 corruptions]{\includegraphics[width=0.22\textwidth]{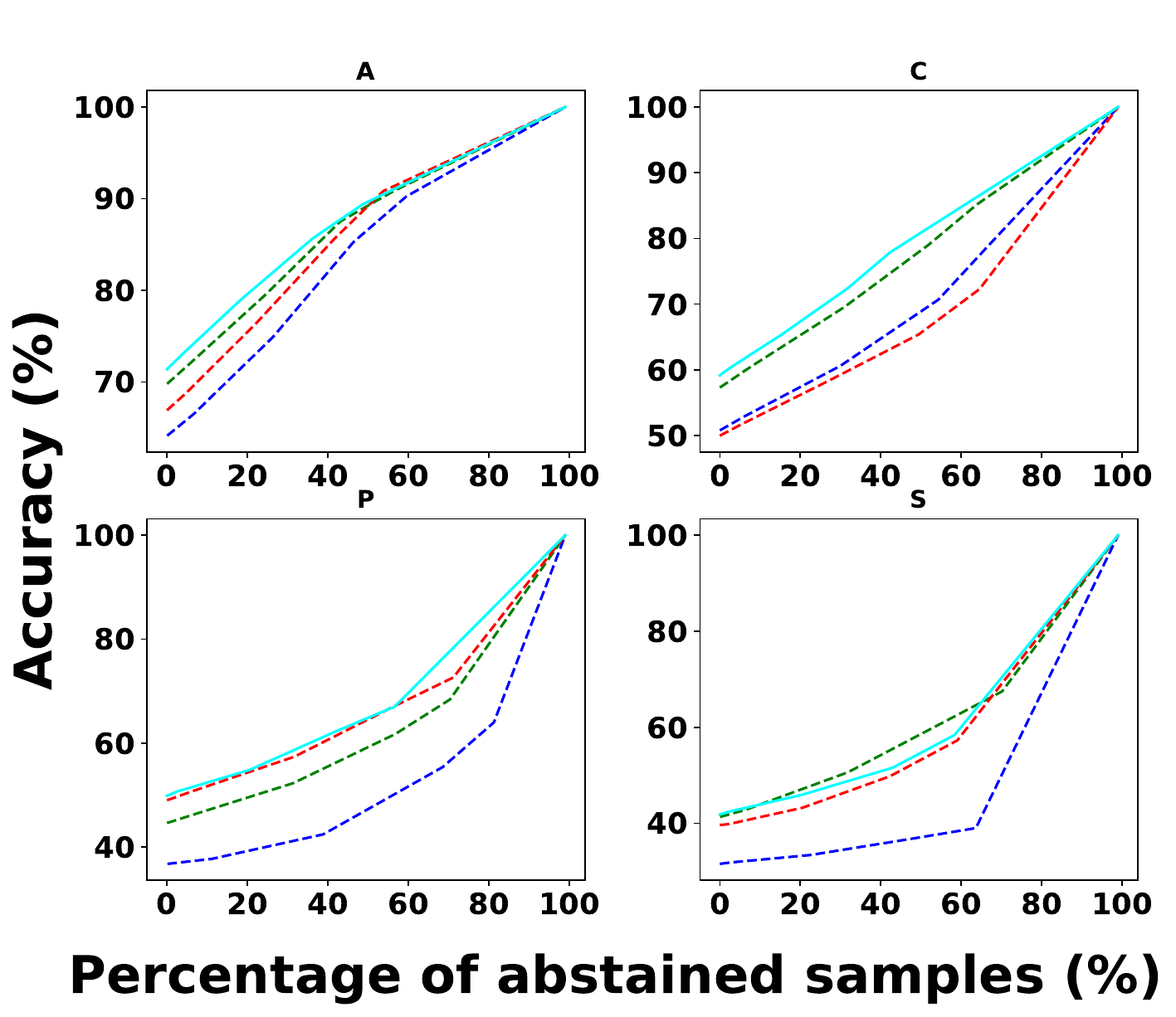}}
  \includegraphics[width=0.25\textwidth]{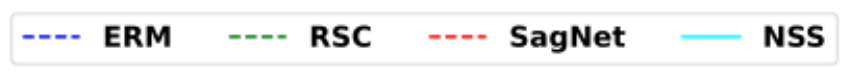}
  }
  \caption{
    Effectiveness of using NSS (with ERM) (solid lines) at producing better risk-averse predictions when evaluated with TT-NSS in comparison to that of other DG methods (dashed lines) in a {\bf single} domain setup.
    NSS-trained classifiers achieve significantly better accuracy on non-abstained samples compared to classifiers trained with ERM and achieve competitive performance to classifiers trained with RSC and SagNet at different abstaining rates on variants of the {\bf PACS} dataset. 
    (See Fig.~\ref{fig:inference_advantage_sconf_vs_conf_pacs} for the explanation of setting.)
   } 
  \label{fig:inference_advantage_NSS_vs_others_sconf_pacs}
\end{figure*}

\subsection{Effectiveness of NSS at improving risk-averse predictions from DG classifiers}
\label{sec:nss_risk_averse}
Here we demonstrate the advantage of using the NSS training procedure for improving the reliability of the classifier's predictions. 
Specifically, we use the NSS losses with that of the ERM-based DG method and minimize the misclassification loss on source domain samples along with minimizing the style misclassification and style consistency losses. 
For training NSS with ERM we used four randomly sampled style images to compute the style smoothed losses in our experiments since we did not observe any significant performance difference with using more images. 
The use of a small number of style-transformed images during NSS training allows us to train DG classifiers without significantly increasing the computational cost compared to that of training with ERM.
The stylized images were generated by using the AdaIN-based decoder pre-trained using data from MS-COCO \cite{lin2014microsoft} as content and Wikiart as style. 
Our results in Table~\ref{table:erm_vs_nss_auc} and Table~\ref{table:erm_vs_nss_auc_M} (in the Appendix) show that classifiers trained with NSS achieve a significantly better area under curve compared to classifiers trained with ERM on PACS, VLCS and OfficeHome datasets in both single and multiple source domain settings.
The improvements in AUC become more evident on variations of these datasets generated by changing to Wikiart style or using common corruptions.
This boost in the AUC is attributed to the style randomization and consistency losses used during NSS training that acts as regularizers and prevents the classifiers from overfitting to specific image styles. 

Results in Fig.~\ref{fig:inference_advantage_NSS_vs_others_sconf_pacs} and Figs.~\ref{fig:inference_advantage_NSS_vs_others_sconf_pacs_M},~\ref{fig:inference_advantage_NSS_vs_others_sconf_vlcs_office},~\ref{fig:inference_advantage_NSS_vs_others_sconf_vlcs_office_M} (in the Appendix) show that classifiers trained with NSS, when evaluated with TT-NSS, achieve better accuracy on non-abstained samples for different abstaining rates and in most cases achieve competitive performance with classifiers trained with RSC and SagNet. 
While in our work we used NSS with ERM, it can be combined with any other DG method such as RSC or SagNet to improve their accuracy on non-abstained samples at different abstaining rates. 
Moreover, training the classifiers with NSS improves the performance of the confidence-based abstaining mechanism as shown in Tables~\ref{table:erm_vs_nss_auc_conf} and~\ref{table:erm_vs_nss_auc_conf_M} (in the Appendix) but even then TT-NSS remains superior in case of severe shifts (such as severity 5 corruptions).







\begin{figure}[tb]
  \centering{
  \subfigure[ERM (SD)]{\includegraphics[width=0.11\textwidth]{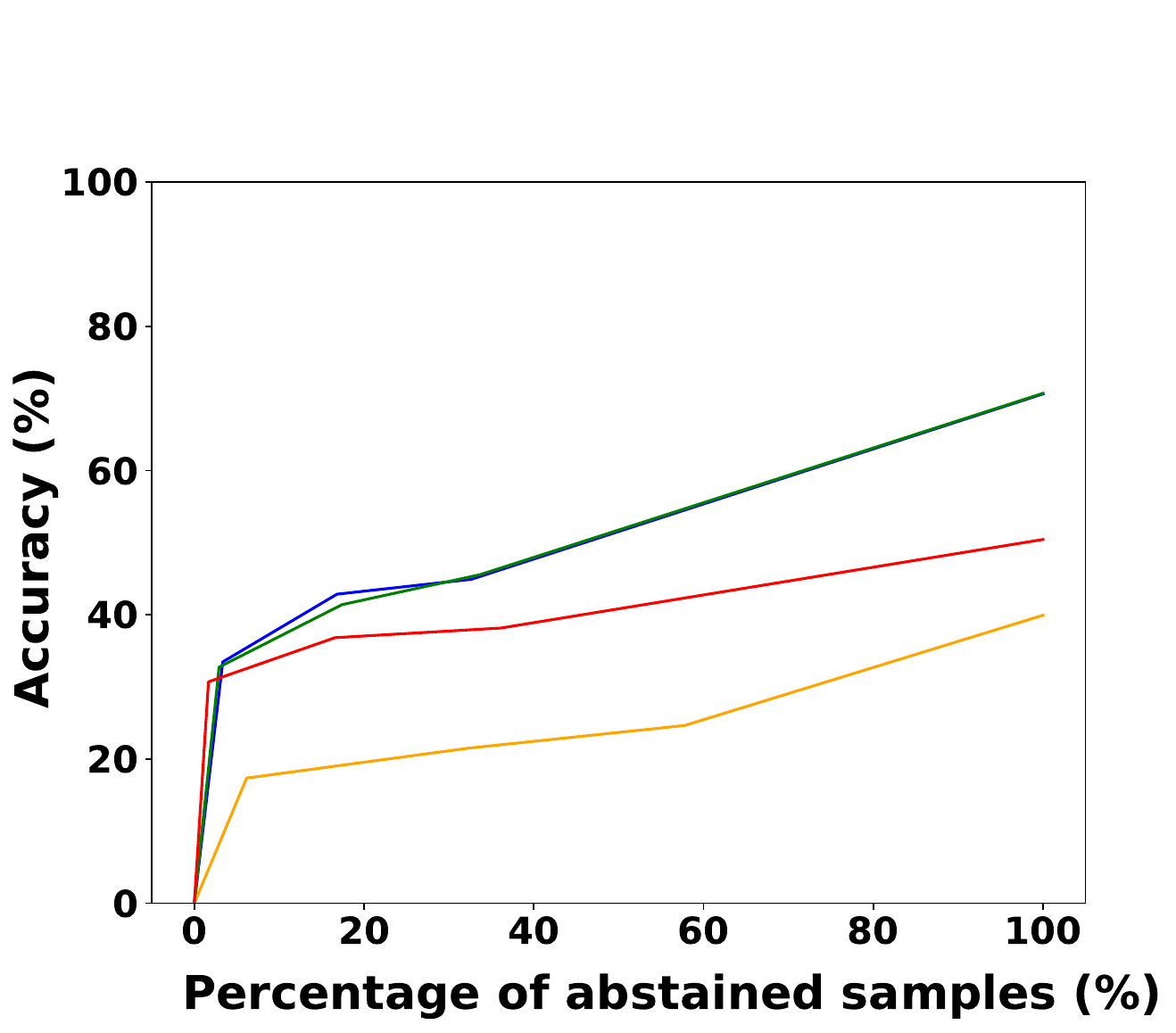}}
  \subfigure[NSS (SD)]{\includegraphics[width=0.11\textwidth]{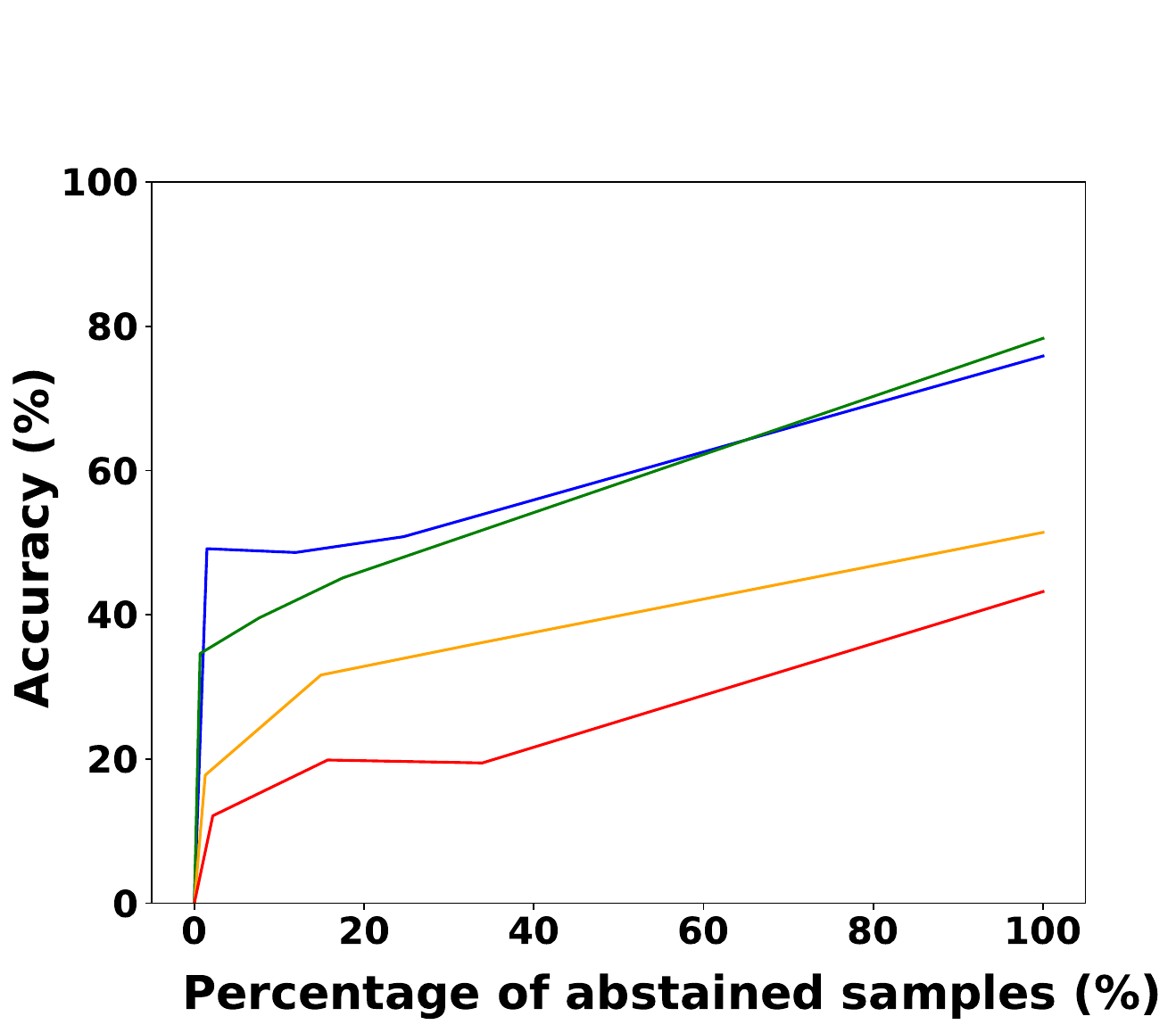}}
  \subfigure[ERM (MD)]{\includegraphics[width=0.11\textwidth]{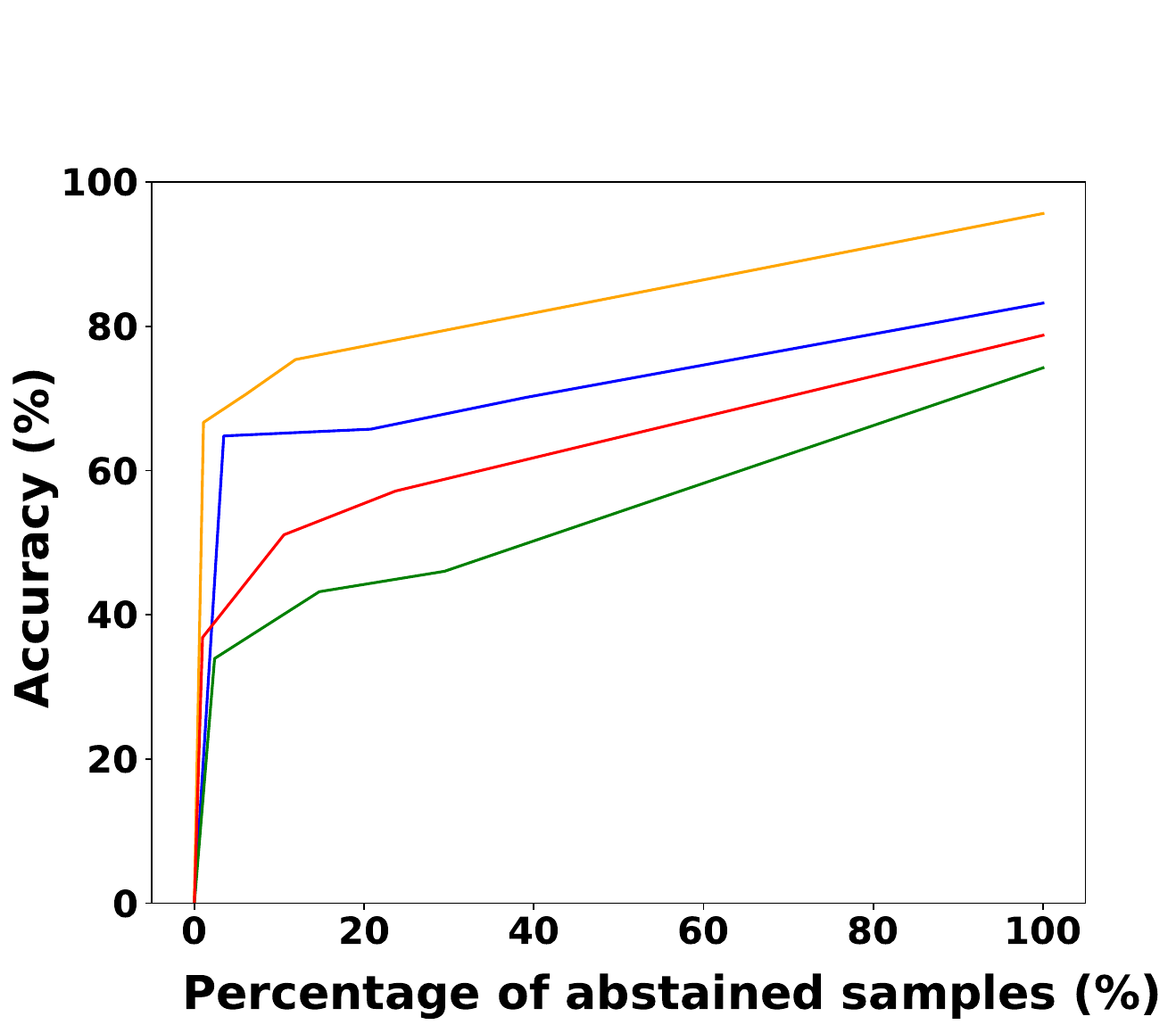}}
  \subfigure[NSS (MD)]{\includegraphics[width=0.11\textwidth]{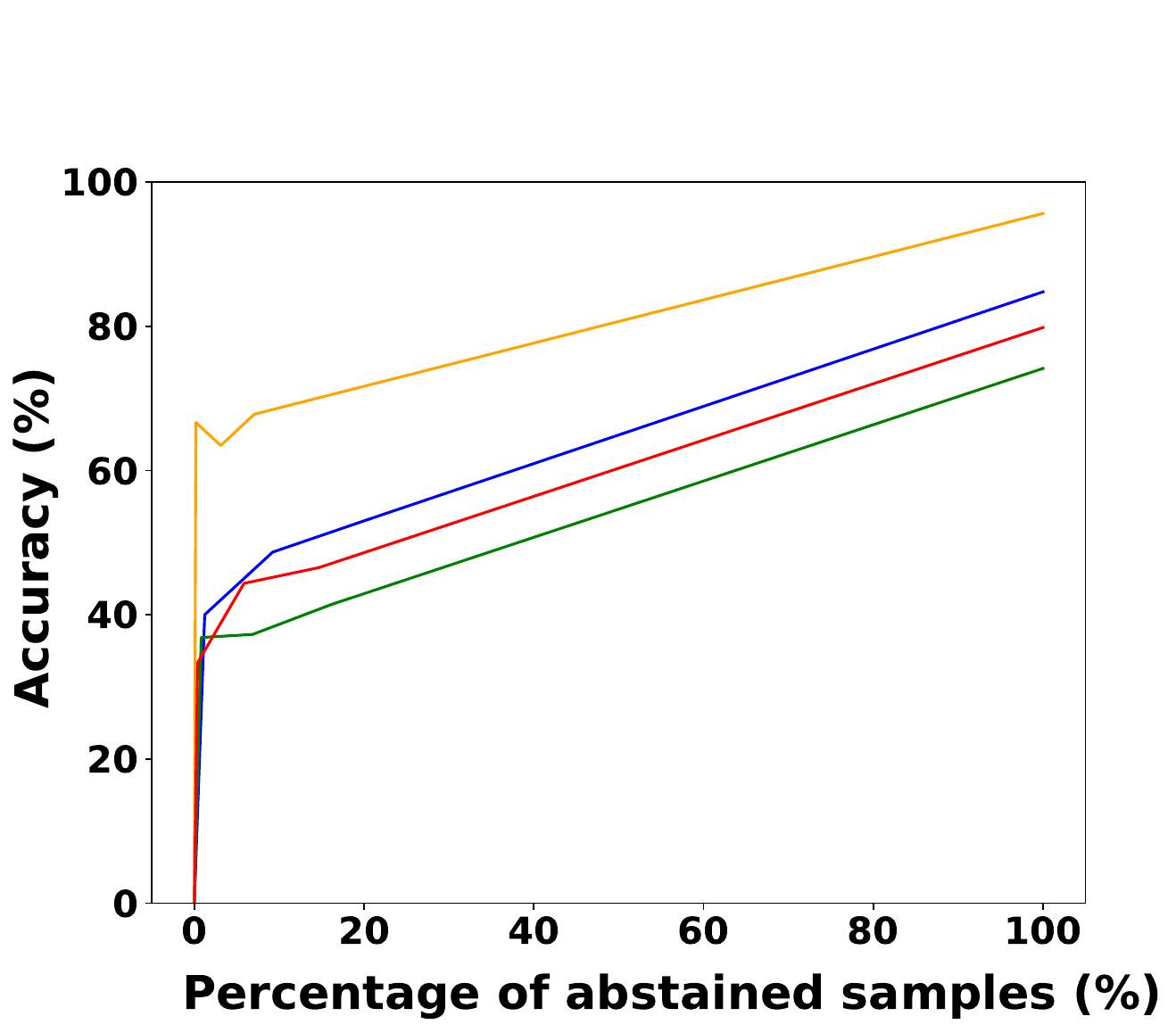}}
  \includegraphics[width=0.25\textwidth]{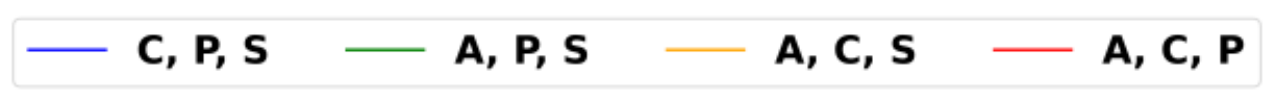}
  \includegraphics[width=0.17\textwidth]{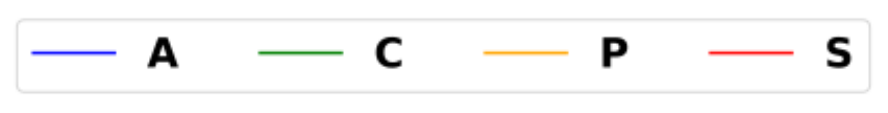}
  }
  \caption{
  Accuracy on samples abstained from a prediction by TT-NSS in single (SD) (a, b) and multiple (MD) (c,d) domain settings on the PACS dataset. (Test domains are denoted in the legend.)
   } 
  \label{fig:accuracy_on_abstained}
\end{figure}

\subsection{Predictions on abstained samples}
Here we evaluate the effectiveness of TT-NSS in correctly abstaining on samples that could lead to misclassifications. We show this by showing the accuracy of the DG classifier on the test samples that were abstained.
Results in Fig.~\ref{fig:accuracy_on_abstained} show that using a small value of the threshold $\alpha$ where TT-NSS abstains on few samples, the accuracy on abstained samples is significantly lower for classifiers trained with ERM and NSS in both single and multiple source domain settings on the PACS dataset (original style). 
This is in  comparison to the standard accuracy of the classifier (recovered at 100\% abstaining rate). 
The low accuracy on abstained samples suggests that TT-NSS correctly refrains from making predictions on ambiguous samples. 
Moreover, the accuracy on abstained samples decreases for most test domains for classifiers trained with NSS compared to classifiers trained with ERM, suggesting that NSS improves the ability of TT-NSS to identify risky samples.

\begin{figure}[tb]
  \centering{
  \subfigure[ERM (SD)]{\includegraphics[width=0.11\textwidth]{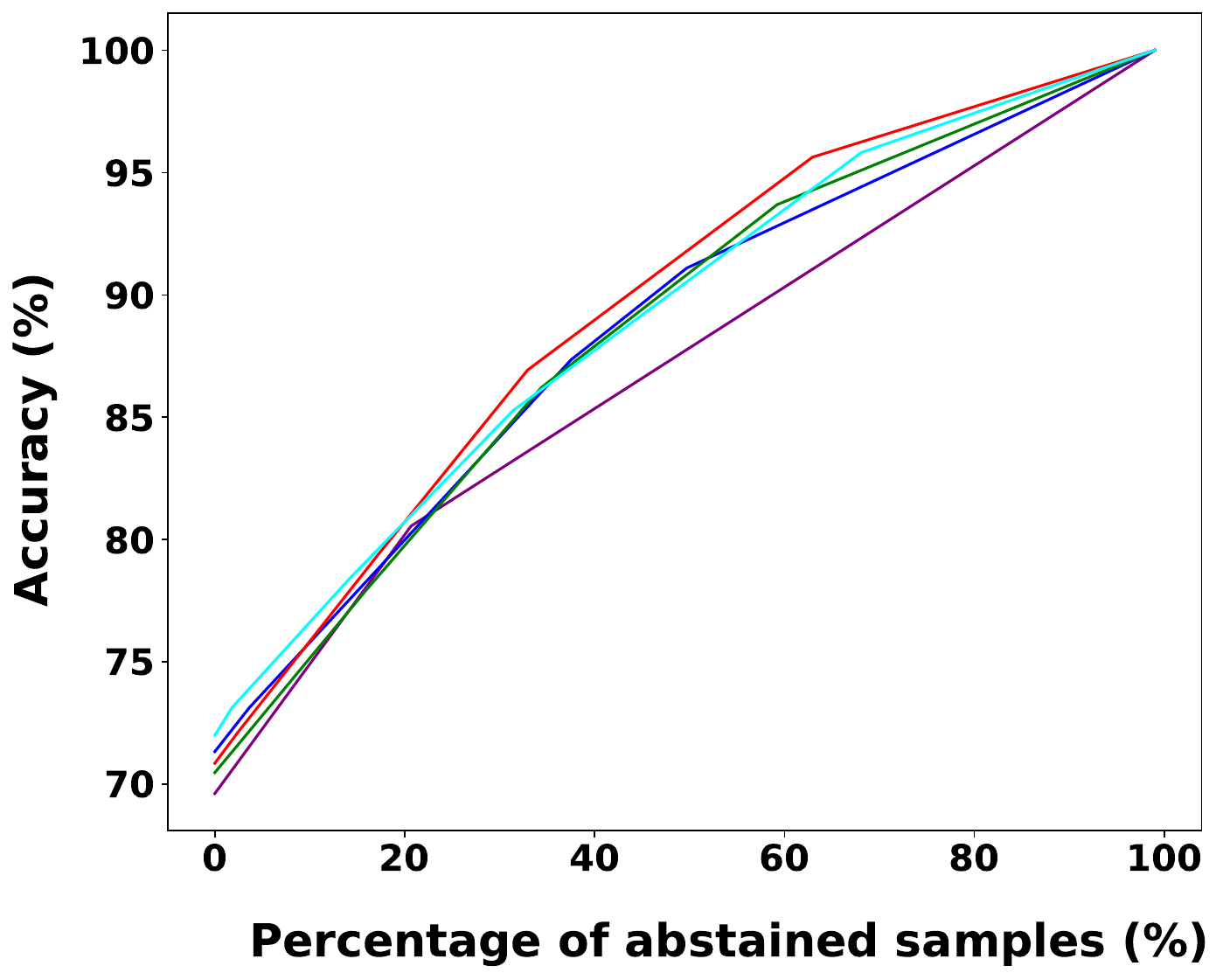}}
  \subfigure[NSS (SD)]{\includegraphics[width=0.11\textwidth]{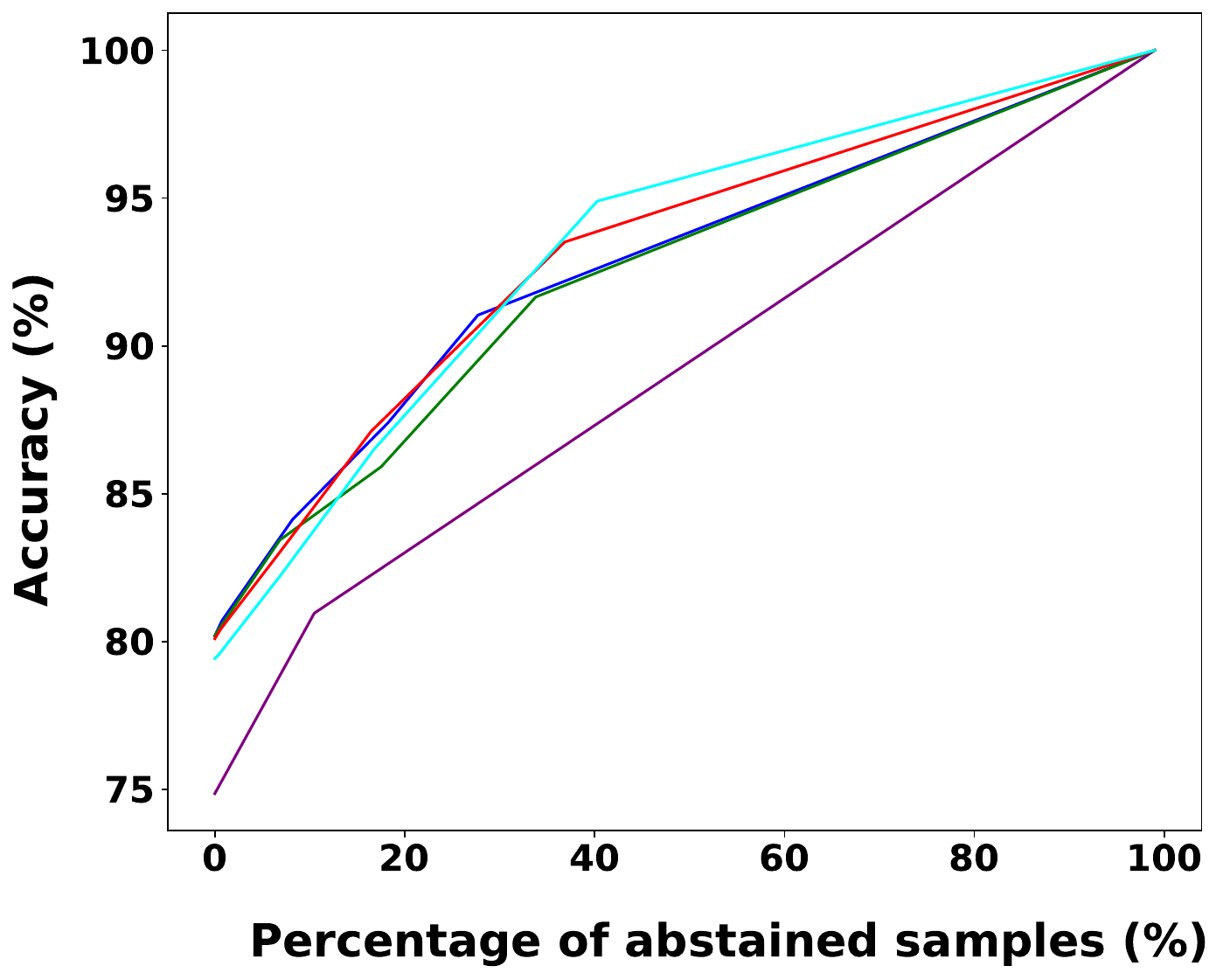}}
  \subfigure[ERM (MD)]{\includegraphics[width=0.11\textwidth]{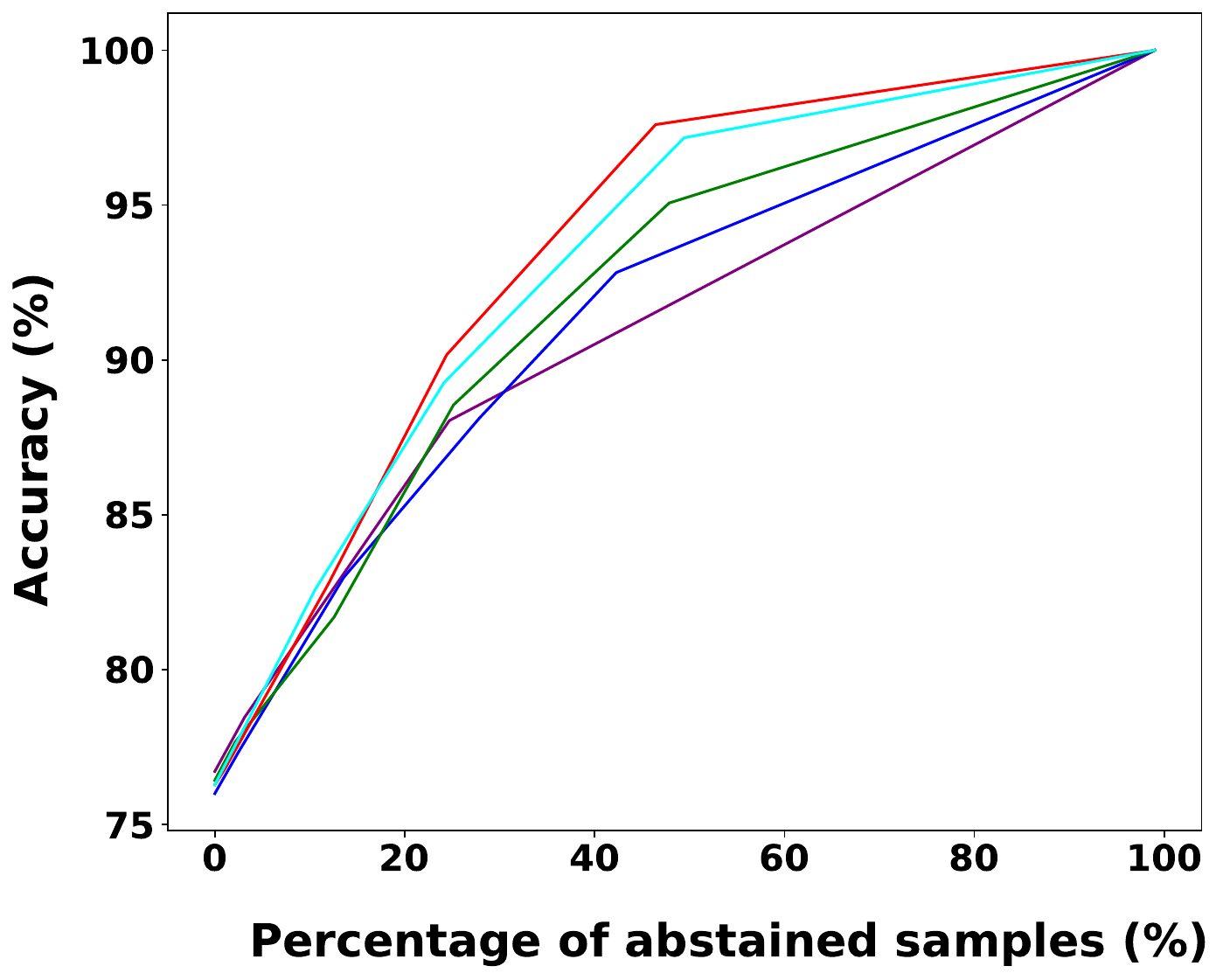}}
  \subfigure[NSS (MD)]{\includegraphics[width=0.11\textwidth]{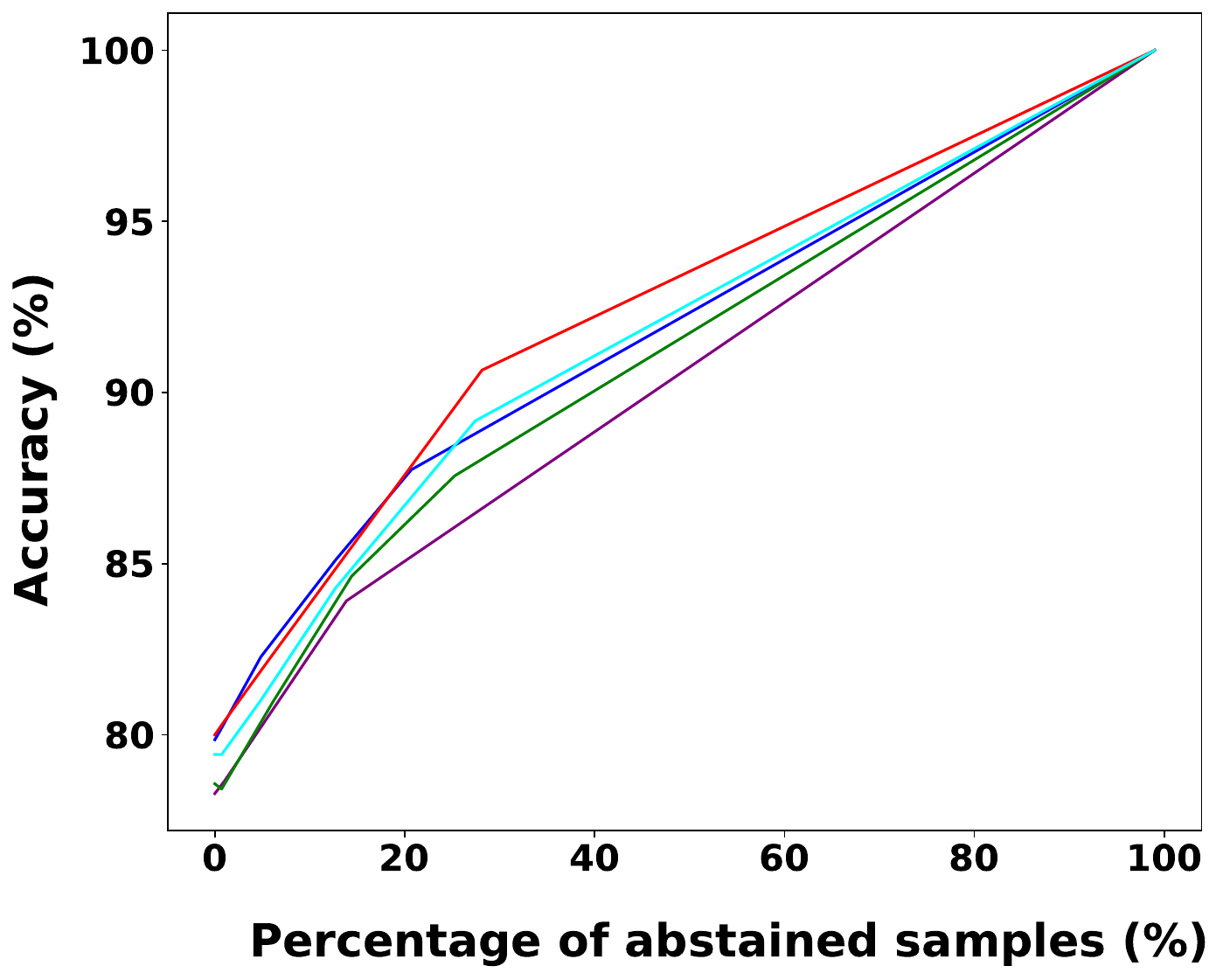}}
  \includegraphics[width=0.2\textwidth]{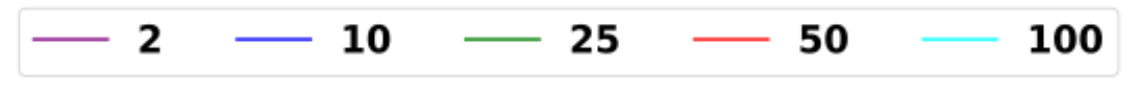}
  \includegraphics[width=0.19\textwidth]{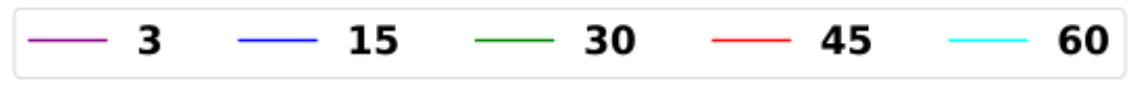}
  }
  \caption{
  The performance of TT-NSS is not significantly affected by the value of $n$ beyond $n=10$ for single (SD) (a, b) and $n=15$ in multiple (MD) (c, d) source domain settings. For the SD setting, the classifier is trained on the Cartoon domain and evaluated on the remaining domains in PACS, and for the MD setting, the classifier is evaluated on the Cartoon domain after training on the rest.
   } 
  \label{fig:effect_of_styles_pacs}
\end{figure}

\subsection{Effect of number of styles}
\label{sec:exp_num_styles}
Here we evaluate the effect of using different numbers of re-stylizations of a single test image, $n$, in TT-NSS using a subsample (see App.~\ref{app:subsample}) of the PACS dataset (original style). 
Results in Fig.~\ref{fig:effect_of_styles_pacs} show that in both single and multi-source domain settings, using a large value of $n$ leads to only a small improvement in the accuracy of non-abstained samples at higher abstaining rates whereas performance at lower abstaining rates remains similar for different values of $n$.
Since using a larger value of $n$ can slow down the inference, we use $n$ as 10 and 15 (5 per domain) in the single and multiple source domain settings. 
Evaluating a single test sample with TT-NSS using 15 styles increases the inference cost by mere 0.26 seconds on our hardware, showing the potential of TT-NSS at producing risk-averse predictions without sacrificing inference efficiency.

\section{Discussion and conclusion}
Our work proposed and demonstrated the effectiveness of incorporating an abstaining mechanism based on NSS to improve the reliability of a DG classifier's predictions on data from unseen domains.
Using advances in neural style transfer, our inference procedure uses the prediction consistency of the classifier on stylized images to predict or abstain on a test sample and requires only black-box access to the DG classifier.
Moreover, we proposed a training procedure to improve the reliability of a  classifier's prediction at different abstaining rates and demonstrated its effectiveness on various datasets and their variations.  
We note that while NSS is effective at gauging the reliability of a classifier's prediction on test samples,
ascertaining the robustness of this prediction to arbitrary style changes is an important open problem and will be the focus of future works. 

\section{Acknowledgment}
This work was supported by the NSF EPSCoR-Louisiana Materials Design Alliance (LAMDA) program \#OIA-1946231 and was performed under the auspices of the U.S. Department of Energy by the Lawrence Livermore National Laboratory under Contract No. DE- AC52-07NA27344 and was supported by the LLNL-LDRD Program under Project No. 23-ERD-030.

{\small
\bibliographystyle{ieee_fullname}
\bibliography{egbib}
}

\clearpage
\appendix
\begin{center}
{\LARGE \bf Appendix}
\end{center}
We present the results of additional experiments in Appendix~\ref{app:additional_experiments} followed by details of datasets used in our work along with implementation details of our algorithm and baselines in Appendix~\ref{app:experimental_details}. 

\section{Additional experiments}
\label{app:additional_experiments}
Here we present the results omitted in the main paper due to space limitations. In App.~\ref{app:comparison_conf_vs_sconf}, we provide additional empirical results for the comparison between TT-NSS and the confidence-based abstaining mechanism on the VLCS dataset and the multi-source domain setting. 
In App.~\ref{app:comparison_nss_vs_others}, we provide additional empirical results demonstrating the effectiveness of models trained with NSS when evaluated with TT-NSS and the confidence-based abstaining mechanism on different datasets in both single and multi-domain settings.

\subsection{Additional results on the comparison of TT-NSS and confidence-based abstaining}
\label{app:comparison_conf_vs_sconf}
Here we present results on the comparison of TT-NSS and confidence-based abstaining using the AUC metric, and present results on the VLCS dataset both in single and multi-domain settings. 
Our results in Tables~\ref{table:conf_vs_sconf_pacs},~\ref{table:conf_vs_sconf_vlcs} and~\ref{table:conf_vs_sconf_pacs_M},~\ref{table:conf_vs_sconf_vlcs_M} show that similar to the results presented in Fig.~\ref{fig:inference_advantage_sconf_vs_conf_pacs} in the main paper, the AUC for the accuracy versus the percentage of abstained samples curve is significantly better for TT-NSS compared to confidence-based abstaining in the single domain setting and is competitive on the multi-domain setting. 
The advantage of TT-NSS becomes clear when evaluated on data from Wikiart and corrupted domains. 
This advantage of TT-NSS holds regardless of the training method used for training the DG classifier or the dataset used.

\begin{figure*}[tb]
  \centering{
    \subfigure[Original  style]{\includegraphics[width=0.2\textwidth]{images/sconf_conf_SD_original_style_pacs.pdf}}
  \subfigure[Wikiart  style]{\includegraphics[width=0.2\textwidth]{images/sconf_conf_SD_wikiart_style_pacs.pdf}}
  \subfigure[Severity 3 corruptions]{\includegraphics[width=0.2\textwidth]{images/sconf_conf_SD_severity_3_style_pacs.pdf}}
  \subfigure[Severity 5 corruptions]{\includegraphics[width=0.2\textwidth]{images/sconf_conf_SD_severity_5_style_pacs.pdf}}
  
  \subfigure[Original  style]{\includegraphics[width=0.2\textwidth]{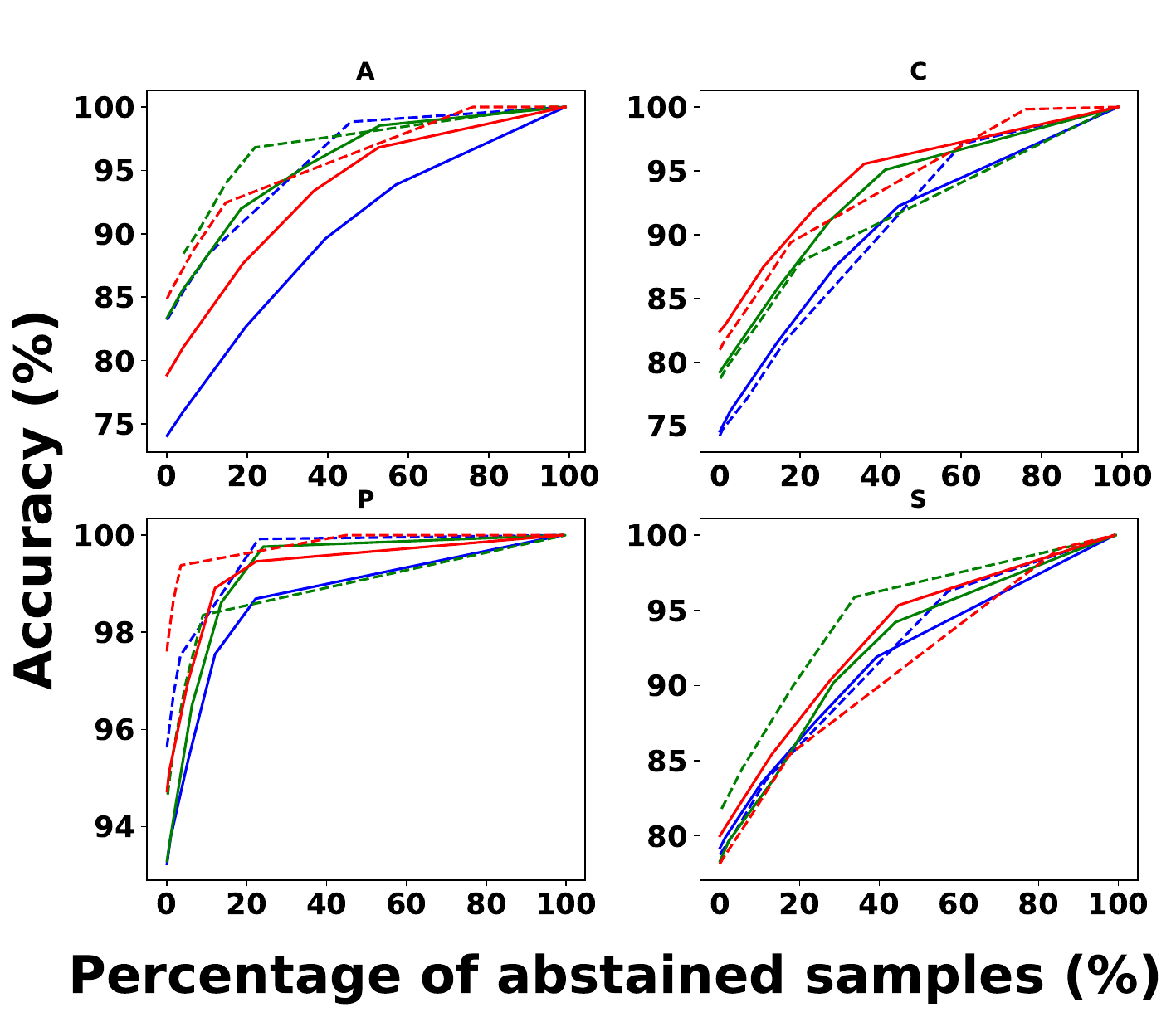}}
  \subfigure[Wikiart  style]{\includegraphics[width=0.2\textwidth]{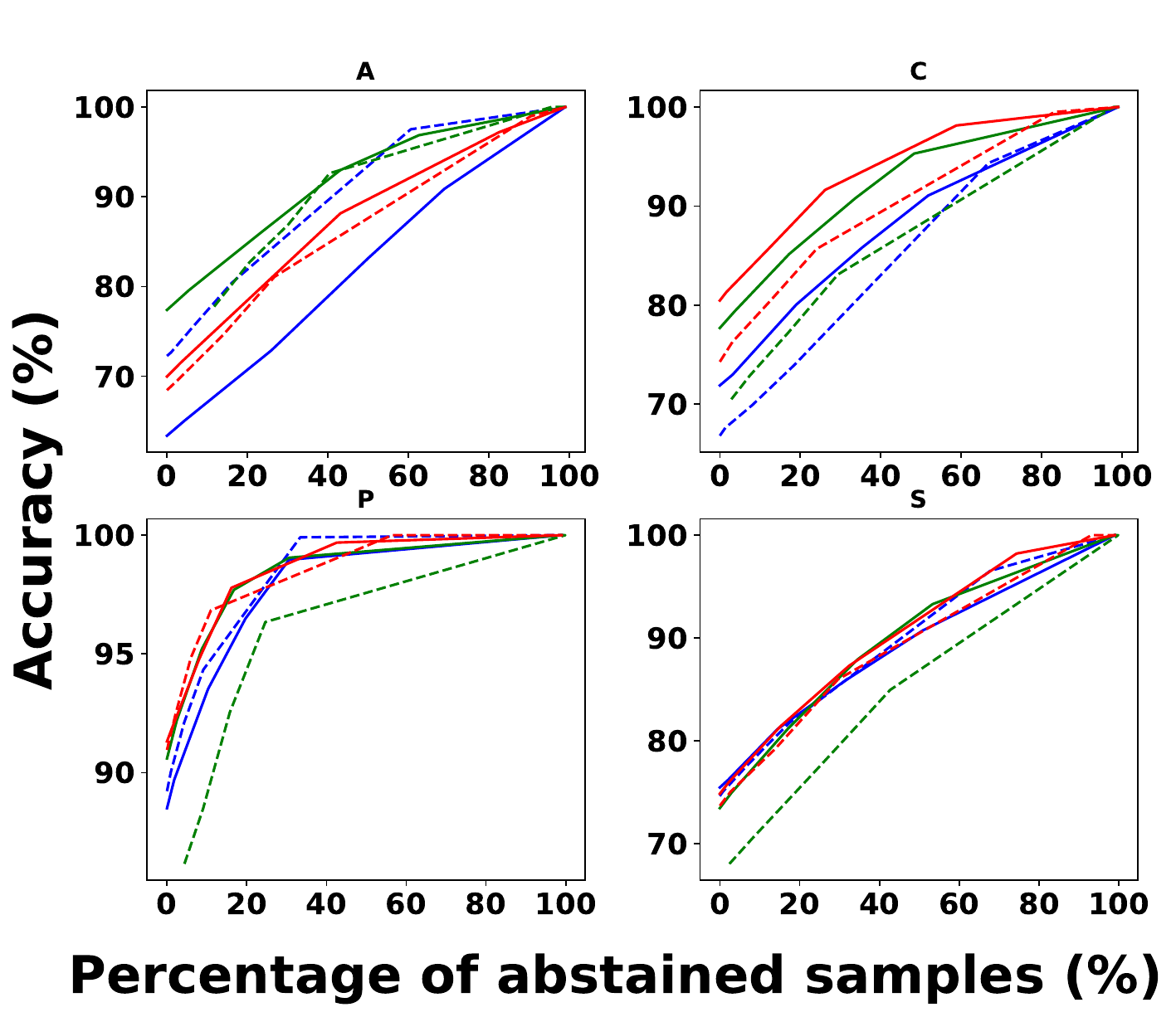}}
  \subfigure[Severity 3 corruptions]{\includegraphics[width=0.2\textwidth]{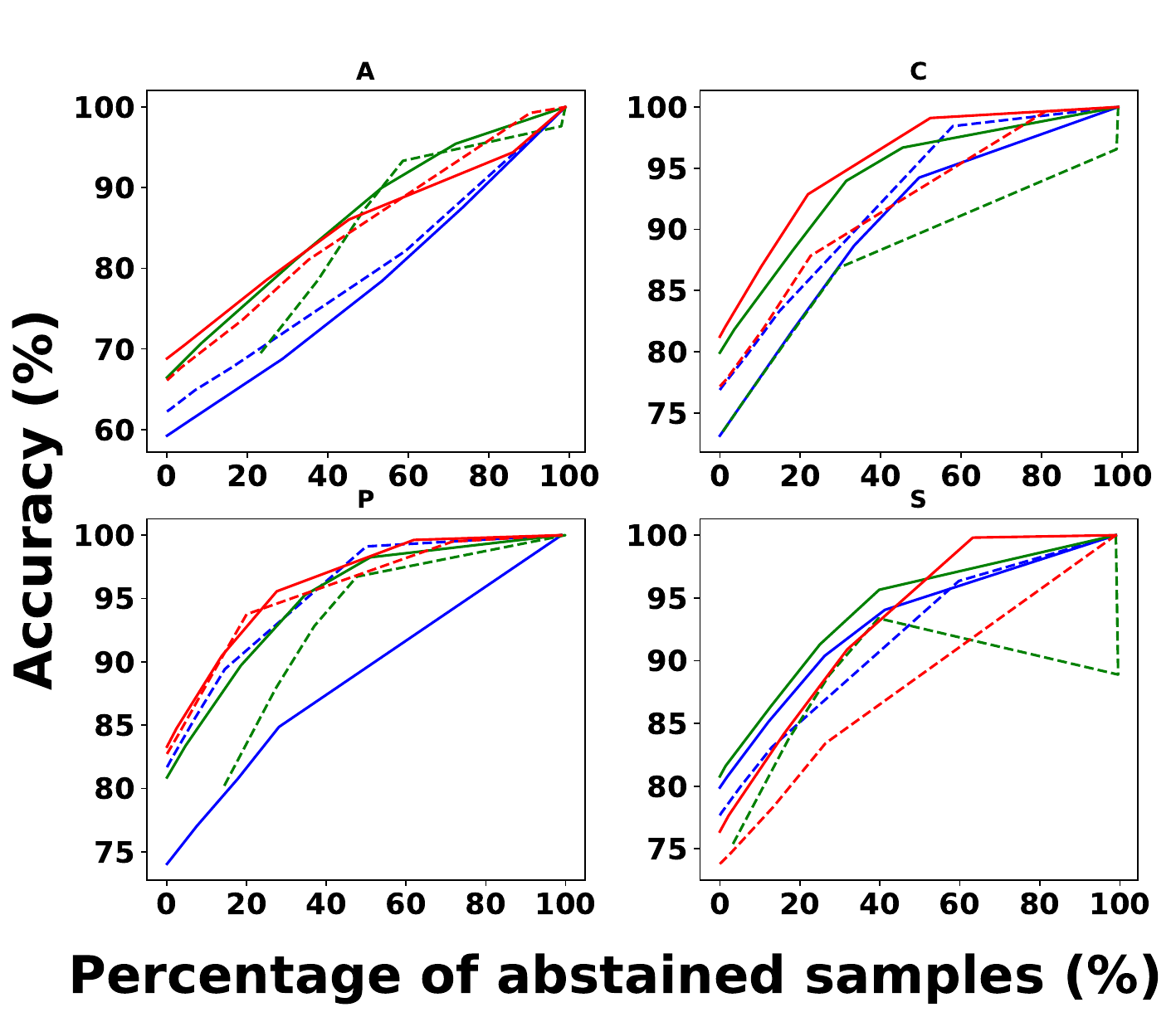}}
  \subfigure[Severity 5 corruptions]{\includegraphics[width=0.2\textwidth]{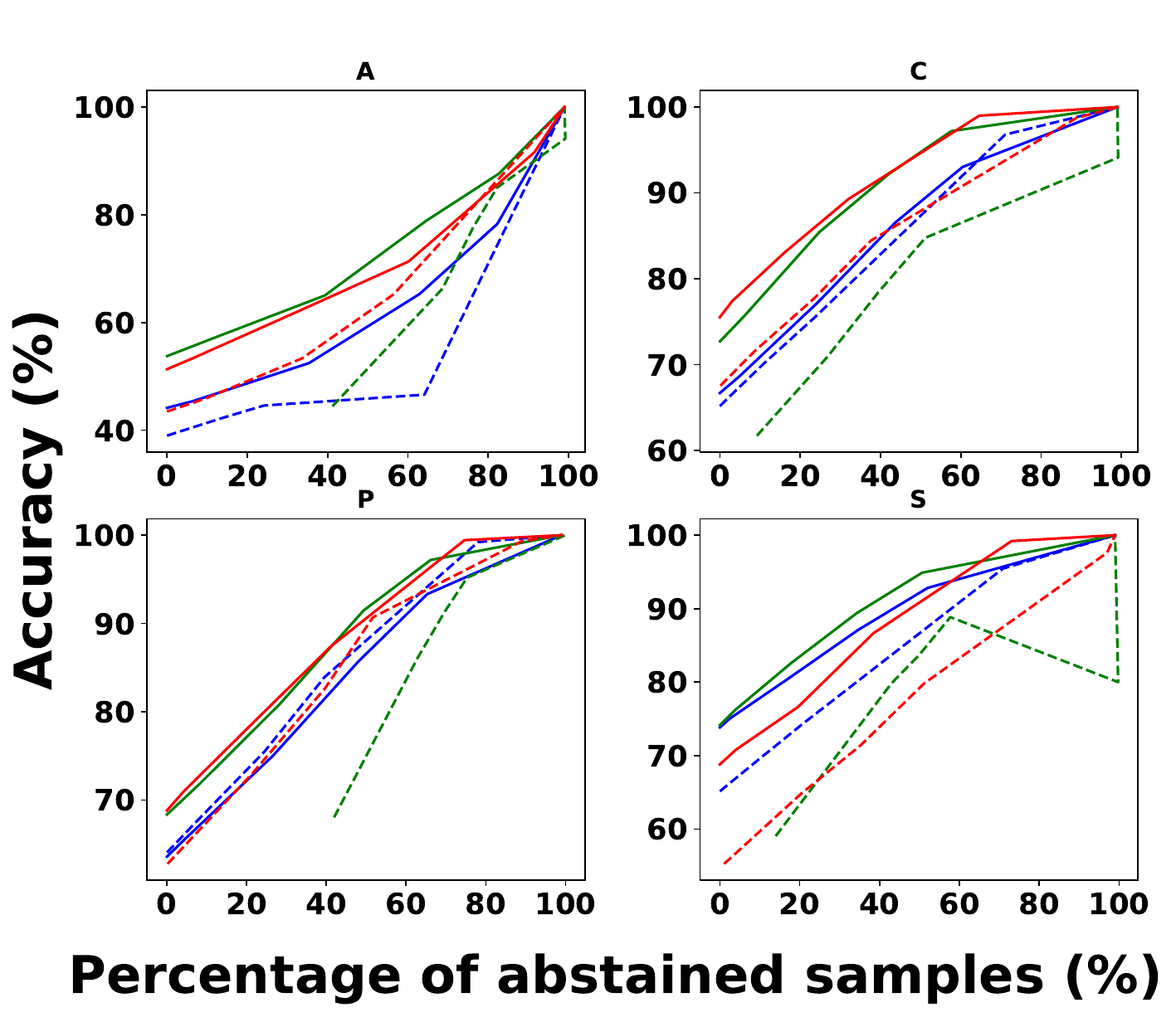}}
  
  \includegraphics[width=0.25\textwidth]{images/sconf_conf_legend.pdf}
  }
  \caption{
    Comparison of TT-NSS (solid lines) and confidence-based method (dashed lines) in a {\bf single} (top row) and {\bf multi}-source (bottom row) domain setup on classifiers trained with ERM. 
    The graphs show accuracy vs abstained points on different variants of the {\bf PACS} dataset ((a) original, (b) wikiart, (c,d) corrupted), and different source/target domains.
    In most domains, the accuracy of the TT-NSS (solid line) is similar to or better than the corresponding accuracy of the confidence-based method (dashed line) for most of the range of the percentage of abstained samples. 
    (Note: The source domain from PACS used for training is denoted in the title and the target domain used for evaluation is denoted in the title in the bottom row.)
   } 
  \label{fig:inference_advantage_sconf_vs_conf_pacs_M}
\end{figure*}

\begin{figure*}[tb]
  \centering{
  \subfigure[Original  style]{\includegraphics[width=0.2\textwidth]{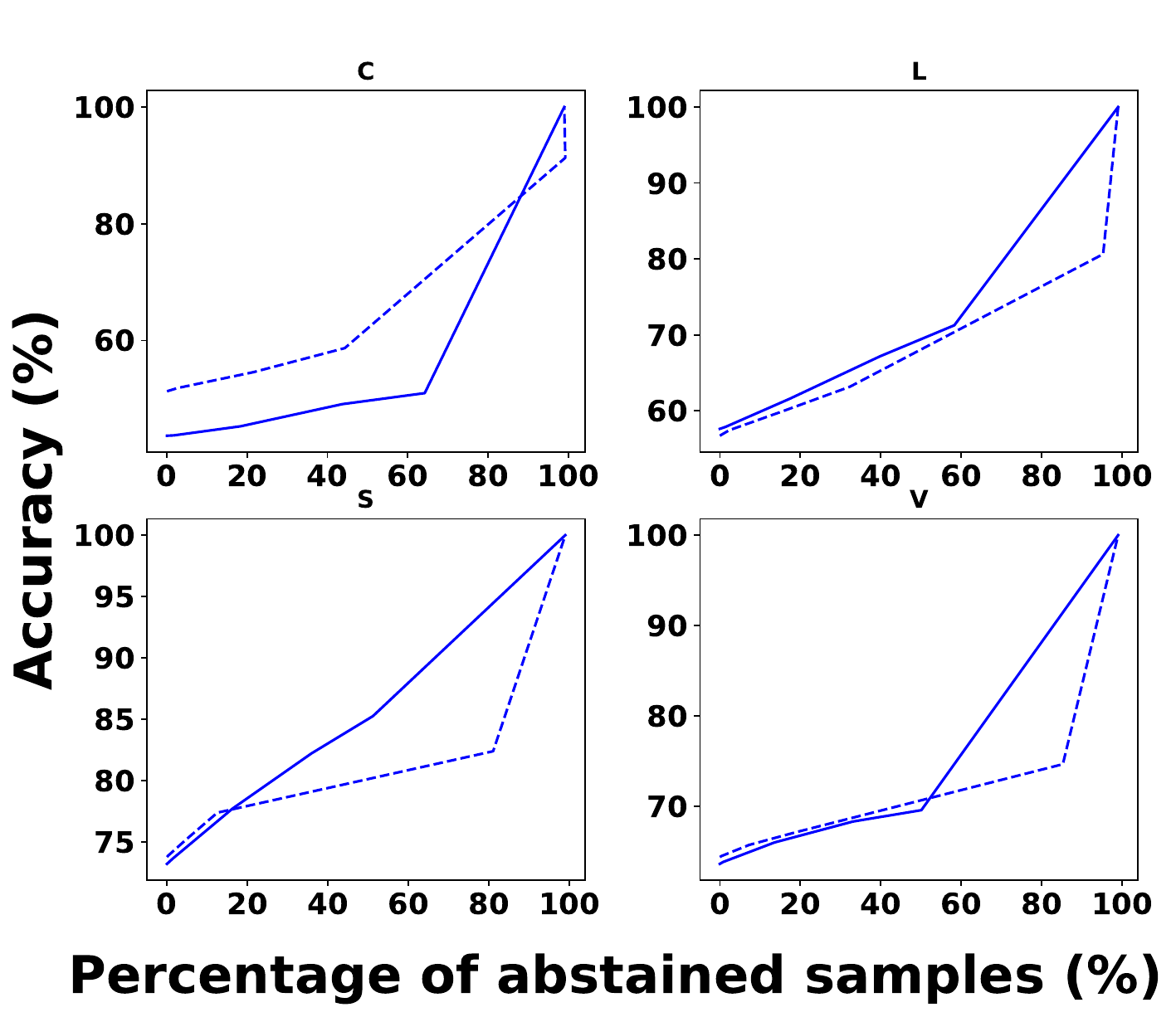}}
  \subfigure[Wikiart  style]{\includegraphics[width=0.2\textwidth]{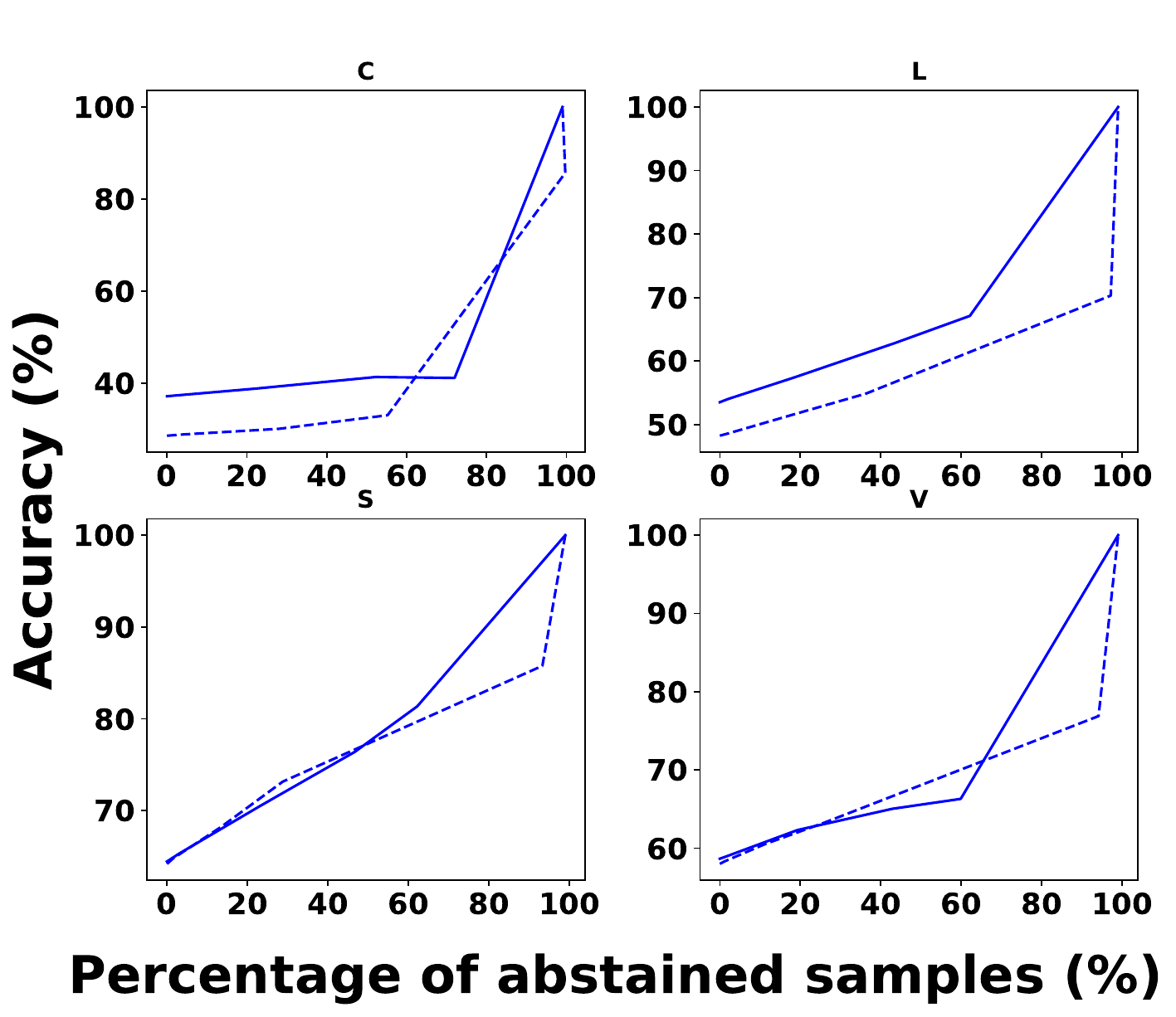}}
  \subfigure[Severity 3 corruptions]{\includegraphics[width=0.2\textwidth]{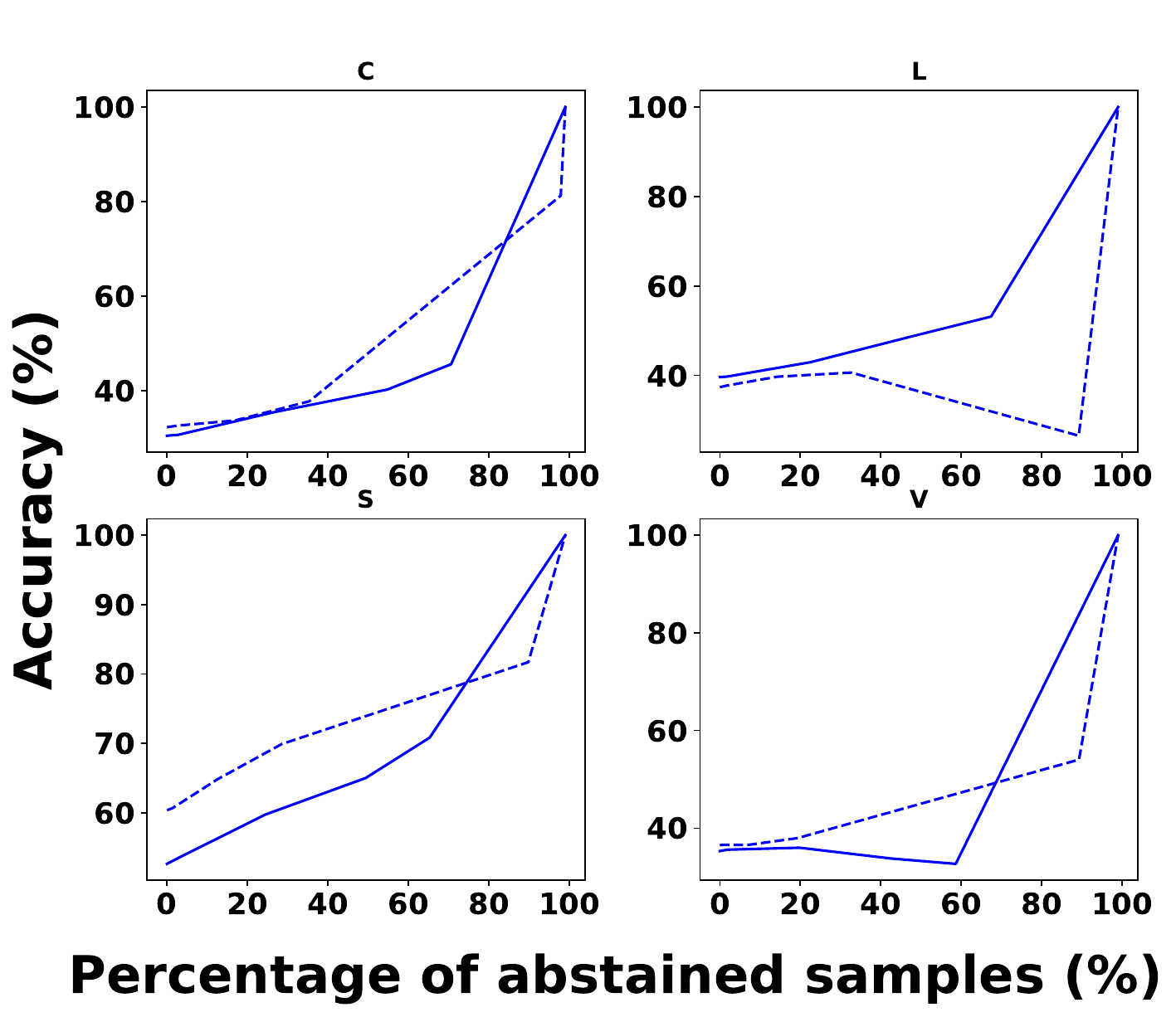}}
  \subfigure[Severity 5 corruptions]{\includegraphics[width=0.2\textwidth]{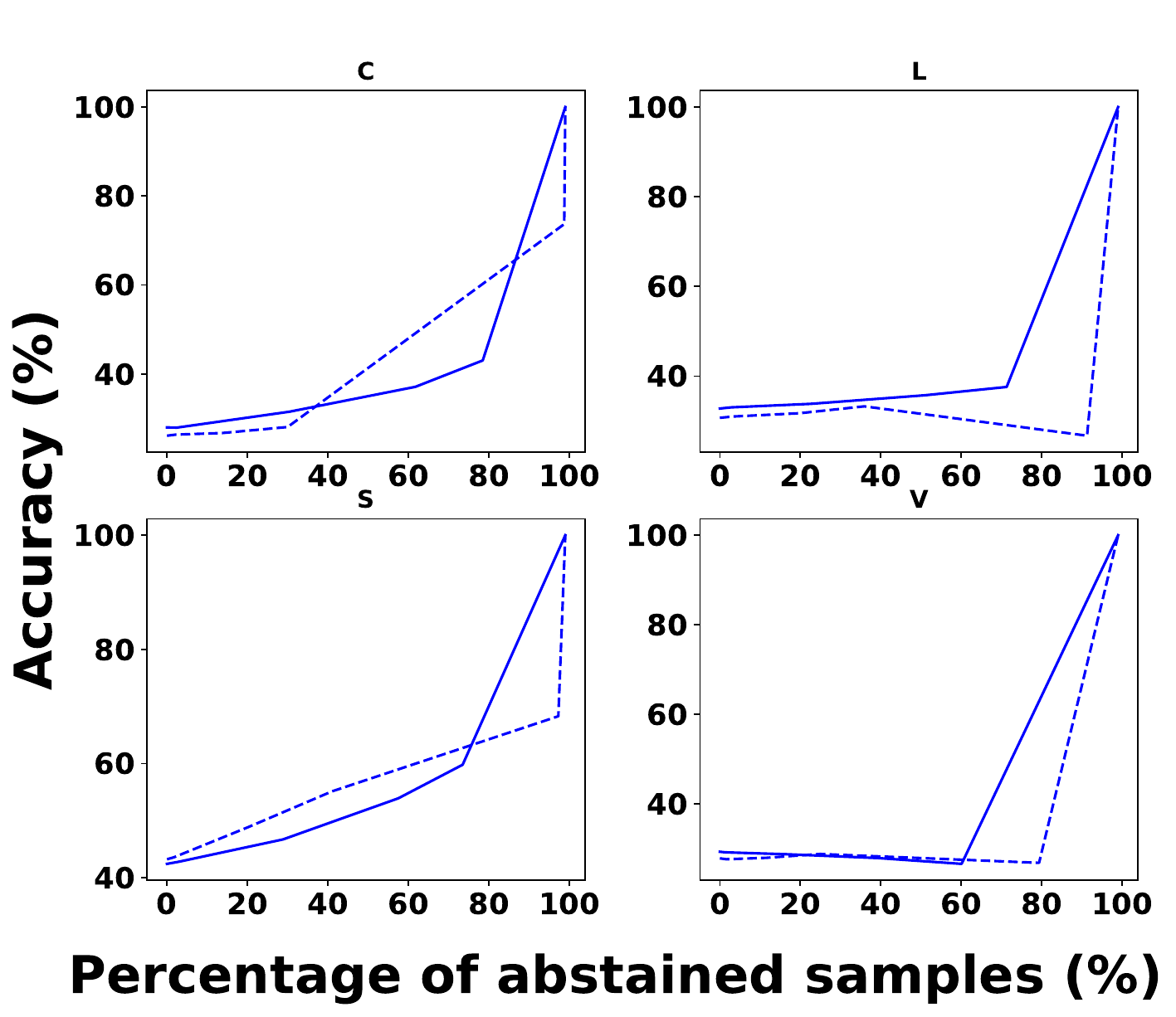}}

  \subfigure[Original  style]{\includegraphics[width=0.2\textwidth]{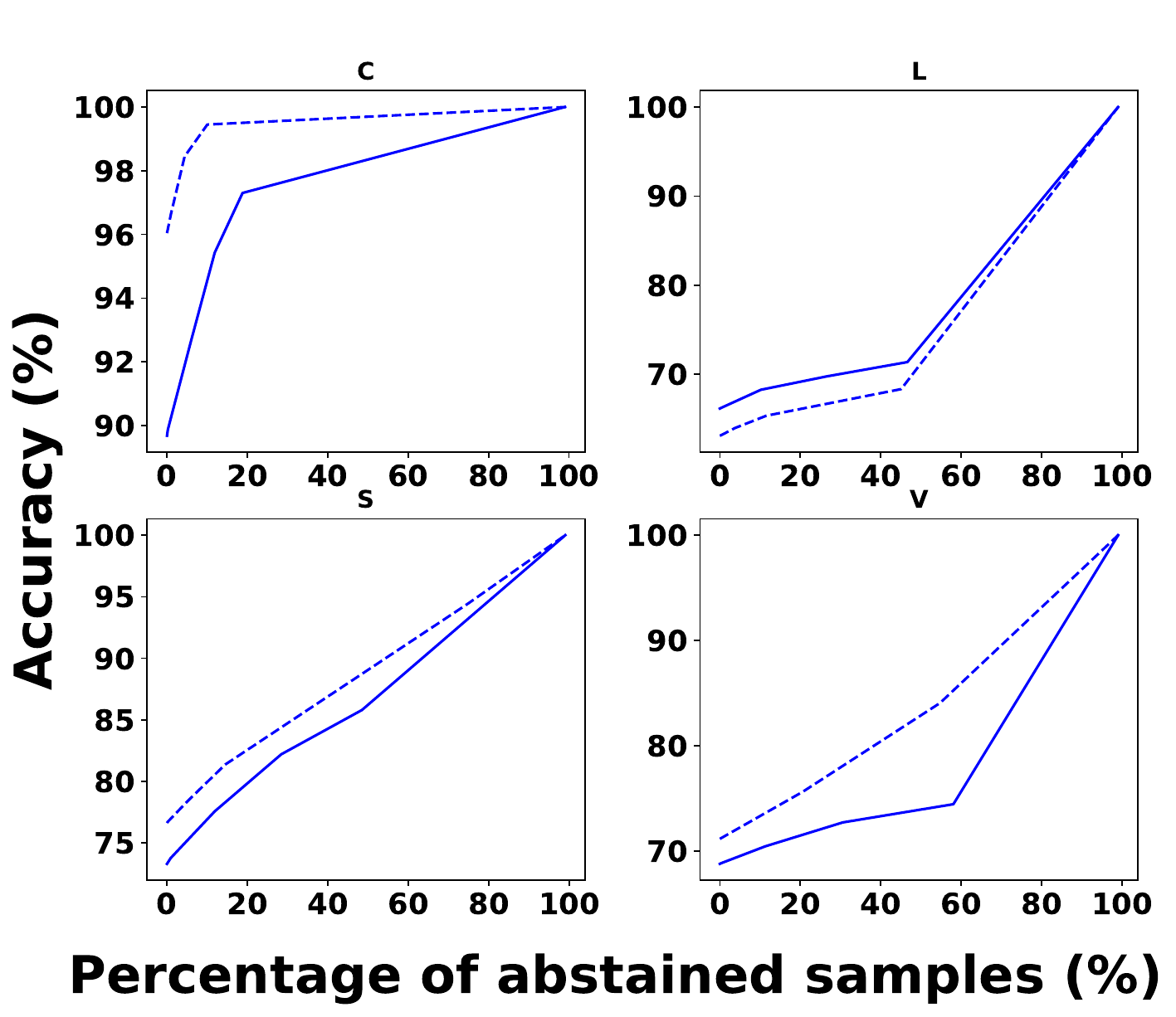}}
  \subfigure[Wikiart  style]{\includegraphics[width=0.2\textwidth]{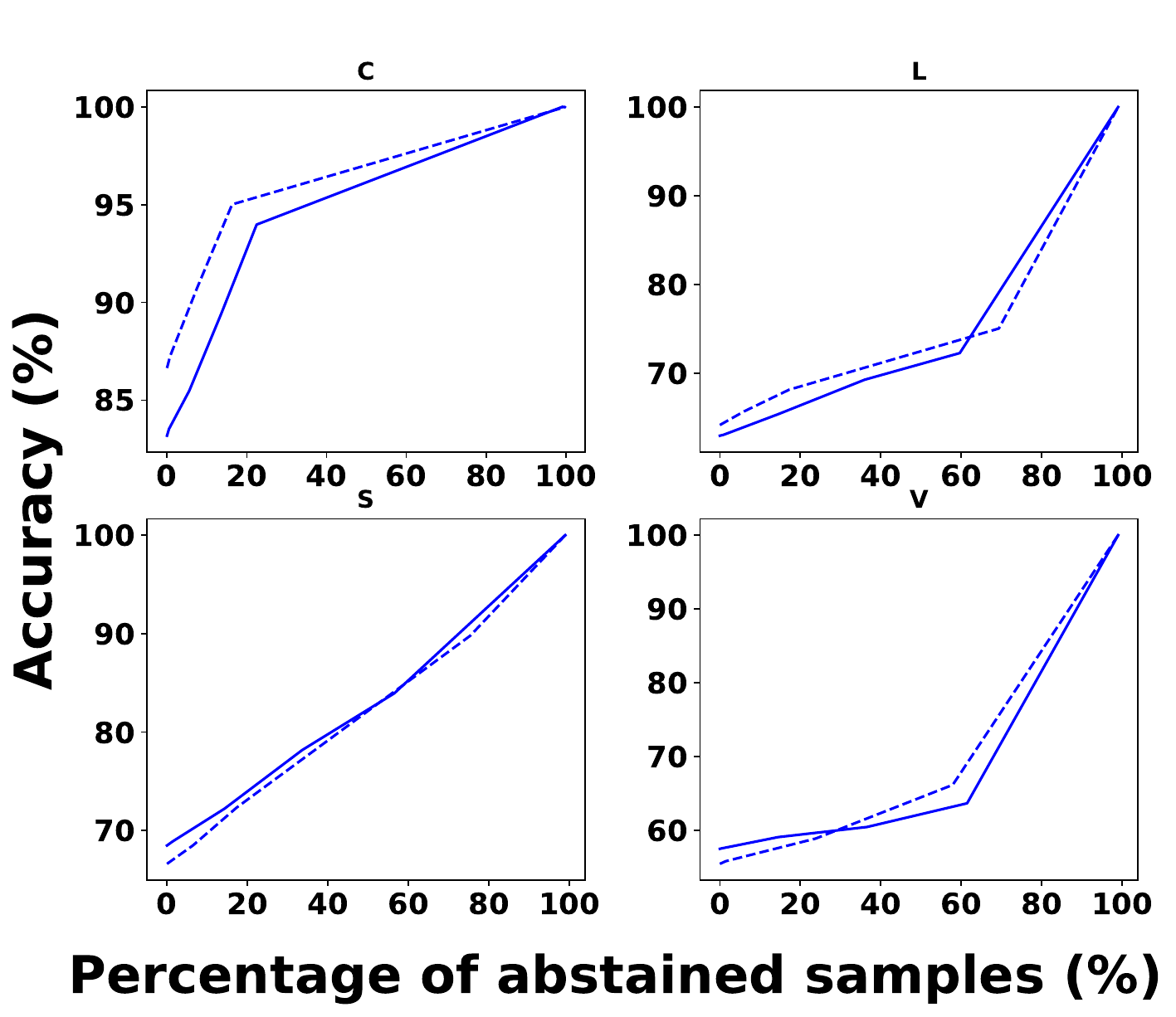}}
  \subfigure[Severity 3 corruptions]{\includegraphics[width=0.2\textwidth]{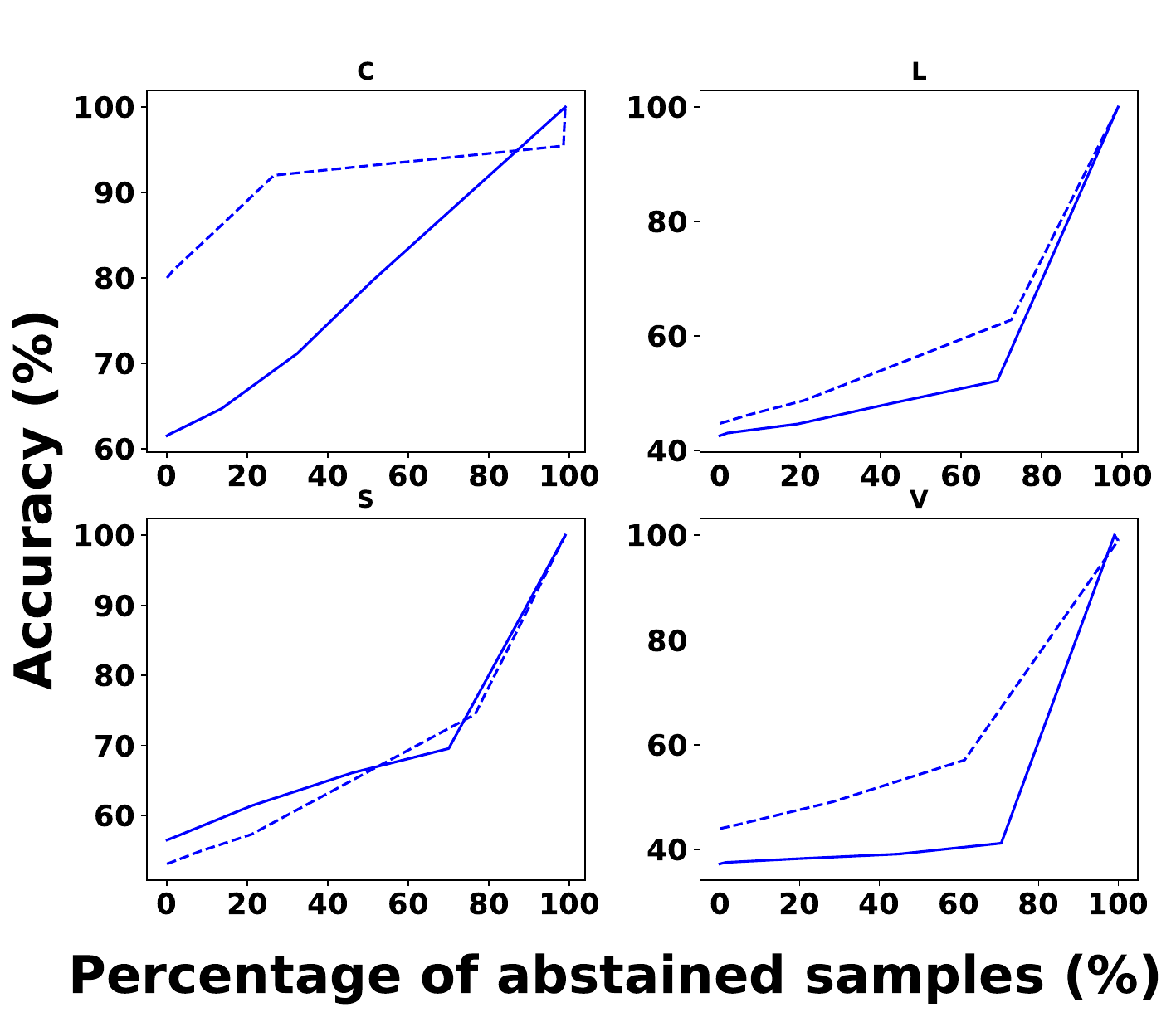}}
  \subfigure[Severity 5 corruptions]{\includegraphics[width=0.2\textwidth]{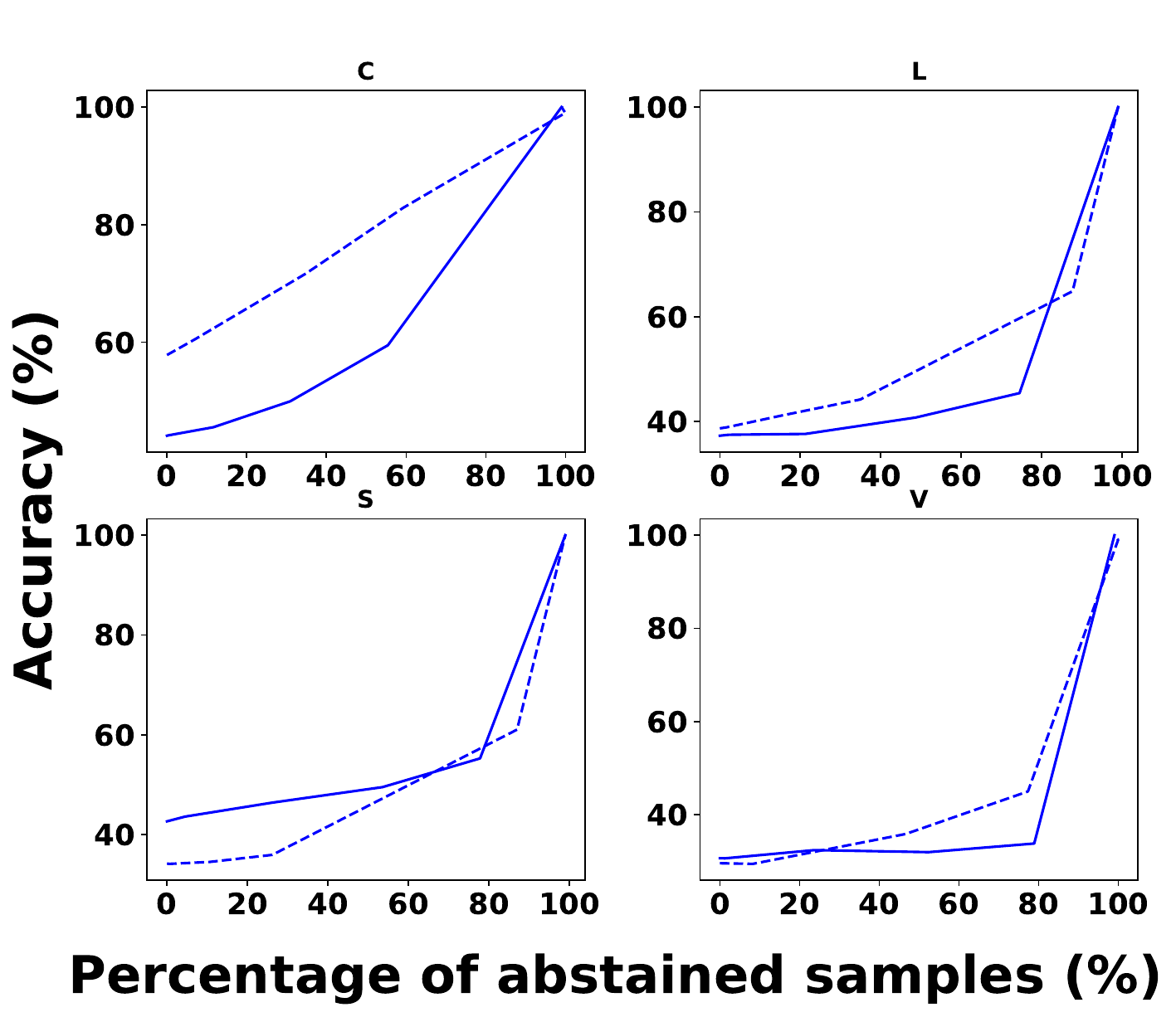}}

  }
  \caption{
    Comparison of TT-NSS (solid lines) and confidence-based method (dashed lines) in a {\bf single} (top row) and {\bf multiple} (bottom row) source domain setup on classifiers trained with ERM. 
    The graphs show accuracy vs abstained points on different variants of the {\bf VLCS} dataset ((a) original, (b) wikiart, (c,d) corrupted), and different source/target domains.
    In most domains, the accuracy of the TT-NSS (solid line) is similar to or better than the corresponding accuracy of the confidence-based method (dashed line) for most of the range of the percentage of abstained samples. 
    (Note: The source domain from VLCS used for training is denoted in the title in the top row and the target domain used for evaluation is denoted in the title in the bottom row.)
   } 
  \label{fig:inference_advantage_sconf_vs_conf_vlcs}
\end{figure*}

In Figs.~\ref{fig:inference_advantage_sconf_vs_conf_pacs_M} and~\ref{fig:inference_advantage_sconf_vs_conf_vlcs}, we show the full accuracy versus percentage of abstained sample curves for classifiers trained on PACS and VLCS dataset in both the single and multi-domain setting. The results show that the performance of the DG classifier when evaluated with TT-NSS remains better or competitive with the performance of the confidence-based abstaining method for most domains and most of the range of abstaining rates. 

\subsection{Additional results on the effectiveness of NSS at improving risk-averse predictions}
\label{app:comparison_nss_vs_others}
\begin{figure*}[tb]
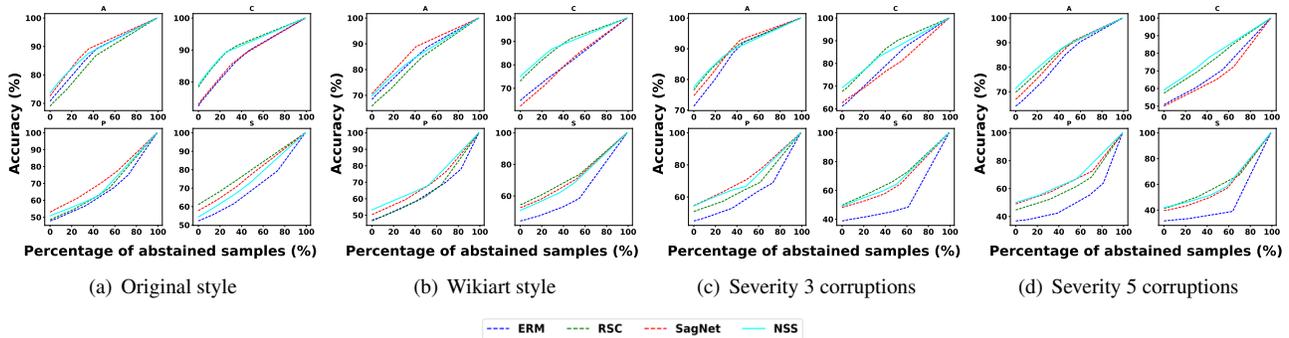

  \centering{
  \subfigure[Original  style]{\includegraphics[width=0.24\textwidth]{images/NSS_others_sconf_SD_original_style_pacs.pdf}}
  \subfigure[Wikiart  style]{\includegraphics[width=0.24\textwidth]{images/NSS_others_sconf_SD_wikiart_style_pacs.pdf}}
  \subfigure[Severity 3 corruptions]{\includegraphics[width=0.24\textwidth]{images/NSS_others_sconf_SD_severity_3_style_pacs.pdf}}
  \subfigure[Severity 5 corruptions]{\includegraphics[width=0.24\textwidth]{images/NSS_others_sconf_SD_severity_5_style_pacs.pdf}}
  \includegraphics[width=0.25\textwidth]{images/NSS_vs_others_legend.pdf}
  }
  \caption{
    Effectiveness of using NSS (with ERM) (solid lines) at improving the ability of DG classifiers at producing risk averse predictions when evaluated with TT-NSS in comparison to that of other DG methods (dashed lines) in a {\bf multi}-domain setup.
    NSS-trained classifiers achieve significantly better accuracy on non-abstained samples compared to classifiers trained with ERM and achieve competitive performance to models trained with RSC and SagNet at different abstaining rates on variants of the {\bf PACS} dataset in a multi-source domain setup.
    (See Fig.~\ref{fig:inference_advantage_sconf_vs_conf_pacs_M} for the explanation of settings.)
   } 
  \label{fig:inference_advantage_NSS_vs_others_sconf_pacs_M}
\end{figure*}

\begin{figure*}[tb]
  \centering{
  \subfigure[Original  style]{\includegraphics[width=0.24\textwidth]{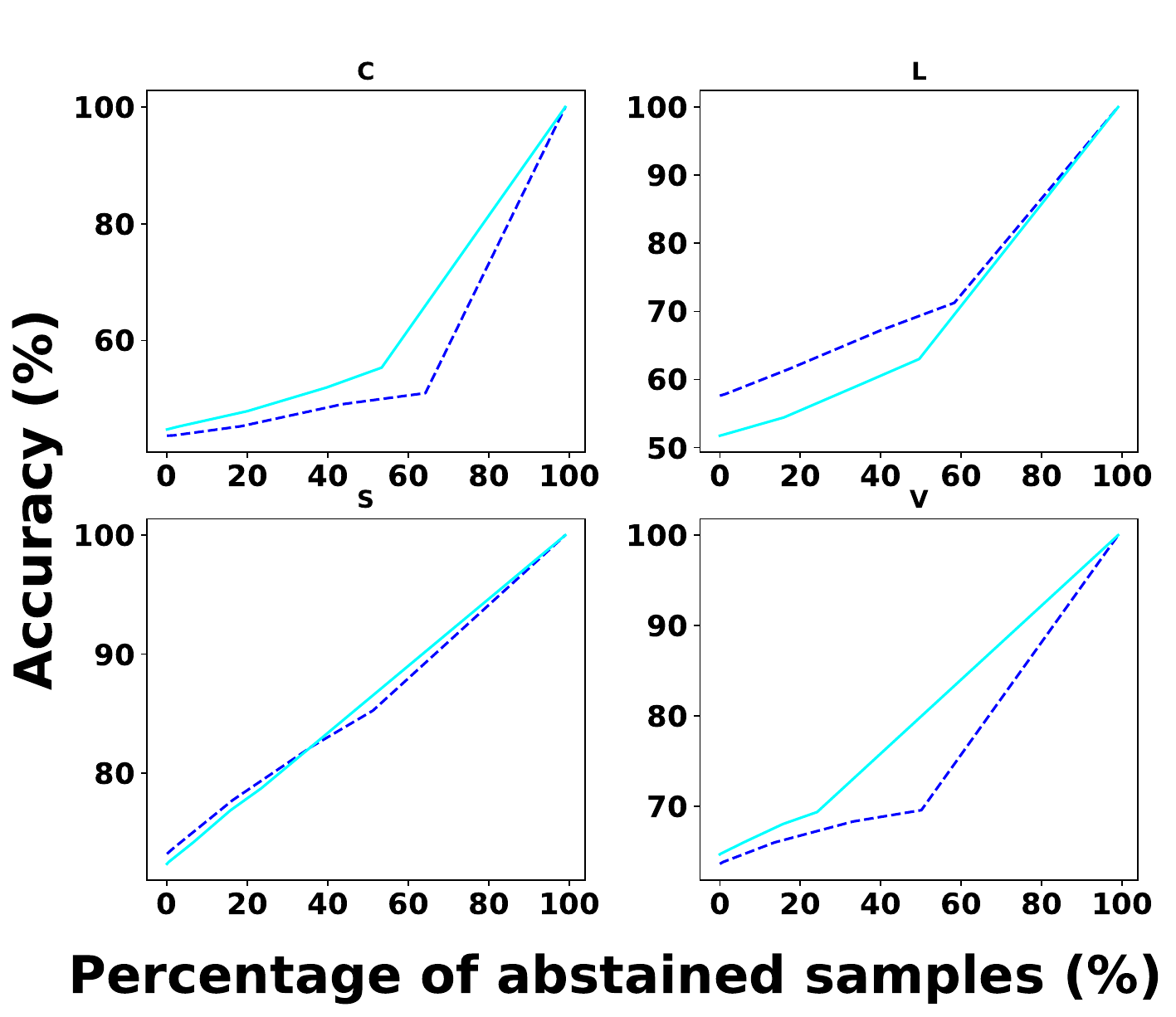}}
  \subfigure[Wikiart  style]{\includegraphics[width=0.24\textwidth]{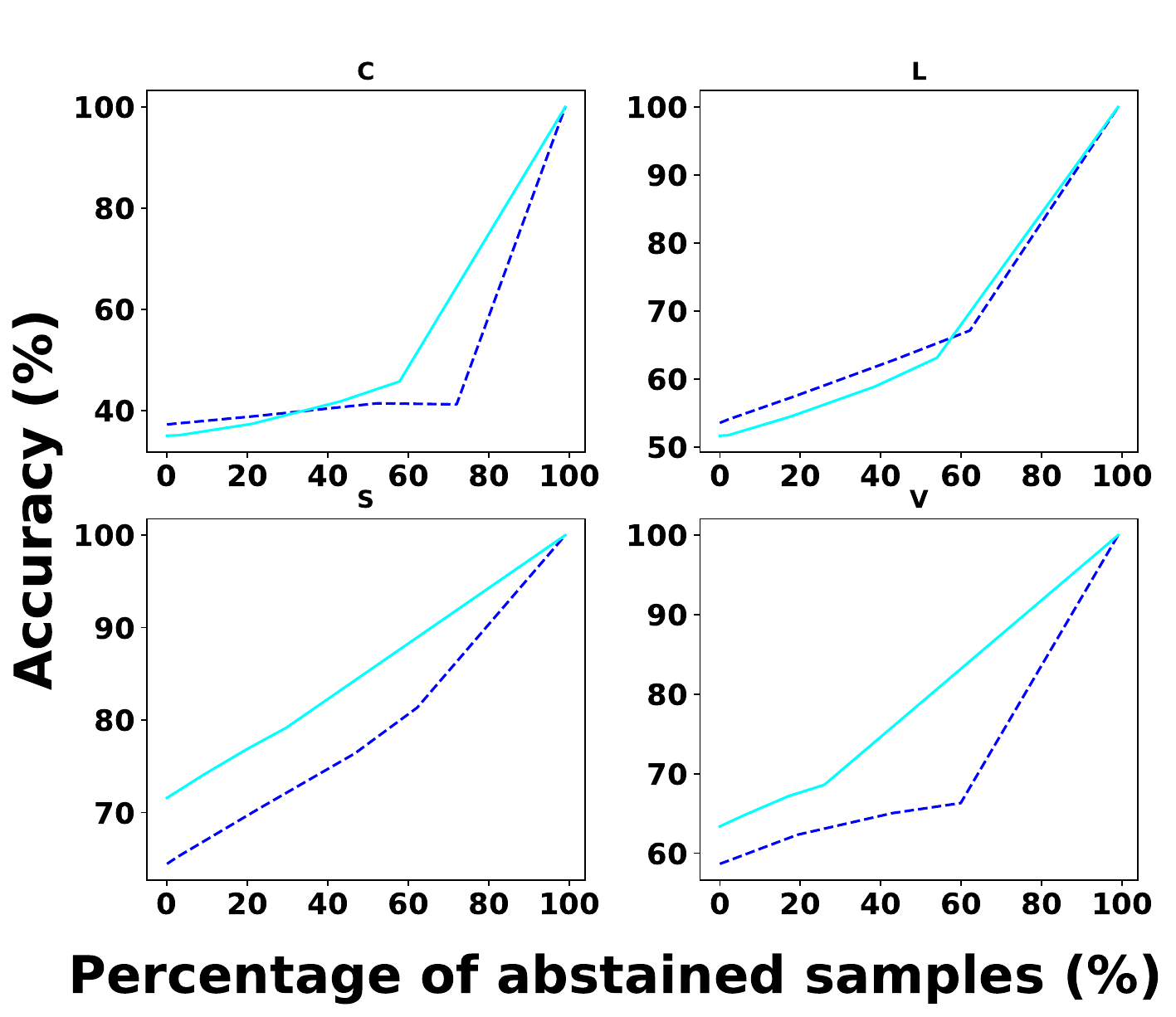}}
  \subfigure[Severity 3 corruptions]{\includegraphics[width=0.24\textwidth]{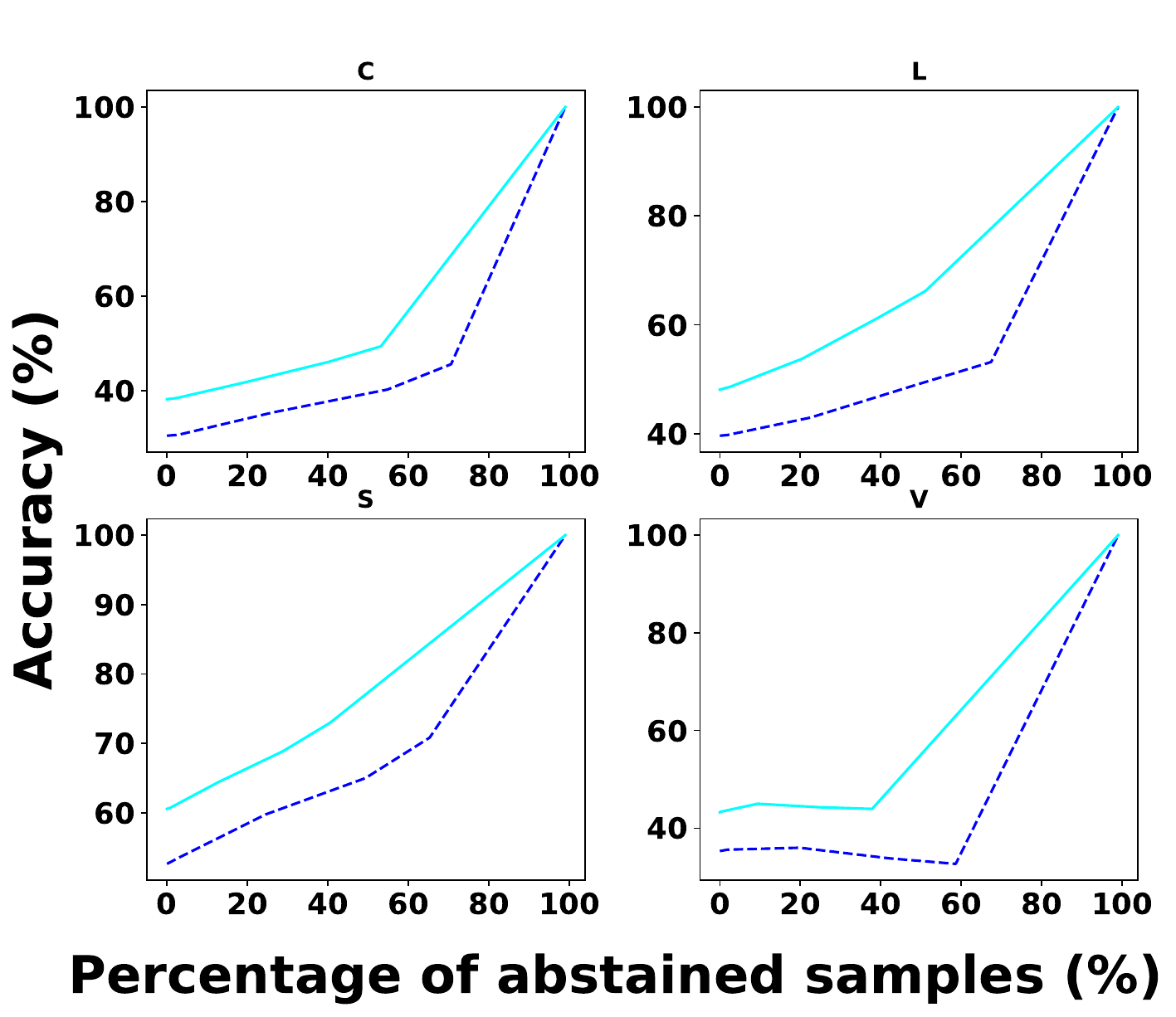}}
  \subfigure[Severity 5 corruptions]{\includegraphics[width=0.24\textwidth]{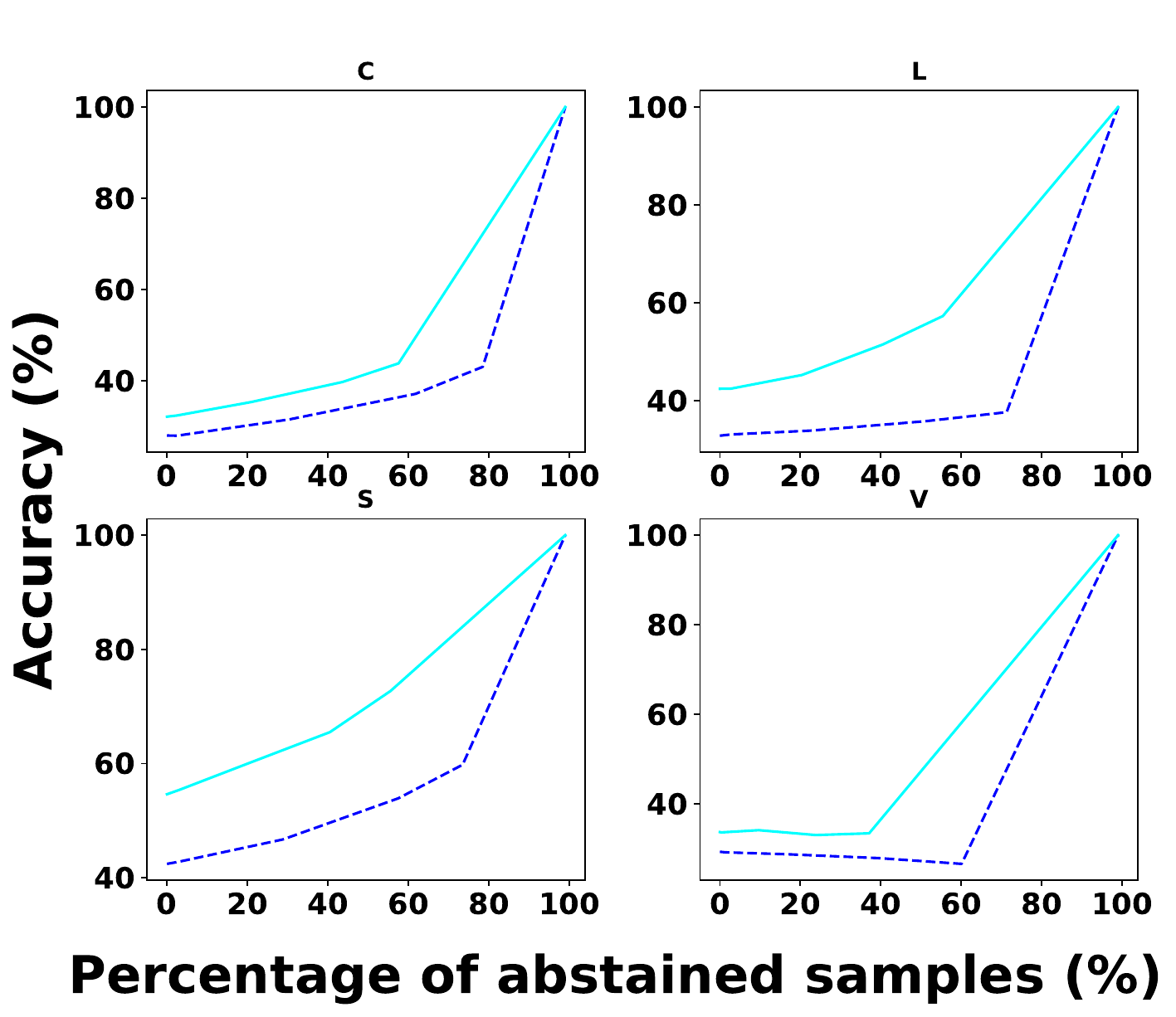}}
  \subfigure[Original  style]{\includegraphics[width=0.24\textwidth]{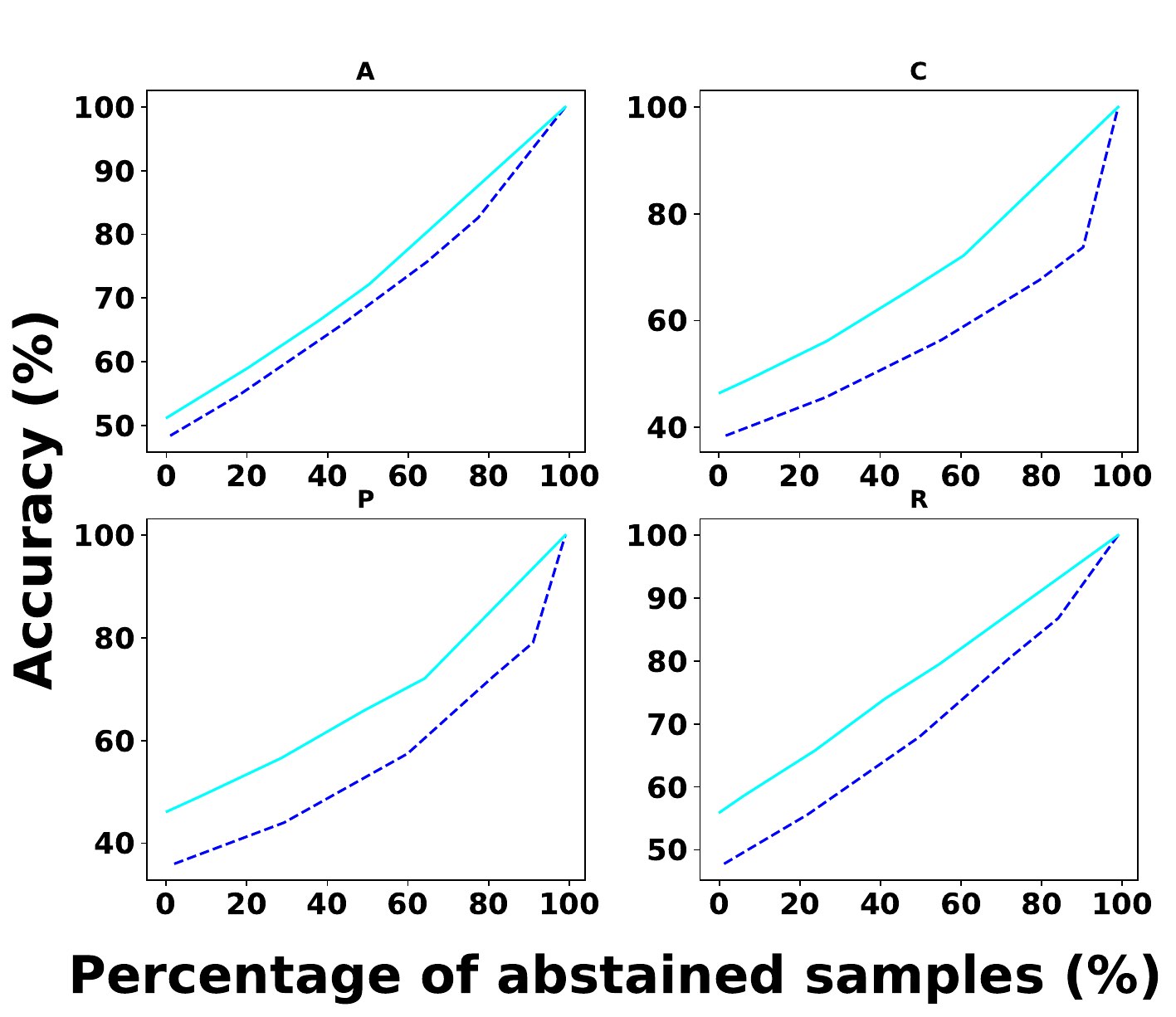}}
  \subfigure[Wikiart  style]{\includegraphics[width=0.24\textwidth]{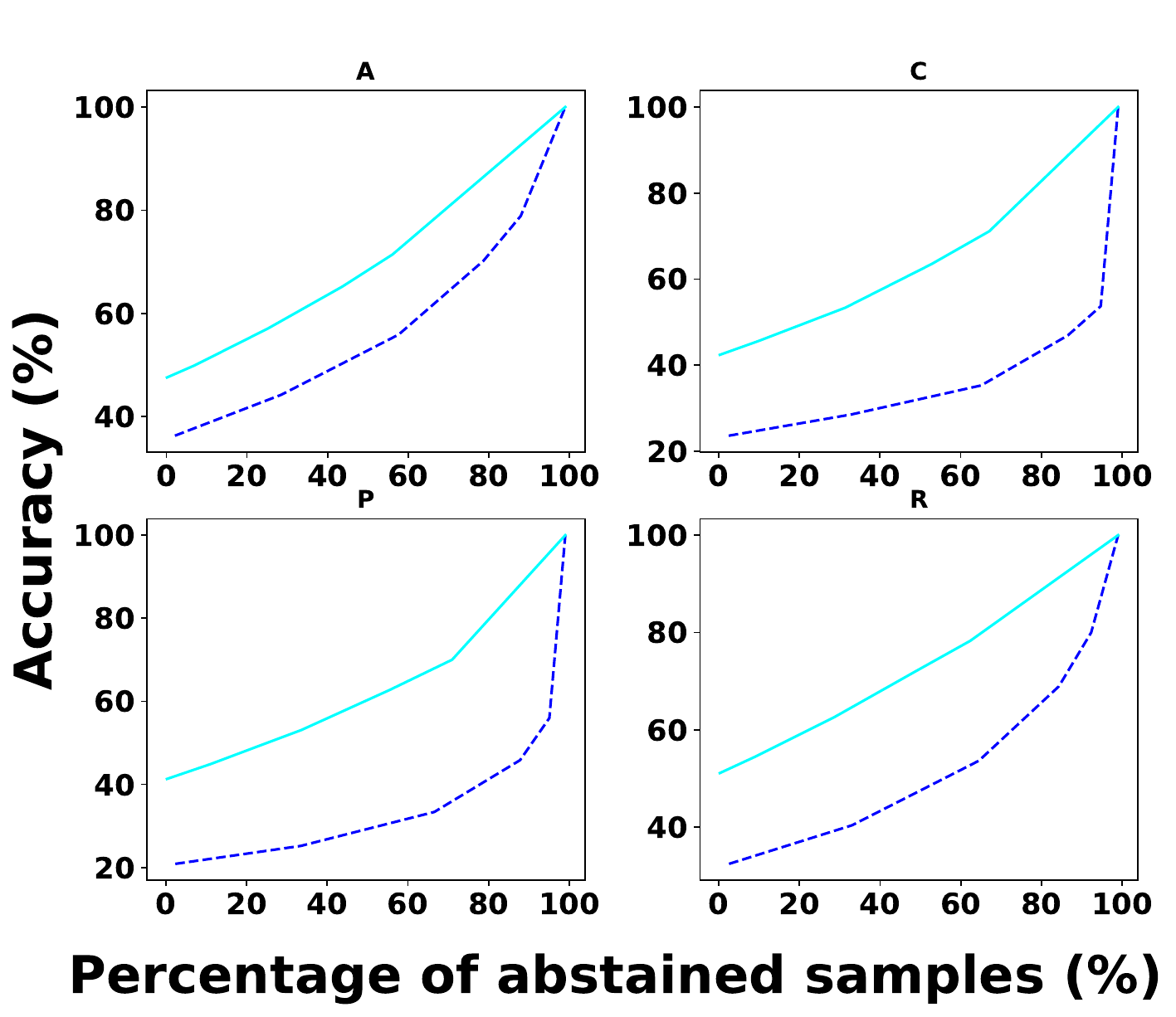}}
  \subfigure[Severity 3 corruptions]{\includegraphics[width=0.24\textwidth]{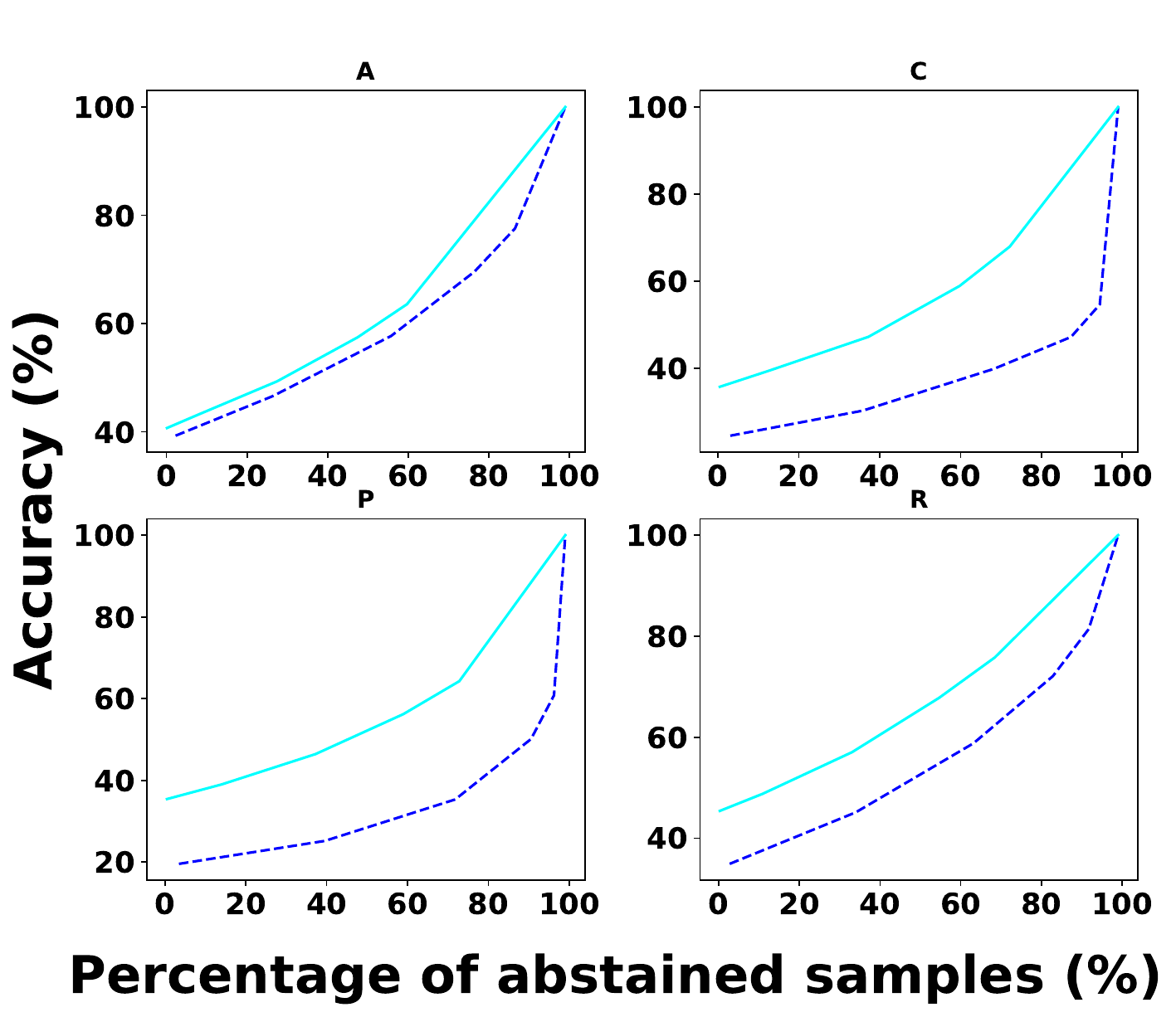}}
  \subfigure[Severity 5 corruptions]{\includegraphics[width=0.24\textwidth]{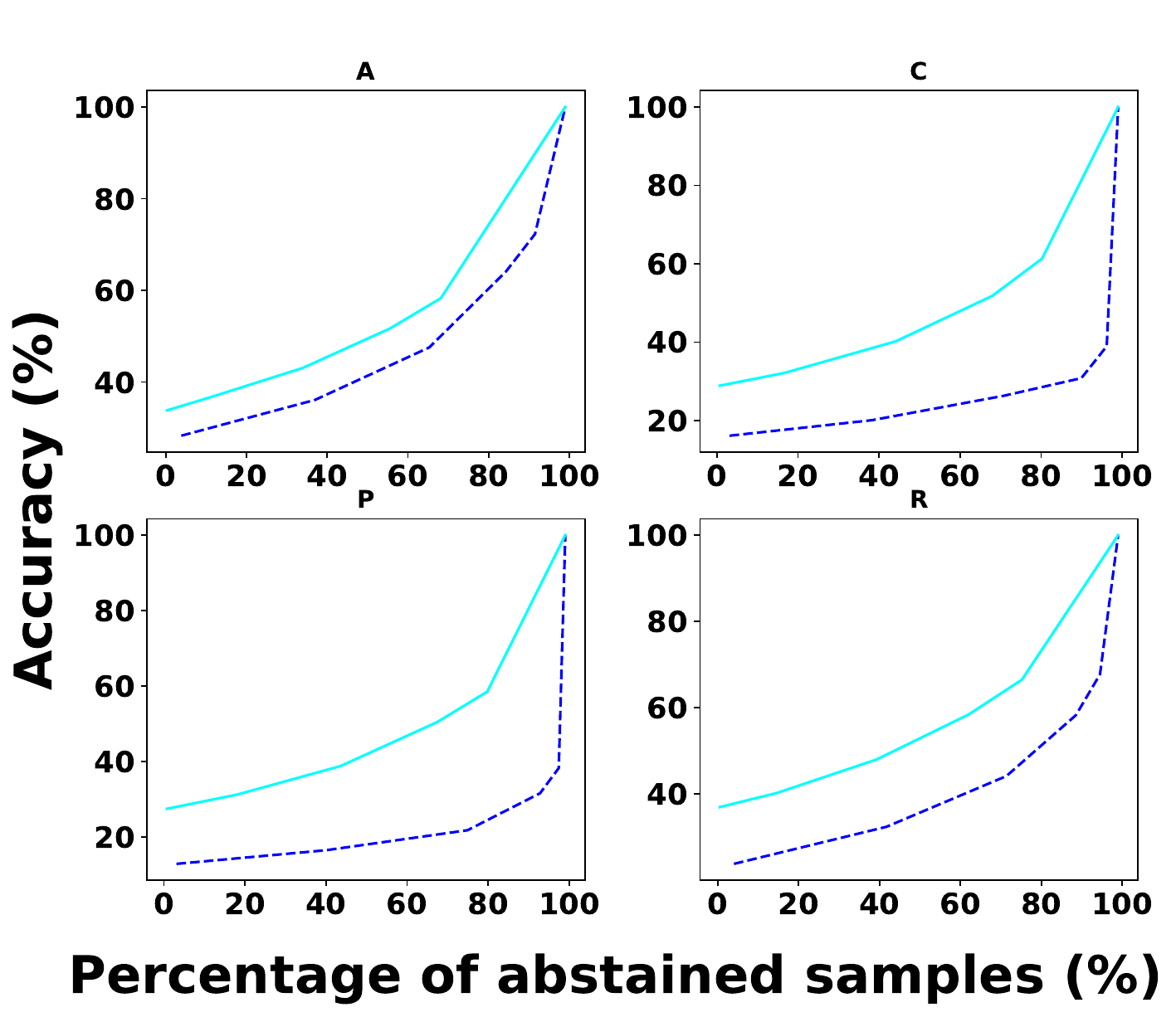}}
  \includegraphics[width=0.15\textwidth]{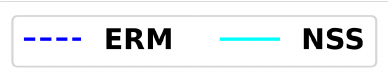}
  }
  \caption{
    Effectiveness of using NSS (with ERM as the base DG method) (solid lines) at improving the ability of DG to produce risk-averse predictions when evaluated with TT-NSS making it superior or competitive to classifiers trained with ERM (dashed lines) on variants of the {\bf VLCS} (top row) and {\bf OfficeHome} (bottom row) dataset in a {\bf single} source domain setup.
    (See Fig.~\ref{fig:inference_advantage_sconf_vs_conf_pacs} for the explanation of settings.)
   } 
  \label{fig:inference_advantage_NSS_vs_others_sconf_vlcs_office}
\end{figure*}

\begin{figure*}[tb]
  \centering{
  
  \subfigure[Original  style]{\includegraphics[width=0.24\textwidth]{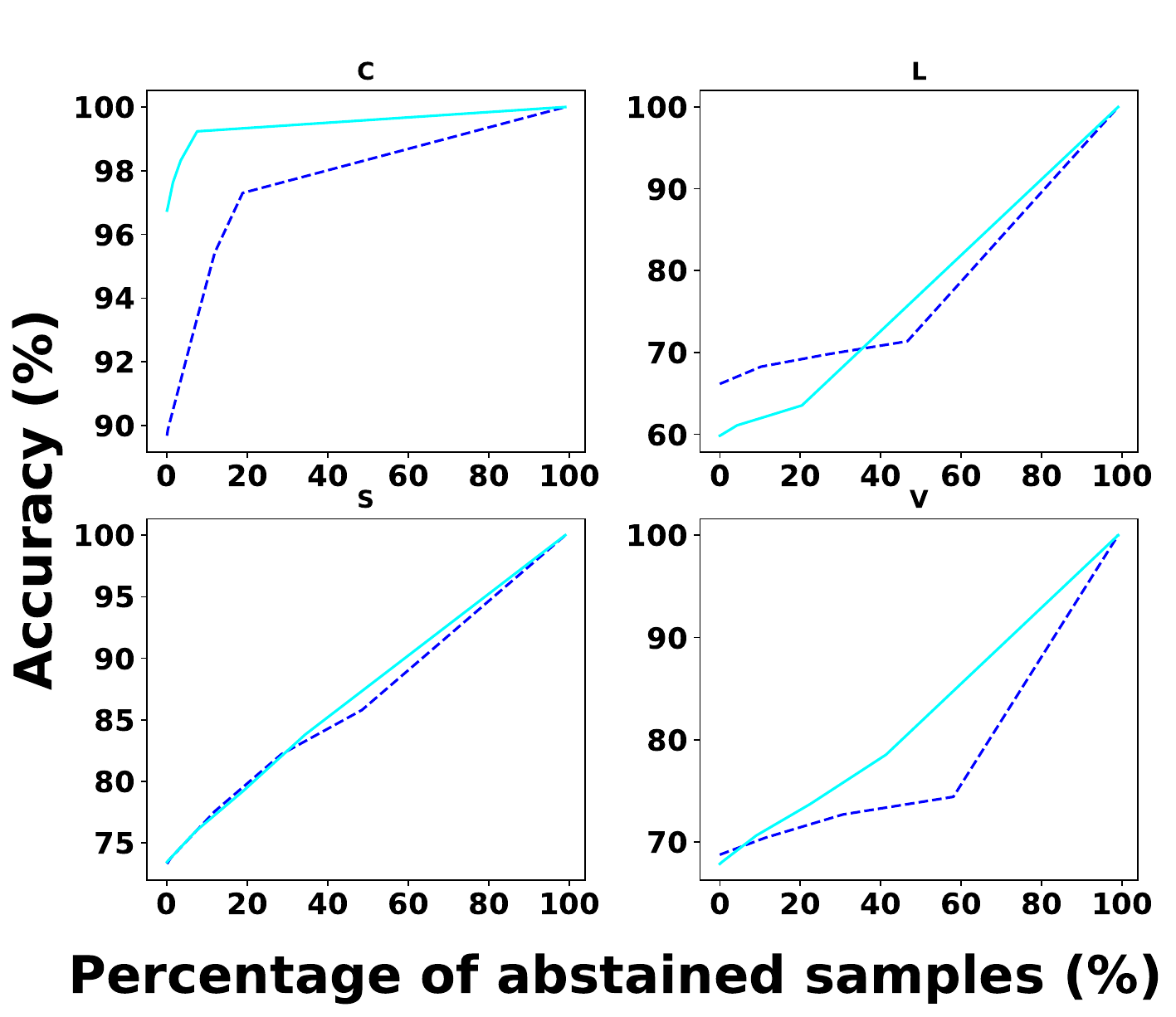}}
  \subfigure[Wikiart  style]{\includegraphics[width=0.24\textwidth]{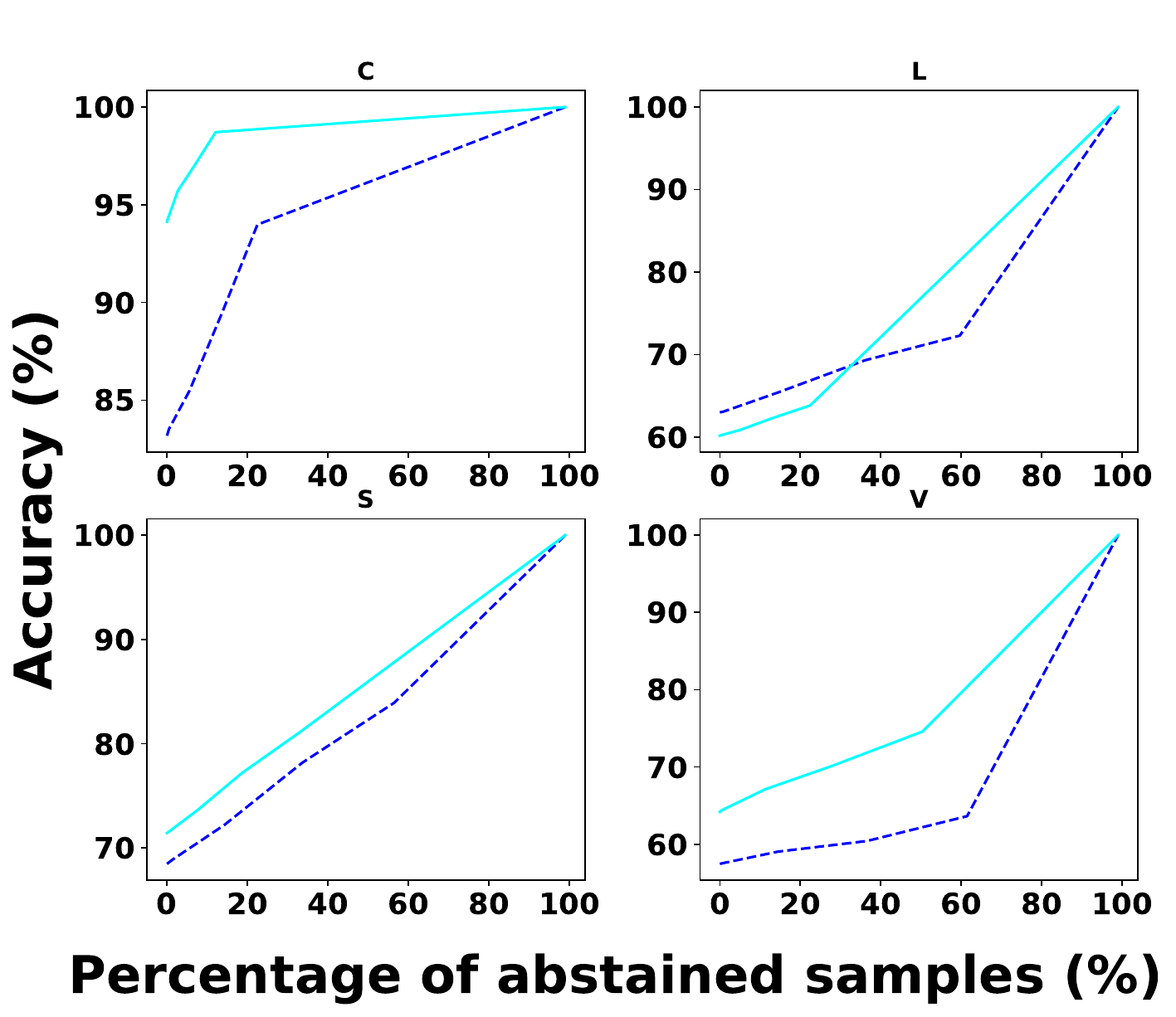}}
  \subfigure[Severity 3 corruptions]{\includegraphics[width=0.24\textwidth]{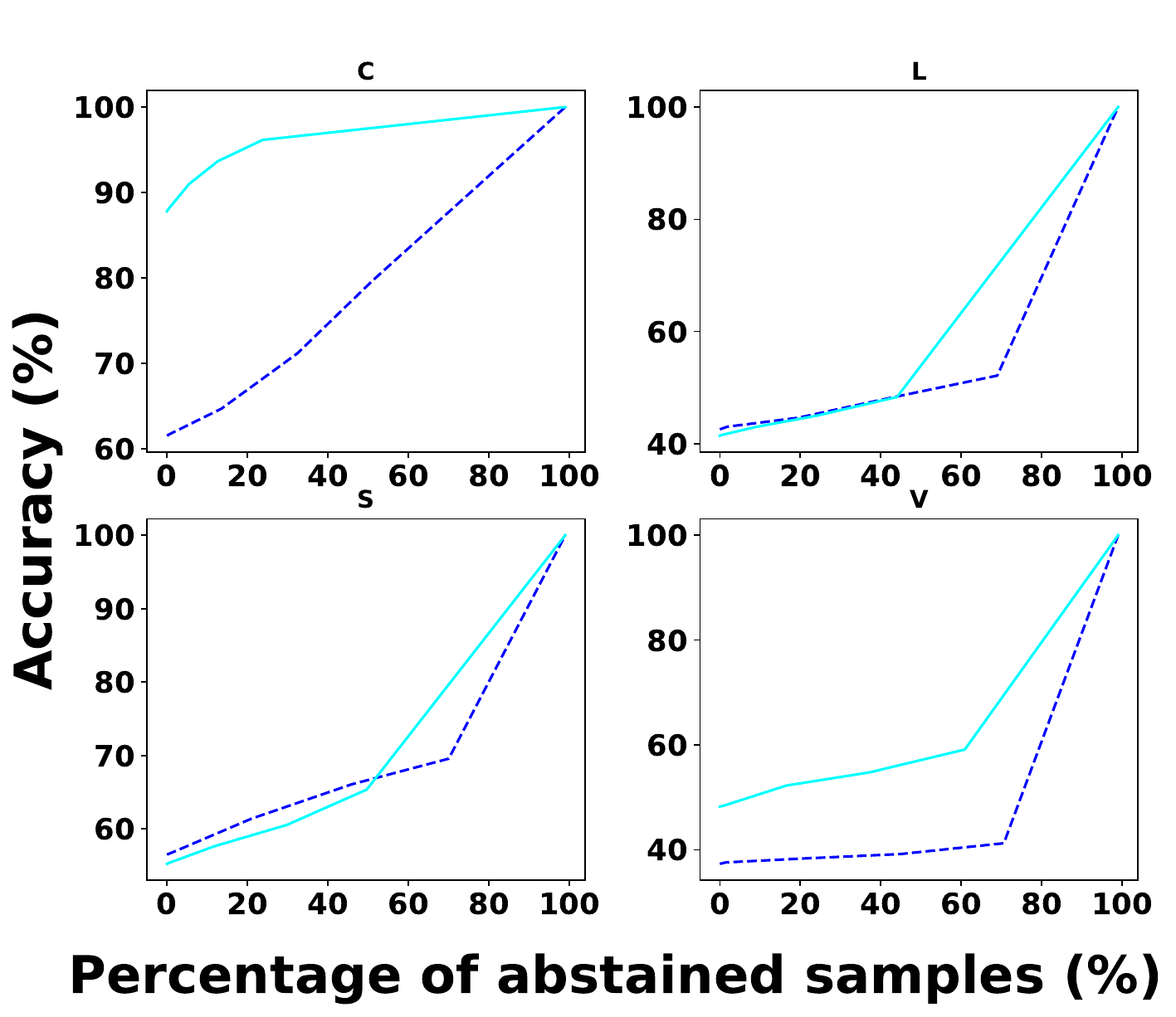}}
  \subfigure[Severity 5 corruptions]{\includegraphics[width=0.24\textwidth]{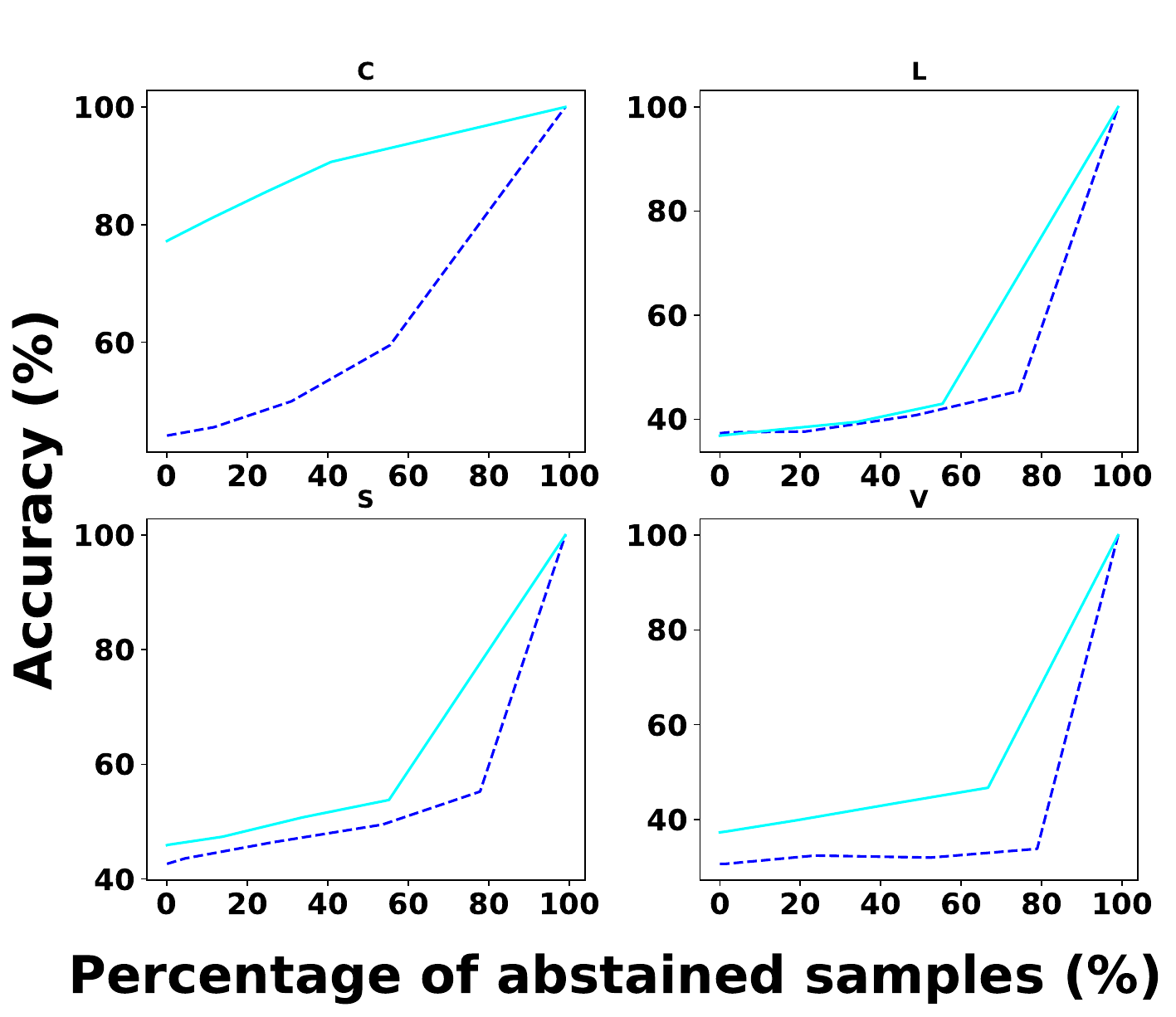}}

  \subfigure[Original  style]{\includegraphics[width=0.24\textwidth]{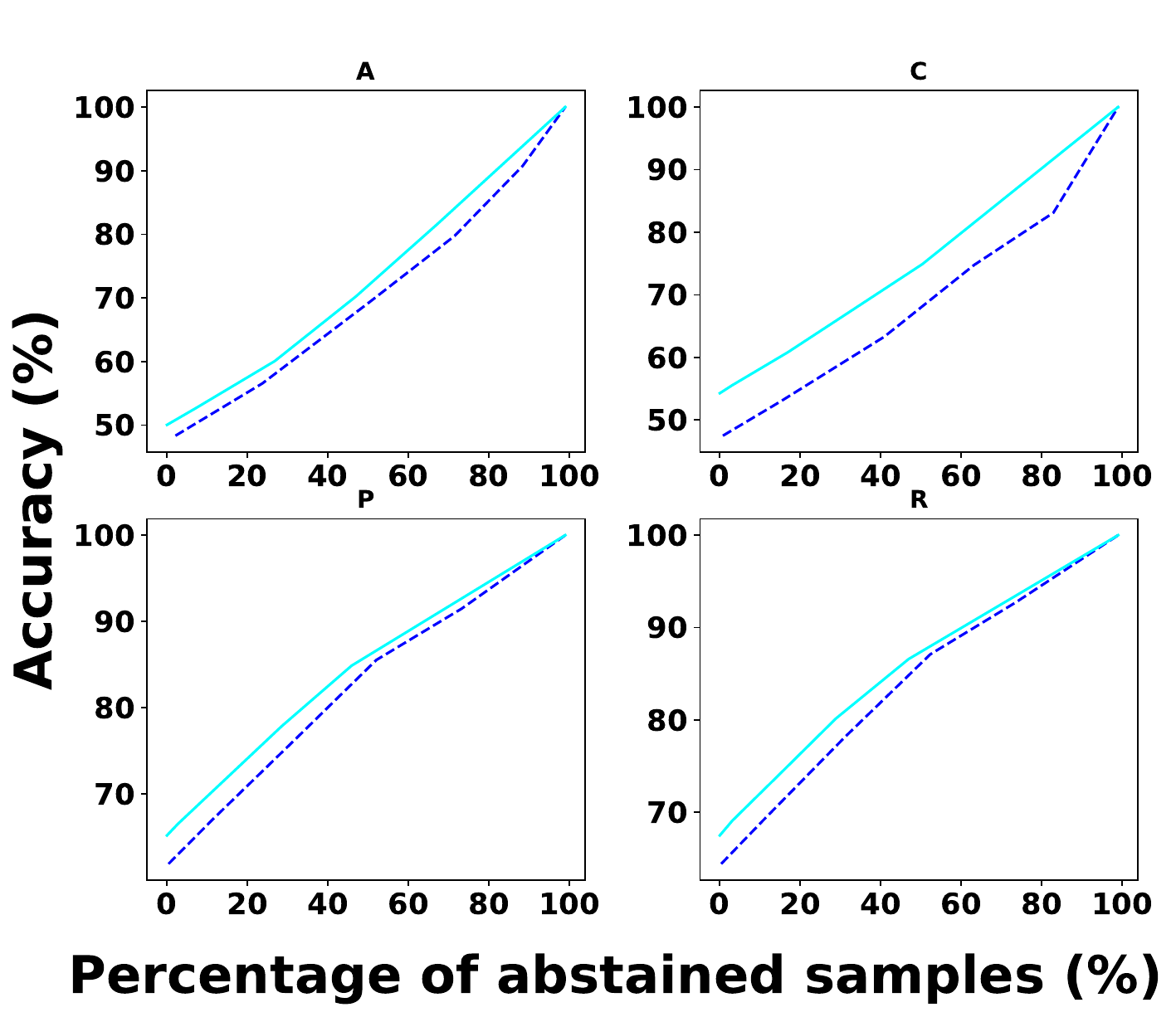}}
  \subfigure[Wikiart  style]{\includegraphics[width=0.24\textwidth]{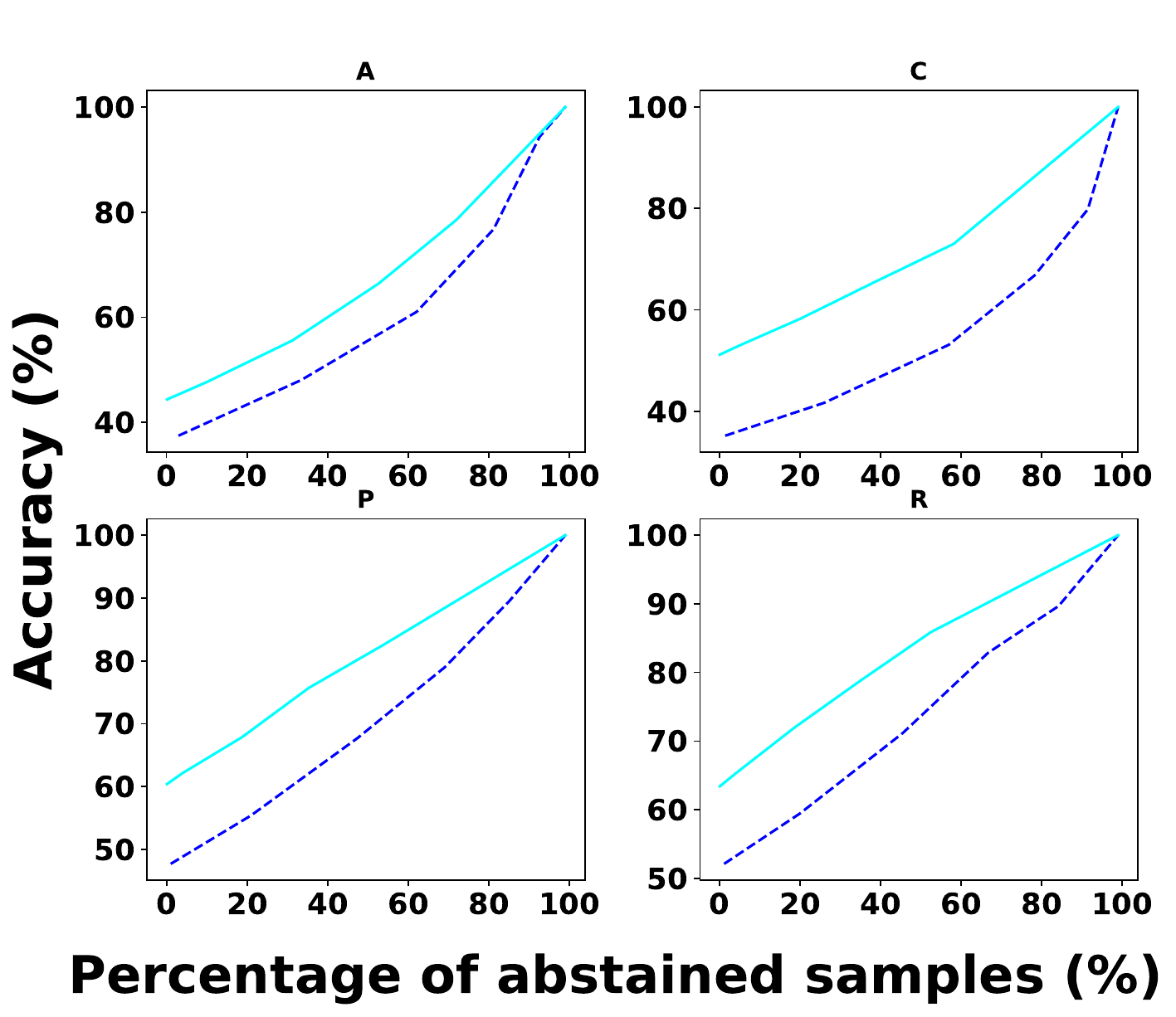}}
  \subfigure[Severity 3 corruptions]{\includegraphics[width=0.24\textwidth]{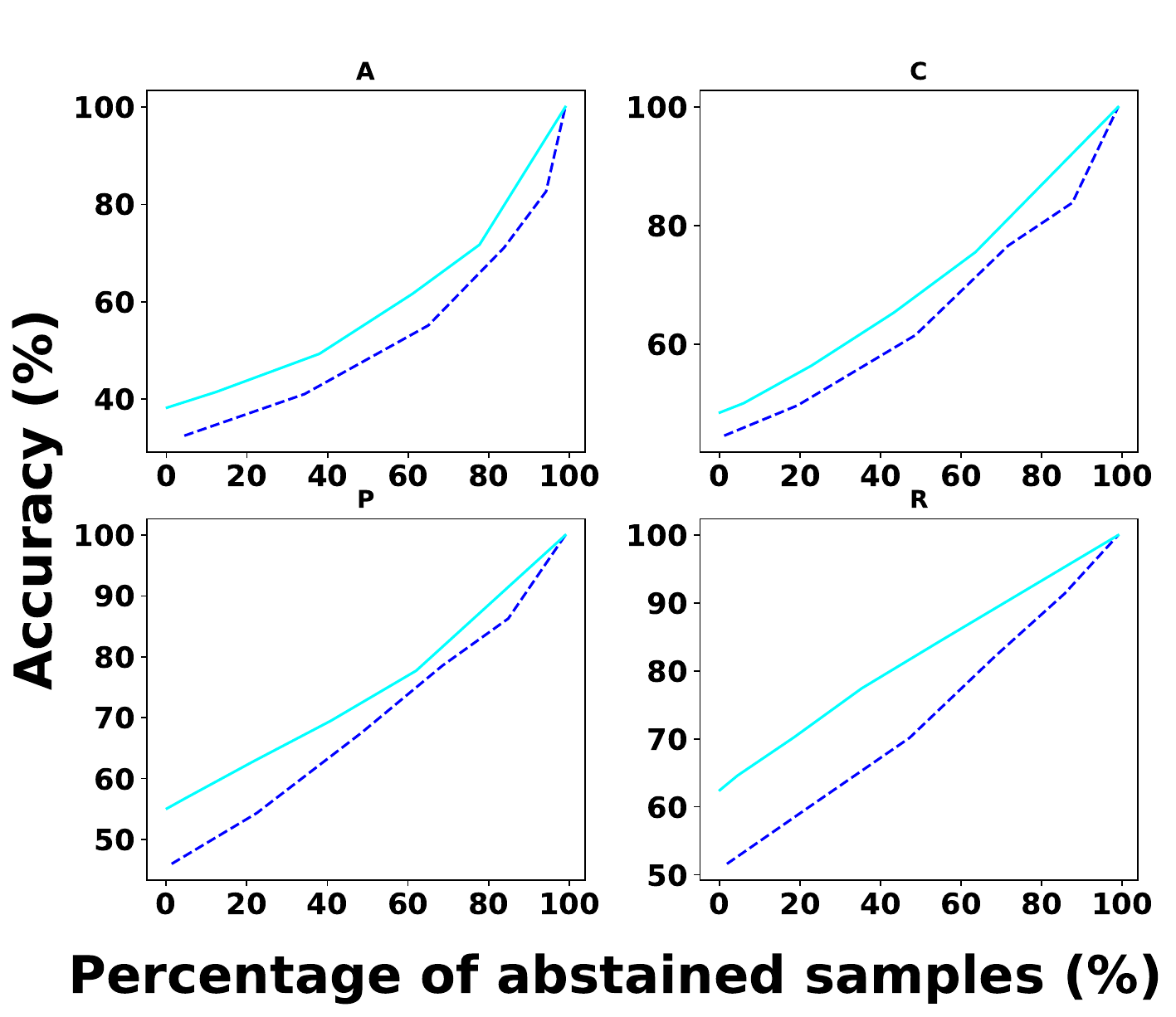}}
  \subfigure[Severity 5 corruptions]{\includegraphics[width=0.24\textwidth]{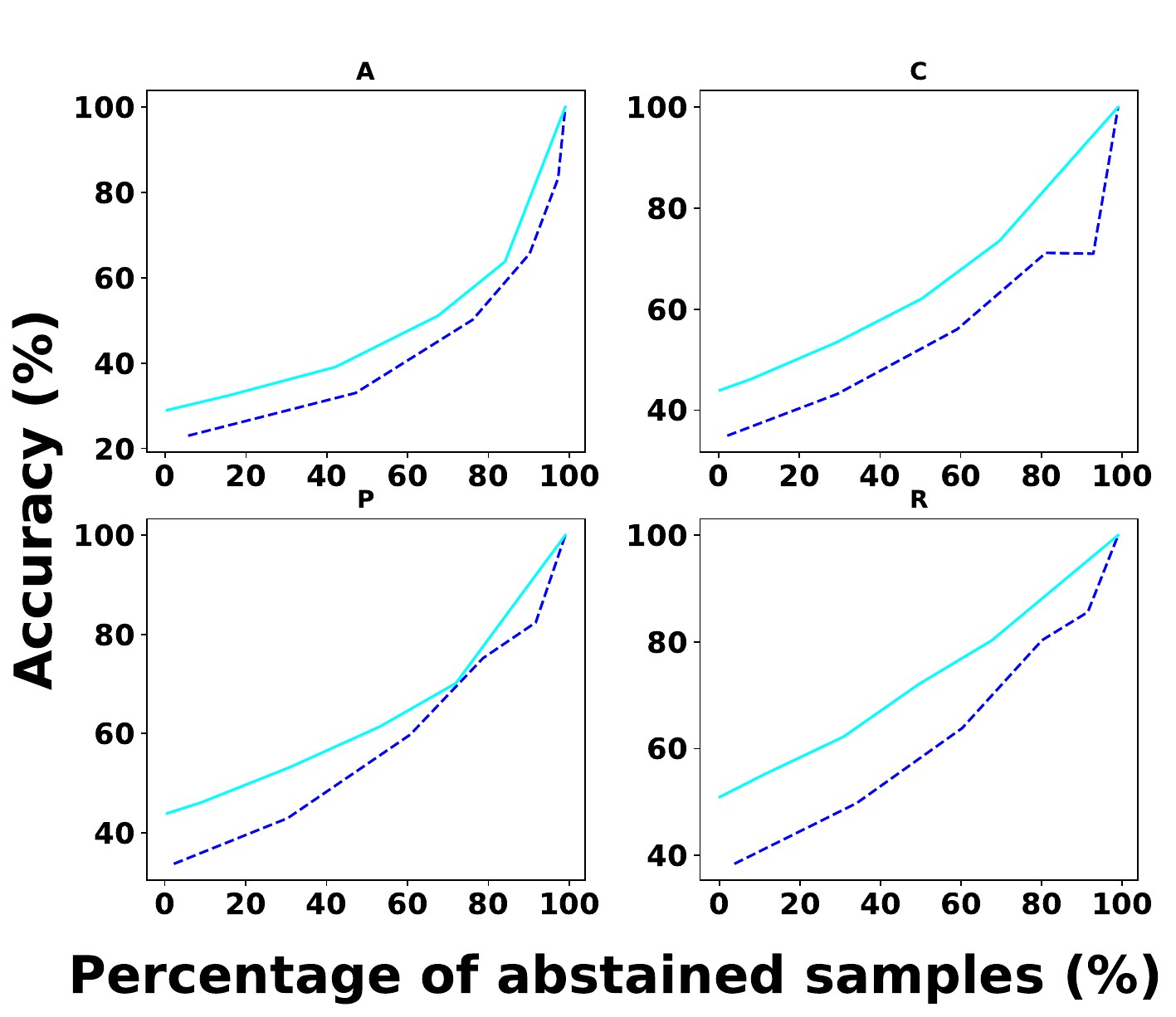}}
  \includegraphics[width=0.15\textwidth]{images/NSS_vs_ERM_legend.pdf}
  }
  \caption{
    Effectiveness of using NSS (with ERM as the base DG method) (solid lines) at improving the ability of DG to produce risk-averse predictions when evaluated with TT-NSS making it superior or competitive to classifiers trained with ERM (dashed lines) on variants of the  {\bf VLCS} (top row) and {\bf OfficeHome} (bottom row) dataset in a {\bf multi}-source domain setup.
    (See Fig.~\ref{fig:inference_advantage_sconf_vs_conf_pacs_M} for the explanation of settings.)
   } 
  \label{fig:inference_advantage_NSS_vs_others_sconf_vlcs_office_M}
\end{figure*}

In this section, we present additional empirical results on the effectiveness of training DG models with NSS (combined with ERM) on different datasets and settings.
Similar to the results in Sec.~\ref{sec:nss_risk_averse} in the main paper, we observe that models trained with NSS achieve consistently better AUC than models trained with ERM on different variants of PACS, VLCS and OfficeHome datasets as shown in Table~\ref{table:erm_vs_nss_auc_M}.
The highest improvement is observed when classifiers are evaluated on test sets corrupted with severity 5 corruptions.
Similar to Fig.~\ref{fig:inference_advantage_NSS_vs_others_sconf_pacs}, we observe in Fig.~\ref{fig:inference_advantage_NSS_vs_others_sconf_pacs_M},~\ref{fig:inference_advantage_NSS_vs_others_sconf_vlcs_office},~\ref{fig:inference_advantage_NSS_vs_others_sconf_vlcs_office_M} that NSS trained models achieve better accuracy on non-abstained samples on most domains compared to the models trained with ERM. 
Incorporating NSS with ERM makes the performance similar to that of other SOTA DG methods such as RSC and SagNet. 
Due to the versatility of NSS to be combined with any DG method, training classifiers with RSC and SagNet in conjunction with NSS could lead to further improvement in the accuracy of the classifier trained with these SOTA DG methods on non-abstained samples when evaluated with TT-NSS.
Lastly, classifiers trained with NSS also perform better in terms of risk-averse predictions when using the confidence-based abstaining mechanism as shown in Tables~\ref{table:erm_vs_nss_auc_conf} and~\ref{table:erm_vs_nss_auc_conf_M}. As mentioned in Sec.~\ref{sec:nss_risk_averse}, TT-NSS remains superior in the presence of severe shifts such as those induced by adding severity 5 corruptions for all the datasets in both single and multi-domain settings. 

\section{Dataset and experimental details}
\label{app:experimental_details}
All codes are written in Python using Tensorflow/Pytorch and were run on an AMD EPYC 7J13 CPU with 200 GB of RAM and an Nvidia A100 GPU. Implementation and hyperparameters are described below.

\subsection{Dataset description}
In this work, we use three popular benchmark datasets 
along with their stylized and corrupted version to evaluate the performance of various methods. 
For single source domain setting, we use 90\% of the data for training and 10\% for hyperparameter tuning, and for multi-domain setting, we use 80\% of the data for training and 20\% for hyperparameter tuning. 

{\bf PACS \cite{li2017deeper}:} This dataset contains images from four domains Art, Cartoons, Photos, and Sketches. 
It contains 9991 images belonging to 7 different classes.

{\bf VLCS \cite{fang2013unbiased}:}
This dataset contains images from four domains Caltech101, LabelMe, SUN09,
PASCAL VOC 2007.
It contains 10729 images belonging to 5 different classes.

{\bf Office-Home \cite{venkateswara2017deep}:}
This dataset contains images from four domains Art, Clipart, Product, and Real.
It contains 15588 images belonging to 65 different classes.

\subsection{Details of the subsample used for reporting the evaluation results in Sec.~\ref{sec:exp_num_styles}}
\label{app:subsample}
As mentioned in Sec.~\ref{sec:experiments}, we use a subsample of the PACS, VLCS and OfficeHome datasets to present the results of using TT-NSS and confidence based abstaining on corrupted variants of the datasets and for the experiment in Sec.~\ref{sec:exp_num_styles} with different values of $n$ in TT-NSS. 
For reporting the results on the corrupted version of the dataset we used 10 images per class from VLCS/PACS and 2 images per class from the OfficeHome dataset. We report average result over all 10 corruption types for this experiment.

For the experiment in Sec.~\ref{sec:exp_num_styles} we used the following subsample.
For the single source domain setting, we report the results on a balanced subsample of the dataset containing 50 images from each class and each target domain for PACS.
For the multi-domain  setting, we use 100 images for each class of the target domain for PACS.
For classes with
fewer samples, we use all the samples from that class

\subsection{Experimental details}

\subsubsection{Reproducing the baselines}
\label{app:reproducing_baselines}
For the RSC\cite{huang2020self} method, we independently run the code using the official implementation published by the authors, using different configurations (\url{https://github.com/DeLightCMU/RSC}). We trained both multi-domain and single-domain RSC\cite{huang2020self} classifiers with the same hyperparameters except for smaller batch size 2 and a learning rate of 0.0001 on one random seed. 
For the SagNet\cite{nam2021reducing}, we reproduce their open-source implementation code with the default configuration on three different random seeds (\url{https://github.com/hyeonseobnam/sagnet}). 
We use the official train and test split of PACS for all three methods. 
Table~\ref{table:reproducr_of_baselines} shows our reproduced results and the results the authors reported in their papers.

\begin{table}
  \begin{center}
    \captionof{table}{%
       Results on single and multi-domain generalization settings using ResNet50 as the backbone on the PACS dataset using RSC \cite{huang2020self} and SagNet \cite{nam2021reducing}. The original work, RSC \cite{huang2020self}, only reports multi-domain results (presented without *) while SagNet \cite{nam2021reducing}, only reported results based on the ResNet-18 backbone in the original paper. 
       We used their official implementation using ResNet-50 as the backbone to obtain results for both single and multi-domain settings (reported with *) (see details in Appendix~\ref{app:reproducing_baselines}). 
      \label{table:reproducr_of_baselines}
    }
    \resizebox{0.99\columnwidth}{!}{
    \begin{tabular}{|c|c|cccc|c|}
    \hline
    DG Setting & Methods & A & C & P & S & Avg. \\
    \hline
    \multirow{2}{*}{Single} & RSC* & 72.55 & 77.30 & 47.88 & 57.54 & 63.82 \\
    & SagNet*  & 77.45 & 78.36 & 52.39 & 53.96 & 65.54 \\
    \hline
   \multirow{3}{*}{Multi} &  RSC & 87.89 & 82.16 & 97.92 & 83.35 & 87.83 \\
   & RSC* & 85.79 & 79.60 & 95.03 & 81.52 & 85.49  \\
    
    & SagNet* & 86.00 & 81.29 & 97.47 & 80.72 & 86.37  \\
    \hline
    \end{tabular}
    }
  \end{center}
\end{table}

\subsubsection{Training classifiers with NSS}
To train the classifiers with NSS, we incorporate style augmentation and style consistency losses computed on stylized versions of the source domain images generated through the AdaIN decoder. 
We additionally incorporate the ERM training loss which minimizes the misclassification on original source domain samples. 
As mentioned in Sec.~\ref{sec:style_adaptation} other  losses used in specific DG algorithms can also be incorporated to improve the quality of risk-averse predictions from classifiers trained with those methods. 
To compute the style consistency loss we use four different styles for every sample in the batch and use a batch size of 16. 
These losses are then used to fine-tune the ResNet50 backbone augmented with a fully connected layer used for classification.
For the multi-domain  setting, the classifier that achieves the highest accuracy on the validation set is used for final evaluation whereas for the single source domain setting, the classifier at the last step is used for final evaluation.

\clearpage

\begin{table}[t]
  \begin{center}
    \captionof{table}{%
      Comparison of the area under the accuracy versus percentage of abstained samples curve for TT-NSS and the confidence-based abstaining mechanism in a {\bf single} domain setting on different variations of the {\bf PACS} dataset. The training domain is denoted in the columns.
      \label{table:conf_vs_sconf_pacs}
    }
    \resizebox{0.85\columnwidth}{!}{
    \begin{tabular}{|c|c|cccc|}

      \hline
      \multicolumn{2}{|c|}{} & A & C & P & S \\
      
      \hline

      Alg. & Evaluation & \multicolumn{4}{|c|}{Original Style}  \\
      
      \hline
      
      \multirow{2}{*}{ERM} & Confidence & {\bf 0.882} & 0.875 & 0.634 & 0.707
\\
& TT-NSS & 0.875 & 0.878 & {\bf 0.662} & 0.702
\\ \hline
\multirow{2}{*}{RSC} & Confidence & {\bf 0.892} & 0.899 & {\bf 0.705} & 0.779
\\
& TT-NSS & 0.858 & {\bf 0.912} & 0.682 & {\bf 0.794}
\\ \hline
\multirow{2}{*}{SagNet} & Confidence & {\bf 0.913} & {\bf 0.91} & {\bf 0.741} & 0.758
\\
& TT-NSS & 0.889 & 0.88 & 0.726 & {\bf 0.771}
\\ \hline
      
      &  & \multicolumn{4}{|c|}{Wikiart Style} \\

      \hline
      
      \multirow{2}{*}{ERM} & Confidence & 0.84 & 0.757 & 0.609 & 0.558
        \\
        & TT-NSS & {\bf 0.854} & {\bf 0.816} & {\bf 0.643} & {\bf 0.626}
        \\ \hline
        \multirow{2}{*}{RSC} & Confidence & 0.823 & 0.766 & 0.63 & 0.662
        \\
        & TT-NSS & {\bf 0.835} & {\bf 0.887} & {\bf 0.654} & {\bf 0.733}
        \\ \hline
        \multirow{2}{*}{SagNet} & Confidence & 0.871 & 0.8 & 0.683 & 0.61
        \\
        & TT-NSS & 0.875 & {\bf 0.813} & {\bf 0.692} & {\bf 0.718}
        \\ \hline
        
       &  & \multicolumn{4}{|c|}{Corrupted with severity 3} \\
      \hline
       
      \multirow{2}{*}{ERM} & Confidence & 0.832 & 0.709 & 0.613 & {\bf 0.612}
\\
& TT-NSS & {\bf 0.886} & {\bf 0.812} & {\bf 0.622} & 0.545
\\ \hline
\multirow{2}{*}{RSC} & Confidence & 0.871 & 0.667 & 0.673 & 0.62
\\
& TT-NSS & {\bf 0.901} & {\bf 0.86} & {\bf 0.682} & {\bf 0.699}
\\ \hline
\multirow{2}{*}{SagNet} & Confidence & 0.903 & 0.78 & 0.725 & 0.629
\\
& TT-NSS & 0.901 & {\bf 0.794} & {\bf 0.731} & {\bf 0.667}
\\ \hline
      
      &  & \multicolumn{4}{|c|}{Corrupted with severity 5} \\
      \hline
       
      \multirow{2}{*}{ERM} & Confidence & 0.696 & 0.579 & 0.418 & {\bf 0.479}
\\
& TT-NSS & {\bf 0.834} & {\bf 0.708} & {\bf 0.519} & 0.468
\\ \hline
\multirow{2}{*}{RSC} & Confidence & 0.728 & 0.449 & 0.564 & 0.465
\\
& TT-NSS & {\bf 0.863} & {\bf 0.776} & {\bf 0.626} & {\bf 0.613}
\\ \hline
\multirow{2}{*}{SagNet} & Confidence & 0.786 & 0.576 & 0.565 & 0.485
\\
& TT-NSS & {\bf 0.855} & {\bf 0.686} & {\bf 0.666} & {\bf 0.59}
\\ \hline

    \end{tabular}
    }
  \end{center}

\end{table}

\begin{table}[t]
  \begin{center}
    \captionof{table}{%
       Comparison of the area under the accuracy versus percentage of abstained samples curve for TT-NSS and the confidence-based abstaining mechanism in a {\bf single} domain setting on different variations of the {\bf VLCS} dataset. The training domain is denoted in the columns.
      \label{table:conf_vs_sconf_vlcs}
    }
    \resizebox{0.85\columnwidth}{!}{
    \begin{tabular}{|c|c|cccc|}

      \hline
      \multicolumn{2}{|c|}{} & A & C & P & S \\
      
      \hline

      Alg. & Evaluation & \multicolumn{4}{|c|}{Original Style}  \\
      
      \hline
      
      \multirow{2}{*}{ERM} & Confidence & {\bf 0.653} & 0.68 & 0.806 & 0.715
\\
& TT-NSS & 0.567 & {\bf 0.724} & {\bf 0.851} & {\bf 0.751}
\\ \hline
      
      &  & \multicolumn{4}{|c|}{Wikiart Style} \\

      \hline
      
      \multirow{2}{*}{ERM} & Confidence & 0.426 & 0.584 & 0.763 & 0.679
\\
& TT-NSS & {\bf 0.477} & {\bf 0.682} & {\bf 0.785} & {\bf 0.704}
\\ \hline
        
       &  & \multicolumn{4}{|c|}{Corrupted with severity 3} \\
      \hline
       
      \multirow{2}{*}{ERM} & Confidence & {\bf 0.504} & 0.381 & {\bf 0.734} & 0.468
\\
& TT-NSS & 0.468 & {\bf 0.551} & 0.689 & {\bf 0.471}
\\ \hline
      
      &  & \multicolumn{4}{|c|}{Corrupted with severity 5} \\
      \hline
       
     \multirow{2}{*}{ERM} & Confidence & {\bf 0.433} & 0.329 & 0.563 & 0.346
\\
& TT-NSS & 0.411 & {\bf 0.439} & 0.567 & {\bf 0.415}
\\ \hline

    \end{tabular}
    }
  \end{center}

\end{table}

\begin{table}
  \begin{center}
    \captionof{table}{%
       Comparison of the area under the accuracy versus percentage of abstained samples curve for TT-NSS and the confidence-based abstaining mechanism in a {\bf multi-}domain setting on different variations of the {\bf PACS} dataset. The domain used for evaluation is denoted in the columns.
      \label{table:conf_vs_sconf_pacs_M}
    }
    \resizebox{0.85\columnwidth}{!}{
    \begin{tabular}{|c|c|cccc|}

      \hline
      \multicolumn{2}{|c|}{} & A & C & P & S \\
      
      \hline

      Alg. & Evaluation & \multicolumn{4}{|c|}{Original Style}  \\
      
      \hline
      
      \multirow{2}{*}{ERM} & Confidence & {\bf 0.95} & 0.902 & 0.986 & 0.915
\\
& TT-NSS & 0.893 & 0.9 & 0.978 & 0.911
\\ \hline
\multirow{2}{*}{RSC} & Confidence & 0.925 & 0.908 & 0.978 & {\bf 0.936}
\\
& TT-NSS & {\bf 0.948} & {\bf 0.926} & 0.983 & 0.917
\\ \hline
\multirow{2}{*}{SagNet} & Confidence & {\bf 0.951} & 0.932 & 0.988 & 0.905
\\
& TT-NSS & 0.927 & 0.939 & 0.984 & {\bf 0.925}
\\ \hline
      
      &  & \multicolumn{4}{|c|}{Wikiart Style} \\

      \hline
      
      \multirow{2}{*}{ERM} & Confidence & {\bf 0.898} & 0.85 & 0.975 & 0.892
\\
& TT-NSS & 0.816 & {\bf 0.876} & 0.97 & 0.886
\\ \hline
\multirow{2}{*}{RSC} & Confidence & 0.81 & 0.842 & 0.915 & 0.828
\\
& TT-NSS & {\bf 0.911} & {\bf 0.916} & {\bf 0.976} & {\bf 0.891}
\\ \hline
\multirow{2}{*}{SagNet} & Confidence & 0.858 & 0.898 & 0.977 & 0.886
\\
& TT-NSS & 0.869 & {\bf 0.933} & 0.977 & 0.897
\\ \hline
        
       &  & \multicolumn{4}{|c|}{Corrupted with severity 3} \\
      \hline
       
      \multirow{2}{*}{ERM} & Confidence & {\bf 0.79} & {\bf 0.918} & {\bf 0.947} & 0.909
\\
& TT-NSS & 0.771 & 0.898 & 0.878 & {\bf 0.923}
\\ \hline
\multirow{2}{*}{RSC} & Confidence & 0.673 & 0.868 & 0.802 & 0.851
\\
& TT-NSS & {\bf 0.856} & {\bf 0.934} & {\bf 0.941} & {\bf 0.933}
\\ \hline
\multirow{2}{*}{SagNet} & Confidence & 0.842 & 0.913 & 0.948 & 0.873
\\
& TT-NSS & 0.845 & {\bf 0.948} & 0.953 & {\bf 0.924}
\\ \hline
      
      &  & \multicolumn{4}{|c|}{Corrupted with severity 5} \\
      \hline
       
      \multirow{2}{*}{ERM} & Confidence & 0.539 & 0.85 & {\bf 0.852} & 0.845
\\
& TT-NSS & {\bf 0.621} & 0.856 & 0.837 & {\bf 0.888}
\\ \hline
\multirow{2}{*}{RSC} & Confidence & 0.405 & 0.734 & 0.505 & 0.673
\\
& TT-NSS & {\bf 0.719} & {\bf 0.904} & {\bf 0.875} & {\bf 0.903}
\\ \hline
\multirow{2}{*}{SagNet} & Confidence & 0.649 & 0.855 & 0.845 & 0.764
\\
& TT-NSS & {\bf 0.696} & {\bf 0.914} & {\bf 0.878} & {\bf 0.877}
\\ \hline

    \end{tabular}
    }
  \end{center}

\end{table}

\begin{table}
  \begin{center}
    \captionof{table}{%
       Comparison of the area under the accuracy versus percentage of abstained samples curve for TT-NSS and the confidence-based abstaining mechanism in a {\bf multi-}domain setting on different variations of the {\bf VLCS} dataset. The domain used for evaluation is denoted in the columns.
      \label{table:conf_vs_sconf_vlcs_M}
    }
    \resizebox{0.85\columnwidth}{!}{
    \begin{tabular}{|c|c|cccc|}

      \hline
      \multicolumn{2}{|c|}{} & A & C & P & S \\
      
      \hline

      Alg. & Evaluation & \multicolumn{4}{|c|}{Original Style}  \\
      
      \hline
      
      \multirow{2}{*}{ERM} & Confidence & 0.986 & 0.752 & {\bf 0.88} & {\bf 0.831}
\\
& TT-NSS & 0.968 & {\bf 0.772} & 0.86 & 0.776
\\ \hline
      
      &  & \multicolumn{4}{|c|}{Wikiart Style} \\

      \hline
      
      \multirow{2}{*}{ERM} & Confidence & {\bf 0.954} & 0.747 & 0.815 & {\bf 0.691}
\\
& TT-NSS & 0.941 & 0.744 & {\bf 0.822} & 0.678
\\ \hline
        
       &  & \multicolumn{4}{|c|}{Corrupted with severity 3} \\
      \hline
       
      \multirow{2}{*}{ERM} & Confidence & {\bf 0.908} & {\bf 0.601} & 0.678 & {\bf 0.599}
\\
& TT-NSS & 0.785 & 0.553 & {\bf 0.692} & 0.476
\\ \hline
      
      &  & \multicolumn{4}{|c|}{Corrupted with severity 5} \\
      \hline
       
     \multirow{2}{*}{ERM} & Confidence & {\bf 0.775} & {\bf 0.526} & 0.483 & {\bf 0.427}
\\
& TT-NSS & 0.626 & 0.477 & {\bf 0.54} & 0.388
\\ \hline

    \end{tabular}
    }
  \end{center}

\end{table}

\clearpage

\begin{table*}
  \begin{center}
    \captionof{table}{%
       Effectiveness of NSS at producing a better AUC score compared to classifiers trained with ERM in a {\bf multiple} source domain setting on PACS, VLCS, and OfficeHome datasets and their variations when evaluated with TT-NSS. (The target domain used for evaluation is denoted in the columns).
      \label{table:erm_vs_nss_auc_M}
    }
    \resizebox{0.85\textwidth}{!}{
    \begin{tabular}{|c|cccc|cccc|cccc|}
    \hline
      & \multicolumn{4}{|c|}{PACS} & \multicolumn{4}{|c|}{VLCS} & \multicolumn{4}{|c|}{OfficeHome}  \\

      \hline
      \multicolumn{1}{|c|}{Alg.} & A & C & P & S & C & S & L & V & A & C & P & R \\
      
      \hline

       & \multicolumn{12}{|c|}{Original Style}  \\
      
      \hline
      
      ERM & 0.893 & {\bf 0.9} & 0.978 & 0.911 & 0.968 & {\bf 0.772} & 0.86 & 0.776 & 0.683 & 0.679 & 0.815 & 0.83
\\
NSS & {\bf 0.95} & 0.884 & {\bf 0.98} & 0.914 & {\bf 0.985} & 0.769 & 0.865 & {\bf 0.818} & {\bf 0.72} & {\bf 0.749} & {\bf 0.836} & {\bf 0.849}
\\
      
      \hline

      & \multicolumn{12}{|c|}{Wikiart Style}  \\
      
      \hline
      
      ERM & 0.816 & 0.876 & 0.97 & 0.886 & 0.941 & 0.744 & 0.822 & 0.678 & 0.578 & 0.534 & 0.692 & 0.726
\\
NSS & {\bf 0.926} & 0.869 & 0.971 & {\bf 0.909} & {\bf 0.98} & {\bf 0.766} & {\bf 0.85} & {\bf 0.775} & {\bf 0.667} & {\bf 0.713} & {\bf 0.798} & {\bf 0.825}
\\
      
      \hline

      & \multicolumn{12}{|c|}{Corrupted with severity 3}  \\
      
      \hline
      
      ERM & 0.771 & 0.898 & 0.878 & 0.923 & 0.785 & 0.553 & 0.692 & 0.476 & 0.5 & 0.64 & 0.677 & 0.715
\\
NSS & {\bf 0.889} & {\bf 0.933} & {\bf 0.943} & {\bf 0.933} & {\bf 0.959} & {\bf 0.605} & {\bf 0.706} & {\bf 0.632} & {\bf 0.587} & {\bf 0.697} & {\bf 0.738} & {\bf 0.812}
\\
      
      \hline

      & \multicolumn{12}{|c|}{Corrupted with severity 5}  \\
      
      \hline
      
      ERM & 0.621 & 0.856 & 0.837 & 0.888 & 0.626 & 0.477 & 0.54 & 0.388 & 0.387 & 0.53 & 0.554 & 0.59
\\
NSS & {\bf 0.792} & 0.854 & {\bf 0.88} & {\bf 0.902} & {\bf 0.898} & {\bf 0.53} & {\bf 0.611} & {\bf 0.517} & {\bf 0.473} & {\bf 0.648} & {\bf 0.628} & {\bf 0.721}
\\
      
      \hline

    \end{tabular}
    }
  \end{center}

\end{table*}


      

      
      
      

      
        
       
      
       



\clearpage

\clearpage

\begin{table*}
  \begin{center}
    \captionof{table}{%
       Effectiveness of NSS at producing a better AUC score compared to classifiers trained with ERM in a {\bf single} source domain setting on PACS, VLCS, and OfficeHome datasets and their variations when evaluated with the confidence-based abstaining mechanism. (The source domain used for training is denoted in the columns).
      \label{table:erm_vs_nss_auc_conf}
    }
    \resizebox{0.85\textwidth}{!}{
    \begin{tabular}{|c|cccc|cccc|cccc|}
    \hline
       & \multicolumn{4}{|c|}{PACS} & \multicolumn{4}{|c|}{VLCS} & \multicolumn{4}{|c|}{OfficeHome} \\

      \hline
      \multicolumn{1}{|c|}{Alg.} & A & C & P & S & C & L & S & V & A & C & P & R \\
      
      \hline

      & \multicolumn{12}{|c|}{Original Style}  \\
      
      \hline
      
      ERM & 0.882 & 0.875 & 0.634 & {\bf 0.707} & 0.653 & 0.68 & 0.806 & 0.715 & 0.743 & 0.717 & 0.699 & {\bf 0.789}
\\
NSS & {\bf 0.907} & {\bf 0.923} & {\bf 0.733} & 0.665 & {\bf 0.671} & 0.687 & {\bf 0.838} & {\bf 0.74} & 0.739 & 0.72 & 0.708 & 0.778
\\
      
      \hline

      & \multicolumn{12}{|c|}{Wikiart Style}  \\
      
      \hline
      
      ERM & 0.84 & 0.757 & 0.609 & {\bf 0.558} & 0.426 & 0.584 & 0.763 & 0.679 & 0.545 & 0.364 & 0.334 & 0.484
\\
NSS & {\bf 0.871} & {\bf 0.885} & {\bf 0.672} & 0.526 & {\bf 0.535} & {\bf 0.655} & {\bf 0.816} & {\bf 0.722} & {\bf 0.705} & {\bf 0.658} & {\bf 0.64} & {\bf 0.749}
\\
      
      \hline

      & \multicolumn{12}{|c|}{Corrupted with severity 3}  \\
      
      \hline
      
      ERM & 0.832 & 0.709 & 0.613 & {\bf 0.612} & 0.504 & 0.381 & 0.734 & 0.468 & 0.596 & 0.412 & 0.411 & 0.586
\\
NSS & {\bf 0.871} & {\bf 0.865} & {\bf 0.754} & 0.549 & {\bf 0.592} & {\bf 0.631} & {\bf 0.771} & {\bf 0.522} & {\bf 0.666} & {\bf 0.586} & {\bf 0.566} & 0.595
\\
      
      \hline

      & \multicolumn{12}{|c|}{Corrupted with severity 5}  \\
      
      \hline
      
      ERM & 0.696 & 0.579 & 0.418 & {\bf 0.479} & 0.433 & 0.329 & 0.563 & 0.346 & 0.416 & 0.243 & 0.223 & 0.388
\\
NSS & {\bf 0.769} & {\bf 0.746} & {\bf 0.667} & 0.434 & 0.454 & {\bf 0.576} & {\bf 0.635} & {\bf 0.4} & {\bf 0.546} & {\bf 0.49} & {\bf 0.415} & {\bf 0.42}
\\
      
      \hline

    \end{tabular}
    }
  \end{center}

\end{table*}

\begin{table*}
  \begin{center}
    \captionof{table}{%
       Effectiveness of NSS at producing a better AUC score compared to classifiers trained with ERM in a {\bf multiple} source domain setting on PACS, VLCS, and OfficeHome datasets and their variations when evaluated with the confidence-based abstaining mechanism. (The target domain used for evaluation is denoted in the columns).
      \label{table:erm_vs_nss_auc_conf_M}
    }
    \resizebox{0.85\textwidth}{!}{
    \begin{tabular}{|c|cccc|cccc|cccc|}
    \hline
       & \multicolumn{4}{|c|}{PACS} & \multicolumn{4}{|c|}{VLCS} & \multicolumn{4}{|c|}{OfficeHome} \\

      \hline
      \multicolumn{1}{|c|}{Alg.} & A & C & P & S & C & L & S & V & A & C & P & R \\
      
      \hline

       & \multicolumn{12}{|c|}{Original Style}  \\
      
      \hline
      
      ERM & 0.95 & 0.902 & 0.986 & 0.915& 0.986 & {\bf 0.752} & {\bf 0.88} & 0.831 & 0.802 & 0.721 & 0.889  & {\bf 0.905}
\\
NSS & 0.955 & 0.896 & 0.985 & 0.922 & 0.987 & 0.706 & 0.86 & 0.829 & 0.783 & {\bf 0.767} & 0.876 & {\bf 0.884}
\\
      
      \hline

      & \multicolumn{12}{|c|}{Wikiart Style}  \\
      
      \hline
      
      ERM & 0.898 & 0.85 & 0.975 & 0.892 & 0.954 & {\bf 0.747} & 0.815 & 0.691 & 0.601 & 0.588 & 0.726 & 0.796
\\
NSS & {\bf 0.927} & {\bf 0.898} & 0.982 & {\bf 0.92} & {\bf 0.982} & 0.705 & {\bf 0.833} & {\bf 0.781} & {\bf 0.707} & {\bf 0.747} & {\bf 0.829} & {\bf 0.838}
\\
      
      \hline

      & \multicolumn{12}{|c|}{Corrupted with severity 3}  \\
      
      \hline
      
      ERM & 0.79 & 0.918 & 0.947 & 0.909 & 0.908 & 0.601 & 0.678 & 0.599 & 0.529 & 0.584 & 0.74 & 0.717
\\
NSS & {\bf 0.887} & 0.909 & 0.955 & {\bf 0.922} & {\bf 0.966} & 0.594 & {\bf 0.735} & {\bf 0.627} & {\bf 0.647} & {\bf 0.735} & {\bf 0.775} & {\bf 0.808}
\\
      
      \hline

      & \multicolumn{12}{|c|}{Corrupted with severity 5}  \\
      
      \hline
      
      ERM & 0.539 & 0.85 & 0.852 & 0.845 & 0.775 & {\bf 0.526} & 0.483 & 0.427 & 0.362 & 0.475 & 0.581 & 0.551
\\
NSS & {\bf 0.735} & {\bf 0.881} & {\bf 0.887} & 0.833 & {\bf 0.91} & 0.508 & {\bf 0.621} & {\bf 0.44} & {\bf 0.528} & {\bf 0.66} & {\bf 0.672} & {\bf 0.688}
\\
      
      \hline

    \end{tabular}
    }
  \end{center}

\end{table*}

\end{document}